\documentclass[accepted]{uai2023} 

\usepackage[american]{babel}
\usepackage{amsfonts}
\usepackage{float}
\usepackage{amsmath}
\usepackage{placeins}

\newcommand{\x}{\boldsymbol{x}}
\newcommand{\z}{\boldsymbol{z}}

\newcommand{\h}{\boldsymbol{h}}

\newcommand{\xib}{\boldsymbol{\xi}}

\newcommand{\thetab}{\boldsymbol{\theta}}
\newcommand{\phib}{\boldsymbol{\phi}}
\newcommand{\psib}{\boldsymbol{\psi}}

\newcommand{\etab}{\boldsymbol{\eta}}

\newcommand{\given}{\,|\,}
\newcommand{\lik}{p(\x \given \thetab)}

\newcommand{\post}{p(\thetab \given \x)}

\newcommand{\prior}{p(\thetab)}
\newcommand{\joint}{p(\thetab, \x)}

\newcommand{\noised}{p(\xib \given \thetab)}
\newcommand{\model}{G(\thetab, \xib)}
\newcommand{\diff}{\mathrm{d}}

\newcommand{\lika}{l_{\etab}(\x \given \thetab)}
\newcommand{\posta}{p_{\phib}(\thetab \given \x)}
\newcommand{\postah}{p_{\phib}(\thetab \given \mathcal{H}_{\psib}(\x))}

\usepackage{algorithm}
\usepackage{algorithmic}

\usepackage{subcaption}
\DeclareMathOperator*{\argmin}{arg\,min}

\newcommand{\twomoonsposterior}[2]{
	\begin{subfigure}[t]{0.40\linewidth}
		\includegraphics[width=\linewidth]{#2}
		\caption{#1}
	\end{subfigure}
}

\newcommand{\covidscenario}[2]{
	\begin{subfigure}[t]{0.40\linewidth}
		\frame{\includegraphics[width=\linewidth]{#2}}
		\caption{#1}
	\end{subfigure}
}

\newcommand{\appendixbenchmark}[3]{
	\begin{figure}
		\centering
		\begin{subfigure}[t]{0.7\linewidth}
			\centering
			\includegraphics[width=1.0\linewidth]{plots/benchmarks/appendix/#2_losses.pdf}
			\caption{Training and validation loss history.}
		\end{subfigure}
		\begin{subfigure}[t]{1.0\linewidth}
			\centering
			\includegraphics[scale=0.25]{plots/benchmarks/appendix/#2_posterior_calibration_diff_separate.pdf}
			\caption{Posterior calibration.}
		\end{subfigure}
		\begin{subfigure}[t]{1.0\linewidth}
			\centering
			\includegraphics[scale=0.25]{plots/benchmarks/appendix/#2_joint_calibration_diff_separate.pdf}
			\caption{Joint calibration.}
		\end{subfigure}
		\caption{\textbf{Benchmark #1, #3.} Loss history, posterior calibration, and joint calibration.}
		\label{fig:app:benchmark:#2}
	\end{figure}
}

\newcommand{\appendixbenchmarkcompressed}[3]{
	\begin{figure}
		\centering
		\begin{subfigure}[t]{0.7\linewidth}
			\centering
			\includegraphics[width=1.0\linewidth]{plots/benchmarks/appendix/#2_losses.pdf}
			\caption{Training and validation loss history.}
		\end{subfigure}\\
		\vspace{0.5cm}
		\begin{subfigure}[t]{0.40\linewidth}
			\includegraphics[width=\linewidth]{plots/benchmarks/appendix/#2_posterior_calibration_diff_separate.pdf}
			\caption{Posterior calibration.}
		\end{subfigure}
		\hspace{1cm}
		\begin{subfigure}[t]{0.40\linewidth}
			\includegraphics[width=\linewidth]{plots/benchmarks/appendix/#2_joint_calibration_diff_separate.pdf}
			\caption{Joint calibration.}
		\end{subfigure}
		\caption{\textbf{Benchmark #1, #3.} Loss history, posterior calibration, and joint calibration.}
		\label{fig:app:benchmark:#2}
	\end{figure}
}

\usepackage[backend=biber,
style=apa,
maxcitenames=2,
minbibnames=2,
maxbibnames=2, 
language=english,
doi=false,
isbn=false,
url=false,
uniquename=false
]{biblatex} 
\DeclareLanguageMapping{english}{english-apa} 
\usepackage{mathtools} 
\usepackage{booktabs} 
\usepackage{tikz} 
\usepackage{xcolor}
\definecolor{darkblue}{RGB}{0,0,128}
\hypersetup{
    colorlinks=true,
	linkcolor=darkblue,
	filecolor=darkblue,
	urlcolor=darkblue,
	citecolor=darkblue,
	pdfauthor={},
	pdftitle={}
}

\newcommand{\new}[1]{\textcolor{black}{#1}}

\title{JANA: Jointly Amortized Neural Approximation of Complex\\Bayesian Models}

\author[1]{\vspace{-1.5em}Stefan T.~Radev}
\author[2]{Marvin Schmitt}
\author[3]{Valentin Pratz}
\author[4]{Umberto Picchini}
\author[3]{\authorcr Ullrich K\"othe$^*$}
\author[2]{Paul-Christian B\"urkner\thanks{Shared senior authorship}}
\affil[1]{%
Cluster of Excellence STRUCTURES\\
Heidelberg University
}
\affil[2]{%
Cluster of Excellence SimTech\\
University of Stuttgart
}
\affil[3]{%
Visual Learning Lab\\
Heidelberg University
}
\affil[4]{%
Department of Mathematical Sciences\\
Chalmers University of Technology \& University of Gothenburg
}

\begin{document}
\maketitle
\begin{abstract}
This work proposes ``jointly amortized neural approximation'' (JANA) of intractable likelihood functions and posterior densities arising in Bayesian surrogate modeling and simulation-based inference. We train three complementary networks in an end-to-end fashion: 1) a summary network to compress individual data points, sets, or time series into informative embedding vectors; 2) a posterior network to learn an amortized approximate posterior; and 3) a likelihood network to learn an amortized approximate likelihood. Their interaction opens a new route to amortized marginal likelihood and posterior predictive estimation -- two important ingredients of Bayesian workflows that are often too expensive for standard methods. We benchmark the fidelity of JANA on a variety of simulation models against state-of-the-art Bayesian methods and propose a powerful and interpretable diagnostic for joint calibration. In addition, we investigate the ability of recurrent likelihood networks to emulate complex time series models without resorting to hand-crafted summary statistics.
\end{abstract}

\section{Introduction}
\label{sec:intro}

Surrogate modeling (SM) and simulation-based inference (SBI) are two 
ingredients of the new generation of methods for simulation science \autocite{lavin2021simulation}.
From a Bayesian perspective, SM seeks to approximate the intractable likelihood function, whereas SBI targets 
the intractable posterior distribution of a complex probabilistic model.
Both problems are hard, as they involve 
integrals which cannot be solved with standard analytical or numerical methods.
Thus, specialized neural approximators have emerged as promising tools for taming the intractable \autocite{cranmer2020frontier}.

\begin{figure*}[t]
    \centering
    \includegraphics[width=\linewidth]{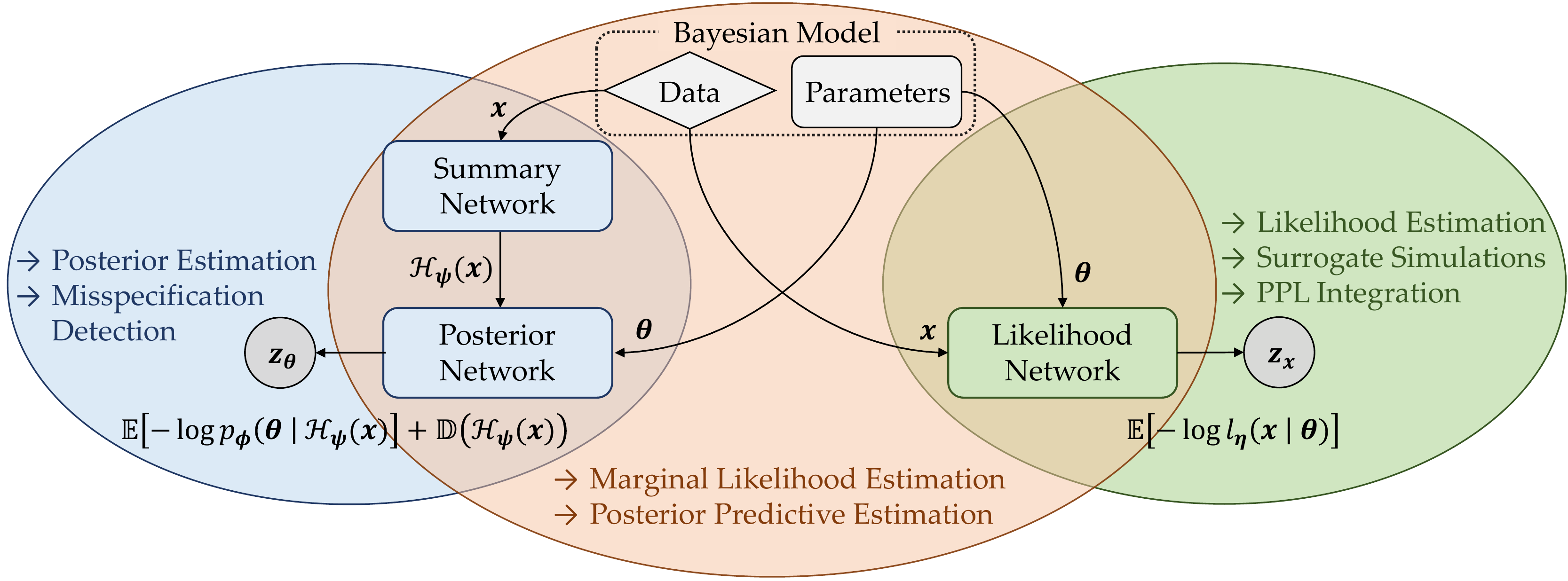}
    \caption{A conceptual illustration of our method for jointly amortized neural approximation (JANA). On the one hand, the summary and posterior network can perform amortized posterior estimation and detect model misspecification. On the other hand, the likelihood network can perform amortized likelihood estimation, surrogate simulations, and interact with probabilistic programming languages. Together, the two networks enable posterior predictive and marginal likelihood estimation, which allow for amortized Bayesian model comparison and validation.
    }
    \label{fig:conceptual}
\end{figure*}

\new{Neural networks trained on model simulations enable \textit{amortized inference}: A pre-trained network can be stored and re-used for Bayesian inference on millions of data sets \autocite{von2022mental}. Crucially, most previous neural approaches have tackled either SM or SBI in isolation, but little attention has been paid to learning both tasks simultaneously.}
\new{To address this gap}, we propose JANA (``Jointly Amortized Neural Approximation''), a Bayesian neural framework for {\em simultaneously amortized} SM and SBI, and show how it enables \new{novel} solutions to challenging downstream tasks like the estimation of marginal and posterior predictive distributions (see \autoref{fig:conceptual}).
JANA also presents a major qualitative upgrade to the BayesFlow framework \autocite{radev2020bayesflow}, which was originally designed for amortized SBI alone.

It is commonly presumed that amortized inference is wasteful \autocite{greenberg2019automatic, papamakarios2016fast} and requires much larger simulation budgets than sequential inference to make up for the much larger prediction domain.
Our results challenge this premise.
Given identical simulation budgets, JANA outperforms or is on par with sequential (i.e., non-amortized) methods, such as ABC-SMC, SNL, and SNPE (see~\autoref{fig:two-moons:posterior}).
Furthermore, we hypothesize that modern neural networks benefit strongly from a broad simulation scope.
Thanks to their excellent generalization capabilities, they can exploit outcomes from the entire prior predictive distribution of a simulation to improve local accuracy for each specific case.
In this sense, amortized inference seems to be a natural by-product of deep probabilistic modeling, and the initial training effort more than repays with global diagnostics, nearly instant estimation at test time, and no loss in accuracy.

We show that JANA unlocks the potential of powerful Bayesian tools for model comparison, validation, and calibration, which are essential in Bayesian workflows \autocite{gelman2020bayesian}, but widely underutilized in current simulation-based analysis.
For one, JANA offers an efficient way to compute \textit{marginal likelihoods} via the probabilistic change-of-variables formula (instead of integration over the model's entire prior space) as a prerequisite for \textit{prior predictive} model selection (i.e., probabilistic Occam's razor).
For another, it can rapidly produce both posterior samples and normalized likelihood estimates of new data instances, as are needed in strong validation procedures of the \textit{posterior predictive} performance \autocite{vehtari_survey_2012}. 
In other words, JANA can directly quantify both prior and posterior predictive performance without resorting to Markov chain Monte Carlo (MCMC) sampling or costly model re-fits, in addition to the well-studied advantages of individual posterior or likelihood networks (see~\autoref{fig:conceptual}). 

In summary, our key contributions are:
\begin{enumerate}
    \item We develop a neural architecture for fully amortized joint posterior estimation and likelihood emulation;
    \item We propose a sensitive and interpretable method to test for joint calibration of the networks;
    \item We extensively validate our new architecture on analytic toy examples and complex simulation models;
    \item We show how our joint architecture solves the challenges of computing both out-of-sample predictive performance and intractable marginal likelihoods;
    \item We demonstrate a recurrent neural likelihood for surrogate simulations in a complex time series model.
\end{enumerate}

\section{Method}
\label{sec:method}

\subsection{Problem Formulation}

\paragraph{Bayesian Models}
We focus on generative Bayesian models specified as a triple $\mathcal{M}{=}\big( \model, \noised, \prior \big)$.
Such models yield observables $\x \in \mathcal{X}$ according to the system
\begin{equation}\label{eq:model}
    \x = \model \quad\textrm{with}\quad \xib \sim \noised,\; \thetab \sim \prior,
\end{equation}
where $G$ denotes a simulation program, $\xib \in \Xi$ denotes externalized randomness (i.e., noise or pseudorandom program states) with density function $\noised$, and $\prior$ encodes prior knowledge about plausible simulation parameters $\thetab \in \Theta$.

\paragraph{Forward Inference}
Running the simulator $G$ with a fixed parameter configuration $\thetab$ and different values of $\xib$ is equivalent to random draws from an \textit{implicit likelihood} $\lik$:
\begin{equation}
    \x \sim \lik \Longleftrightarrow \x =\model \quad \textrm{with} \quad \xib \sim \noised
\end{equation}
In theory, implicit likelihoods can be obtained by marginalizing the joint distribution $p(\xib, \x \given \thetab)$ over all possible execution trajectories of the simulation program (i.e., over $\xib$), but this is typically intractable \autocite{cranmer2020frontier}.

\paragraph{Inverse Inference}
In Bayesian analysis, we want to infer a model's latent parameters $\thetab$ from manifest data $\x$ through the probabilistic factorization of the joint distribution into prior and (implicit) likelihood:
\begin{equation}
    \post \propto \joint = \prior \int_{\Xi} p(\xib, \x \given \thetab)\,\diff \xib.
\end{equation}
Since we assume that the likelihood is not available in closed form, we also cannot access the posterior $\post$ and perform parameter estimation through gold-standard Bayesian methods, such as MCMC \autocite{carpenter2017stan}.

\paragraph{Marginal Likelihoods}
In addition to estimating parameters, modelers often want to compare and assign preferences to competing models.
From a Bayesian perspective, the canonical measure of evidence for a given model is the \textit{marginal likelihood} (aka the \textit{prior predictive distribution}),
\begin{equation}
    p(\x) = \int_{\Theta} \int_{\Xi} \prior\,p(\xib, \x \given \thetab)\,\diff \xib\,\diff \thetab, \label{eq:marg_lik}
\end{equation}
which is doubly intractable for complex models because both involved integrals are highly difficult to approximate with sufficient precision \autocite{meng_simulating_1996}.
However, the estimation of the marginal likelihood is central to Bayesian model comparison, since it naturally embodies a probabilistic version of Occam’s razor by penalizing the prior complexity of a model \autocite{mackay2003information}.
Thus, it allows us to express our preference for a simpler model over a more complex one, given that both models can account for the observed data equally well.

\paragraph{Posterior Predictive Distribution}
Bayesian models can also be compared and validated on the basis of their posterior predictive performance \autocite{vehtari_survey_2012}.
However, many posterior predictive metrics rely on the likelihood density being available analytically. 
In particular, this is true for the expected log-predictive density (ELPD), which is a widely-applied, general-purpose metric to measure (out-of-sample) posterior predictive performance when no application-specific utilities are known \autocite{vehtari2017practical}. 
For $K$ (new) observations \smash{$\x^{(k)}_{\rm new}$} not previously seen by the model, the ELPD can be defined as
\begin{equation}\label{elpd}
    \text{ELPD} = \sum_{k=1}^{K} \log \int_{\Theta} p(\x^{(k)}_{\rm new} \given \thetab) \, \post \, \diff \thetab.
\end{equation}

The ELPD has a strong connection to information theory \autocite{vehtari_survey_2012} and is widely used in Bayesian cross-validation \autocite{vehtari2017practical}, where it is one of the most prominent sources of computational intractability.

\paragraph{Probabilistic Symmetry}
Our joint training will leverage the symmetry in the arguments of $\post$ and $\lik$, along with the fact that a single run of the simulator (Eq.~\ref{eq:model}) yields a reusable tuple of parameters and synthetic data $(\thetab, \x)$.
However, many simulation models are characterized by a relatively low-dimensional parameter space $\Theta$ (e.g., low-dimensional vectors) and a rather high-dimensional data space with a rich structure $\mathcal{X}$ (e.g., multivariate time series or sets of exchangeable observations).
Thus, we need different neural architectures, each separately aligned with the structural properties of $\post$ and $\lik$.


\begin{figure}[t]
    \centering
    \includegraphics[width=\linewidth]{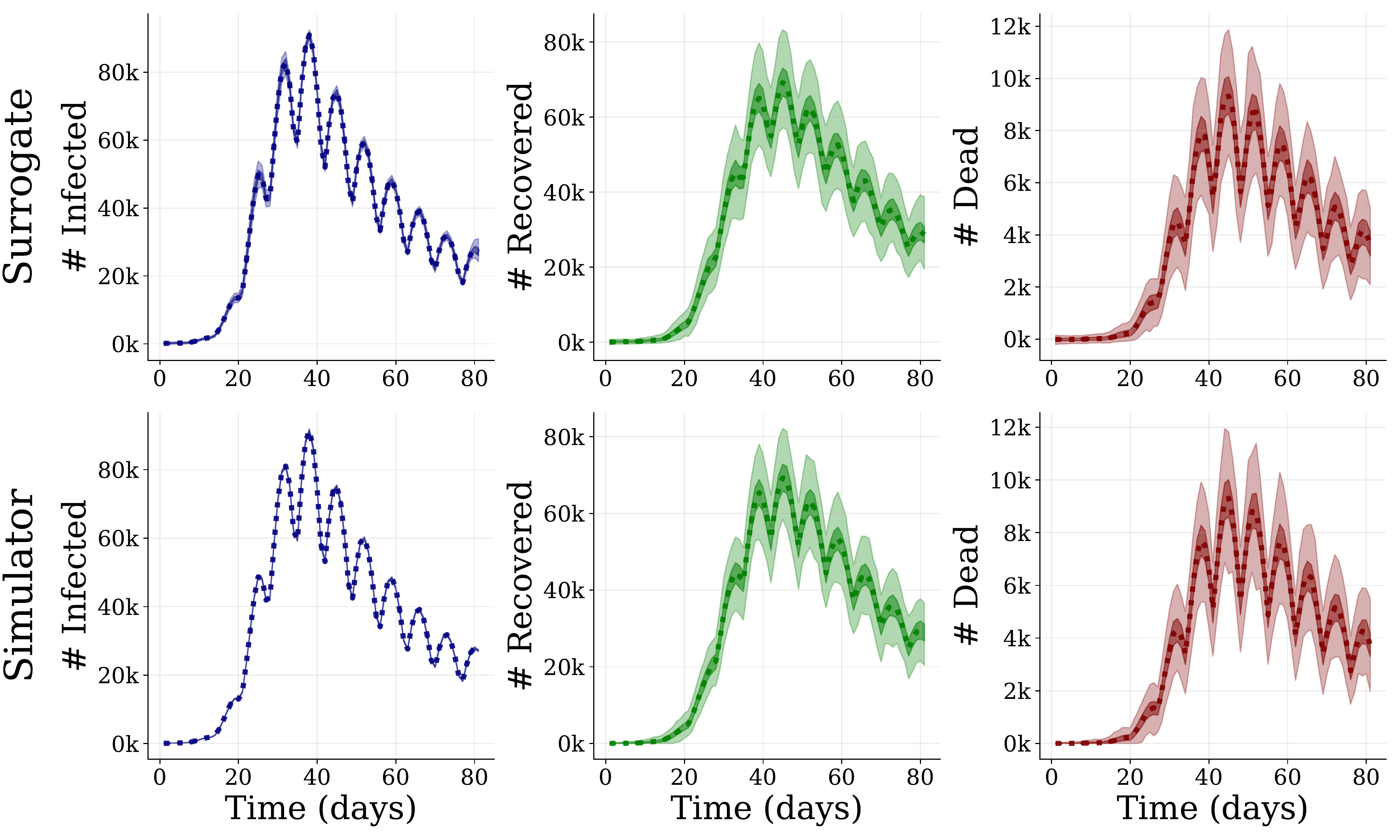}\\
    \includegraphics[width=0.3\linewidth]{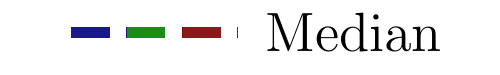}
    \includegraphics[width=0.3\linewidth]{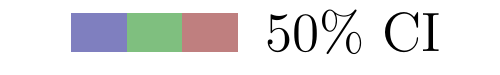}
    \includegraphics[width=0.3\linewidth]{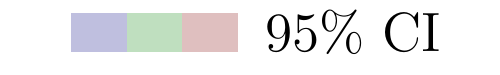}
    \caption{Recurrent likelihood networks can emulate complex Bayesian stochastic differential equation models of disease outbreaks (see~\textbf{Experiment 4}).The top and bottom row each depict $1\,000$ simulations (same $\thetab$) from the surrogate and the actual simulator, respectively.}
    \label{fig:covid}
\end{figure}

\subsection{Posterior Network}
\label{sec:post_net}

The posterior network $\mathcal{P}_{\phib}$ implements a normalizing flow between $\thetab$ and a latent variable $\z_{\thetab}$ with a simple density (e.g., Gaussian) given observed or simulated data $\x$:
\begin{align}
\posta &= p(\z_{\thetab})\,\left|\det \left(\frac{\partial \z_{\thetab}}{\partial \thetab}\right)\right| \label{eq:post}\\
\z_{\thetab} &= \mathcal{P}_{\phib}(\thetab;\x).
\end{align} 
The normalizing flow is realized via a conditional invertible neural network (cINN) composed by a series of conditional coupling layers with affine and/or spline transformations.
Since the observed or simulated data will typically have a complex structure and/or contain varying numbers of observations, the posterior cINN includes a trainable summary network sub-module $\mathcal{H}_{\psib}$ \autocite[see][]{radev2020bayesflow} which we optimize alongside to extract maximally informative data representations $\mathcal{H}_{\psib}(\x)$ in an end-to-end manner.

The design of the conditional coupling layers follows the work of \autocite{durkan2019neural, ardizzone2019guided, ardizzone2018analyzing}, since compositions of such layers exhibit favorable theoretical properties \autocite{draxler2022whitening} and remarkable empirical performance on high-dimensional unstructured data \autocite{kingma2018glow, dinh2016density} or complex Bayesian models in various domains \autocite{bellagente2022understanding, von2022mental, radev2021outbreakflow, bieringer2021measuring}.
However, any other coupling design can be used as a plug-in replacement.

\subsection{Likelihood Network}
\label{sec:lik_net}

The likelihood network $\mathcal{L}_{\etab}$ implements a normalizing flow between $\x$ and a (multivariate) Gaussian latent variable $\z_{\x}=\mathcal{L}_{\etab}(\x;\thetab)$ given a parameter configuration $\thetab$,
\begin{align}
\lika = p(\z_{\x})\,\left|\det \left(\frac{\partial \z_{\x}}{\partial \x}\right)\right|. \label{eq:lik}
\end{align} 
This formulation is similar to the pushforward expression for the posterior network (Eq.~\ref{eq:post}), but with $\thetab$ swapped for $\x$.
The likelihood network, like the posterior network, is also implemented as a cINN. 
As the conditioning information is now the parameter vector $\thetab$ (and not a complex data structure), it can be fed directly to the conditional coupling layers of the cINN without an additional summary network. 

However, since the data $\x$ (i.e., simulator outputs) is typically in non-vector form, the design of the coupling layers needs to be tailored according to the probabilistic symmetry of $\lik$.
Learning $\lik$ in its raw form is typically much harder than learning the likelihood $p(\mathcal{H}(\x) \given \thetab)$ of some (learned or hand-crafted) summary statistics $\mathcal{H}(\x)$, since the latter are already in a compressed vector form and do not require specialized architectures. 
JANA can learn either $p(\mathcal{H}(\x) \given \thetab)$ or $\lik$, as required by the particular application or dictated by the (un-)availability of good summary statistics.
In our experiments, we directly target $\lik$ and the \textbf{Appendix} details how to design likelihood networks for exchangeable or Markovian data.

\subsection{Simulation-based training}

In contrast to previous joint learning approaches \autocite{wiqvist2021sequential, glockler2022snvi}, we aim for a fully amortized approach: Once the networks have converged, we want to evaluate the normalized densities $\posta$ and $\lika$ for \textit{any} pair $(\thetab, \x)$ consistent with a generative Bayesian model $\mathcal{M}$.
In addition, we want to generate conditional random draws $\thetab \given \x$ and $\x \given \thetab$ from both networks for parameter estimation and surrogate modeling.
Finally, we want to prescribe a simple distribution to the summary network outputs $p\big(\mathcal{H}_{\psib}(\x)\big)$ in order to detect atypical data during inference (i.e., model misspecification) and highlight potential posterior errors \autocite{schmitt2022bayesflow}.
Thus, we minimize the following criterion:
\begin{equation}\label{eq:cost}
\begin{aligned}
    \min_{\phib, \psib, \etab} \mathbb{E}_{ \joint}\big[&-\left(\log p_{\phib}(\thetab \given \mathcal{H}_{\psib}(\x) \right) + \log l_{\etab}(\x \given \thetab))  \big] \\&+ \lambda \cdot \mathbb{MMD}^2\big[p(\mathcal{H}_{\psib}(\x))\,||\,\mathcal{N}(\boldsymbol{0}, \mathbb{I})\big]
\end{aligned}
\end{equation}
where $\mathbb{MMD}^2$ is the maximum mean discrepancy \autocite[MMD;][]{Gretton2012} between the distribution of summary network outputs and a unit Gaussian density.
This divergence imposes a probabilistic structure on the summary space learned by $\mathcal{H}_{\psib}(\x)$ and enables error detection and model criticism during inference \autocite[to be explained shortly, see also][]{schmitt2022bayesflow}.
\new{We approximate the expectation over $\joint$ via online or offline simulations from the generative model $\mathcal{M}$ and train the three networks until convergence (see the \textbf{Appendix} for a detailed derivation of simulation-based training).}

Proper minimization of the criterion in Eq.~\ref{eq:cost} results in correct posterior and likelihood approximation, along with an interpretable summary space.
However, the objective promises self-consistency only in the ``small world'', as it does not guarantee correct posterior inference or likelihood evaluation in the real world when there may be a severe simulation gap.
This is due to the fact that simulation-based training optimizes the expectation with respect to the Bayesian joint model $\joint$, but not (necessarily) the empirical data distribution $p^*(\x)$.
Thus, the MMD term allows us to detect potential simulation gaps during inference via distribution matching \autocite{schmitt2022bayesflow}.
Moreover, the posterior network can serve as a ``critic'' for the likelihood network by rejecting surrogate simulations which are judged to be highly unlikely under the true simulator. 

\subsection{Validation Methodology: Joint Calibration}
\label{sec:val}

Faithful uncertainty representation (i.e., calibration) is an essential precondition for self-consistent and interpretable simulation-based inference.  
Simulation-based calibration \autocite[SBC;][]{talts2018validating} is a general diagnostic method which considers the performance of a sampling algorithm over the entire joint distribution $p(\thetab, \x)$, regardless of the specific probabilistic structure of a model.

SBC leverages the generative nature of Bayesian models as well as the self-consistency of the Bayesian joint model $\model$ in the following sense: For all quantiles $q \in (0, 1)$, all uncertainty regions $U_q(\thetab \mid \x)$ of $\post$ are well calibrated, as long as the generating distribution of the assumed model is equal to true data-generating distribution and posterior computation is exact \autocite{talts2018validating}. 
We can formally write this property as
\begin{equation}
\label{eq:sbc}
  q = \int_{\mathcal{X}} \int_{\Theta} \mathbb{I}_{\left[ \thetab^* \in U_q(\thetab \given \x) \right]} \, p(\x \given \thetab^*) \, p(\thetab^*) \, \diff \x \, \diff \thetab^*,
\end{equation}
where $\thetab^*$ is the true data-generating parameter and $\mathbb{I}_{\left[\cdot\right]}$ is the indicator function.
If the posterior network $\mathcal{P}_{\phib}$ generates draws from the true posterior and the likelihood network $\mathcal{L}_{\etab}$ mimics the simulator perfectly, then the equality implied by Eq.~\ref{eq:sbc} holds regardless of the particular form of the true likelihood or the true posterior. 
Thus, any violation of this equality indicates some error incurred by joint training, so we refer to our validation procedure as joint simulation-based calibration (JSBC).

The reasons for faulty JSBC can be any combination of (i) inaccurate representation of the posterior; (ii) inaccurate representation of the likelihood; or (iii) an erroneous implementation of the simulation model itself. 
To differentiate between (i) and (ii), we can first run standard SBC for the posterior network using data draws from the actual simulator instead of the likelihood network. 
If this check passes, but subsequently JSBC fails, the calibration problems must stem from the likelihood network.
Thereby, we can use the posterior network for \emph{model criticism} of the likelihood network, which would otherwise be infeasible for most Bayesian models.

As part of a Bayesian workflow \autocite{gelman2020bayesian}, calibration procedures can quickly become infeasible for non-amortized methods, as they require independent posterior draws from hundreds or thousands of simulated data sets.
However, we can effortlessly assess the calibration of amortized methods, since we can obtain many posterior draws from thousands of data sets in a matter of seconds.
In practice, we follow \textcite{sailynoja2022graphical} by transforming the posterior draws intro fractional rank statistics and computing their empirical cumulative distribution functions (ECDFs). 
This method provides \textit{simultaneous confidence bands} and eliminates the need to manually select a binning parameter (e.g., as required by histogram-based methods).

\subsection{Use Cases for Joint Learning}


\paragraph{Posterior Predictive Estimation}
Estimating the expected predictive performance of a Bayesian model (Eq.~\ref{elpd}) requires an analytic expression for the pointwise \smash{$p(\x^{(k)}_{\rm new} \given \thetab)$} at arbitrary new data \smash{$\x^{(k)}_{\rm new}$} \autocite{burkner_nfloo_2021}.
For this reason, the ELPD cannot be computed for Bayesian models with intractable likelihoods or sequential neural estimators.

Moreover, even if the likelihood itself were analytic, the integral in Eq.~\eqref{elpd} would still be intractable for most models. 
It can be efficiently approximated using posterior draws, but doing so in the context of cross-validation requires importance sampling or costly model refits \autocite{vehtari2017practical}.
Hence, evaluating the ELPD for arbitrary cross-validation schemes critically requires both the amortized likelihood and posterior approximator.

Given data used for model fitting $\x$ and upcoming data \smash{$\x^{(k)}_{\rm new}$}, the two networks can estimate a model’s expected predictive performance in two steps.
First, we can obtain a large amount of $S$ random draws from the amortized posterior given $\x$:
\begin{equation}
    \thetab^{(s)} \sim \postah \textrm{ for } s = 1,...,S.
\end{equation}
Then, the likelihood network can approximate the ELPD at all $\x_{\rm new}^{(k)}$ given $\{\thetab^{(s)}\}$ via its Monte Carlo estimate:
\begin{equation}\label{eq:elpd}
    \widehat{\textrm{ELPD}} = \sum_{k=1}^{K} \log \frac{1}{S} \sum_{s=1}^S l_{\psib}(\x^{(k)}_{\rm new} \given \thetab^{(s)})
\end{equation}
In the context of cross-validation (CV), $\x$ and \smash{$\x_{\rm new}$} refer to a random data split, and we can estimate the predictive performance of a Bayesian model by summing over the \smash{$\widehat{\textrm{ELPDs}}$} from all data splits. 
In \textbf{Experiment 3}, we demonstrate this for leave-one-out (LOO)-CV, which is one of the most expensive validation methods.

\paragraph{Marginal Likelihood Estimation}
Bayesian (prior) predictive model comparison depends on computing a marginal likelihood (Eq.~\ref{eq:marg_lik}).
We can leverage the probabilistic change of variable, which results directly from Bayes' rule:
\begin{align}\label{eq:log_marg}
    \log \widehat{p}(\x) &= \log \lika + \log\prior\\ &- \log \postah \nonumber.
\end{align}
Thus, for any data set, we can obtain an estimate of the log marginal likelihood by evaluating Eq.~\ref{eq:post} and Eq.~\ref{eq:lik}, along with the prior density $\prior$.
Evaluating all above terms is infeasible with standard Bayesian methods, since either the normalized posterior, the likelihood, or both quantities are typically intractable. 
Bridge sampling \autocite{meng_simulating_1996} enables the approximation of marginal likelihoods from posterior draws, but only works for models with analytical likelihoods and in tandem with non-amortized MCMC.

From a Bayesian perspective, evaluating Eq.~\ref{eq:log_marg} across multiple data sets amounts to \textit{amortized bridge sampling}.
At the same time, we can use Eq.~\ref{eq:log_marg} for assessing non-convergence or problems during inference by evaluating the right-hand side for a fixed $\x$ and different $\thetab$ drawn from the approximate posterior.
Under perfect convergence, the right-hand side of Eq.~\ref{eq:log_marg} is independent of $\thetab$, so any ensuing variation is a measure of pure approximation error.

\textbf{Surrogate Simulators}
In some modeling scenarios, the simulator might be a large-scale computer program implementing a complex generative algorithm \autocite{lavin2021simulation}.
Thus, a simulation-based inference workflow might be severely limited by the inability to obtain a large amount of simulations in a reasonable time. 
In such cases, an amortized surrogate simulator can generate additional data for the posterior network or a black-box optimizer \autocite{gutmann2016bayesian}.
A notable advantage of neural surrogate simulators is that they can directly emulate complex data without summary statistics (see~\autoref{fig:covid}).
In addition, they can render a non-differentiable simulator differentiable for downstream tasks, such as amortized design optimization \autocite{ivanova2021implicit} or interact with MCMC samplers \autocite{fengler2021likelihood, boelts2022flexible}.

\section{Related Work}
\label{sec:rel_work}


\paragraph{Approximate Bayesian Computation} An established approach to SBI is embodied by approximate Bayesian computation \autocite[ABC;][]{marin2012approximate, sisson2018handbook}. 
ABC is a family of algorithms where the simplest one, ``ABC rejection'', generates draws from an approximate posterior by repeatedly proposing parameters from the prior distribution, and then simulating a corresponding synthetic data set by running the simulator with the proposed parameters.
More sophisticated ABC samplers are Sequential Monte Carlo \autocite[ABC-SMC;][]{beaumont2009adaptive,smc,del2012adaptive,picchini2022guided} and Markov chain Monte Carlo ABC \autocite[ABC-MCMC;][]{marjoram2003markov,picchini2014inference}. 
In ABC, raw data are typically reduced via summary functions.
However, \textit{hand-crafted} summary statistics are often insufficient, which results in a leak of information about the parameters \autocite{marin2018likelihood}. 
Recent work has used neural networks to learn informative summary statistics of model parameters in ABC \autocite{jiang2017learning, wiqvist2019partially,chen2020neural}. 

\paragraph{Synthetic Likelihoods and Particle MCMC} Despite being intuitive to grasp and use, the above ABC methods are notoriously inefficient, typically requiring millions of model simulations, which can be prohibitive for expensive simulators. 
Another established SBI alternative, also based on data-reduction via summary statistics, is \textit{synthetic likelihood} \autocite{wood2010statistical,price2018bayesian}, which is more suitable for high-dimensional summary statistics. 
Since synthetic likelihood is typically implemented in tandem with an MCMC sampler where multiple data sets are simulated at each proposed $\thetab$, it can also be computationally intensive. 
Particle MCMC \autocite{andrieu2010particle} is a simulation-based method for exact Bayesian inference which has found considerable success, especially for state-space models. 
However, particle MCMC could be infeasible when multiple inference runs are required to separately fit several different data sets.

\begin{figure}[t]
    \centering
    \begin{minipage}[c]{0.05\linewidth}
    \,
    \end{minipage}
    \begin{minipage}[c]{0.46\linewidth}
        \centering Posterior Calibration
    \end{minipage}
    \begin{minipage}[c]{0.46\linewidth}
        \centering \quad Joint Calibration
    \end{minipage}\\
    \begin{minipage}[c]{0.05\linewidth}
        \rotatebox{90}{Gaussian Mixture}
    \end{minipage}
    \begin{minipage}[c]{0.46\linewidth}
        \includegraphics[width=\linewidth]{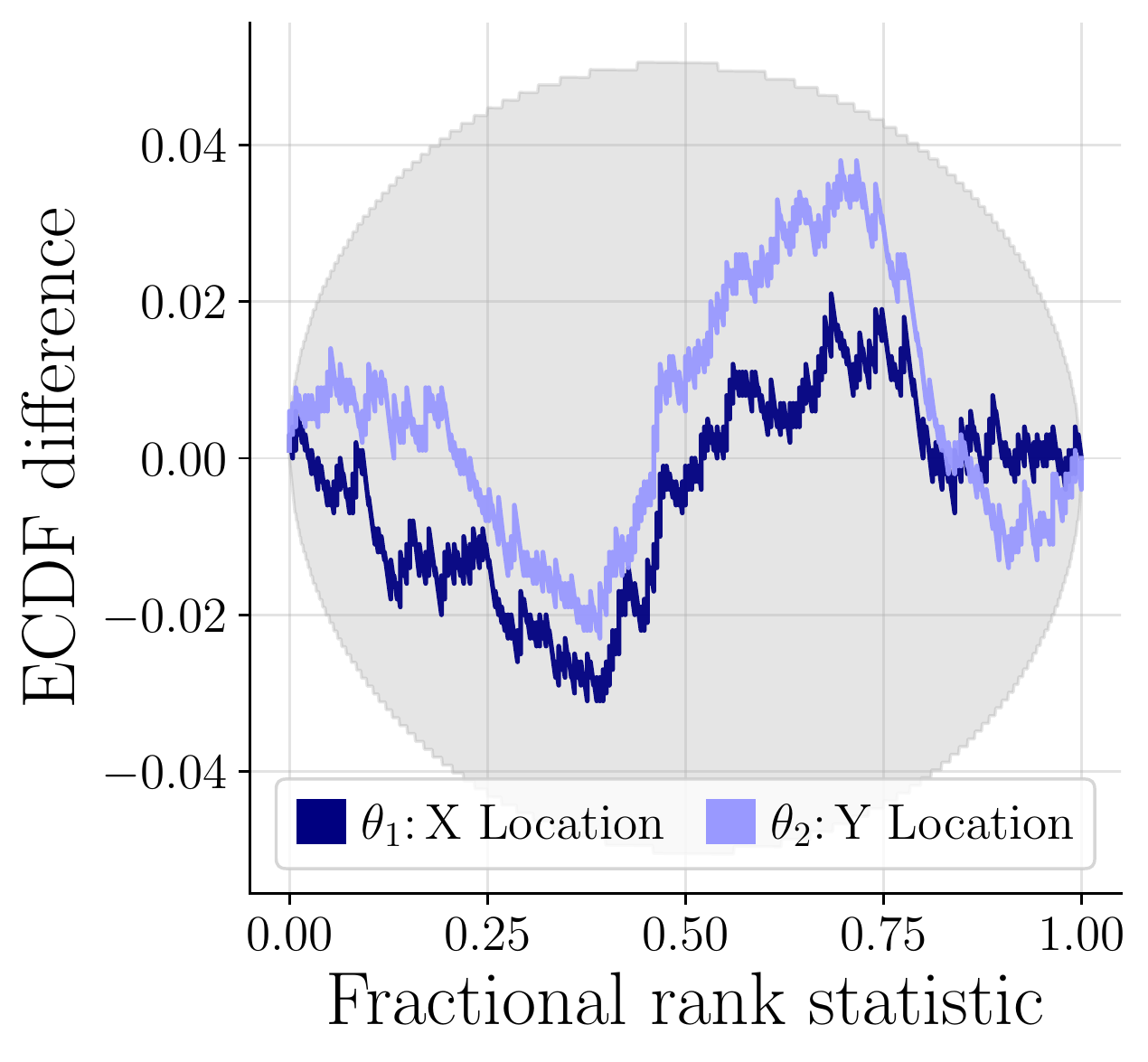}
    \end{minipage}
    \begin{minipage}[c]{0.46\linewidth}
        \includegraphics[width=\linewidth]{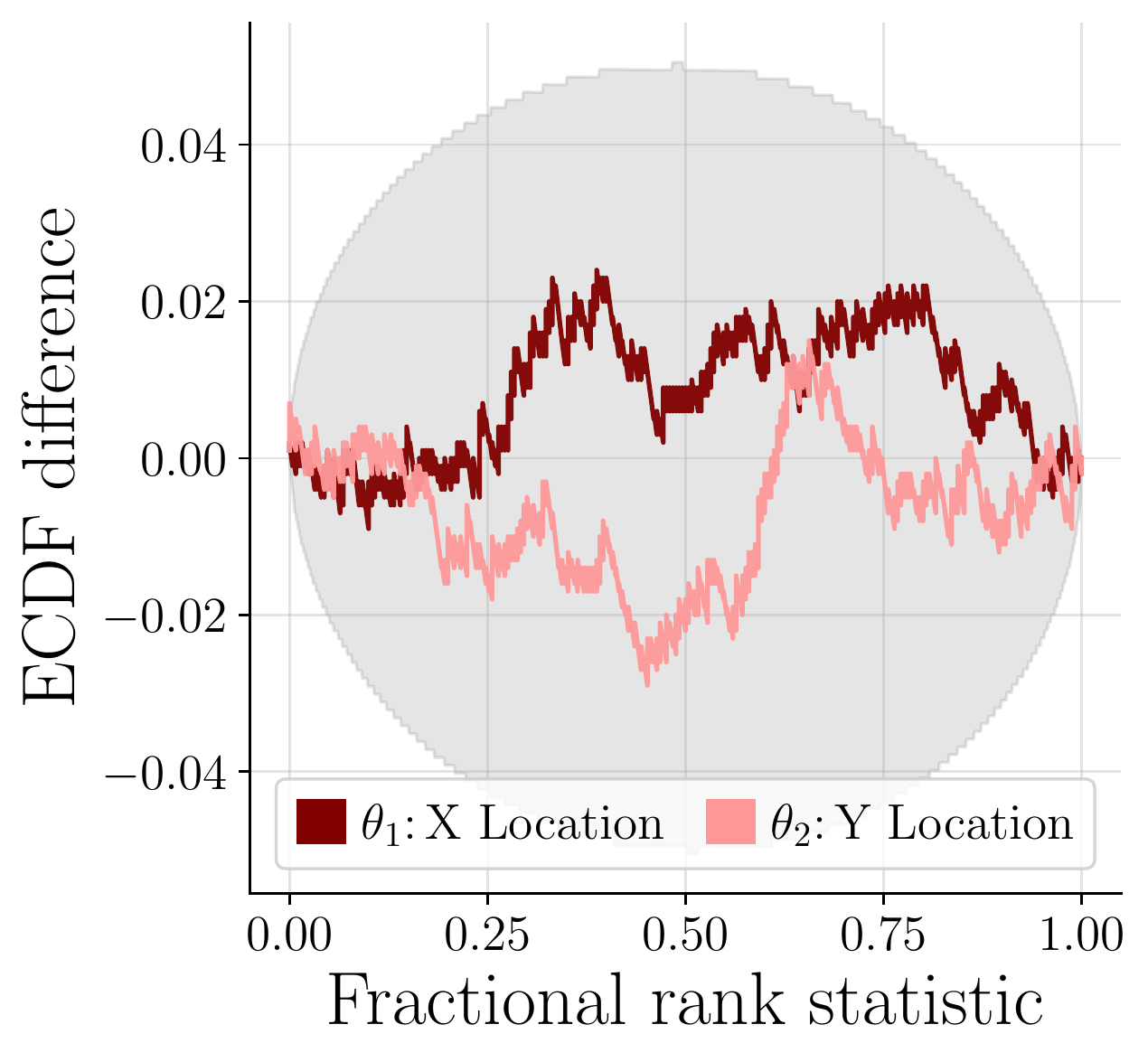}
    \end{minipage}\\
    \begin{minipage}[c]{0.05\linewidth}
        \rotatebox{90}{SIR}
    \end{minipage}
    \begin{minipage}[c]{0.46\linewidth}
        \includegraphics[width=\linewidth]{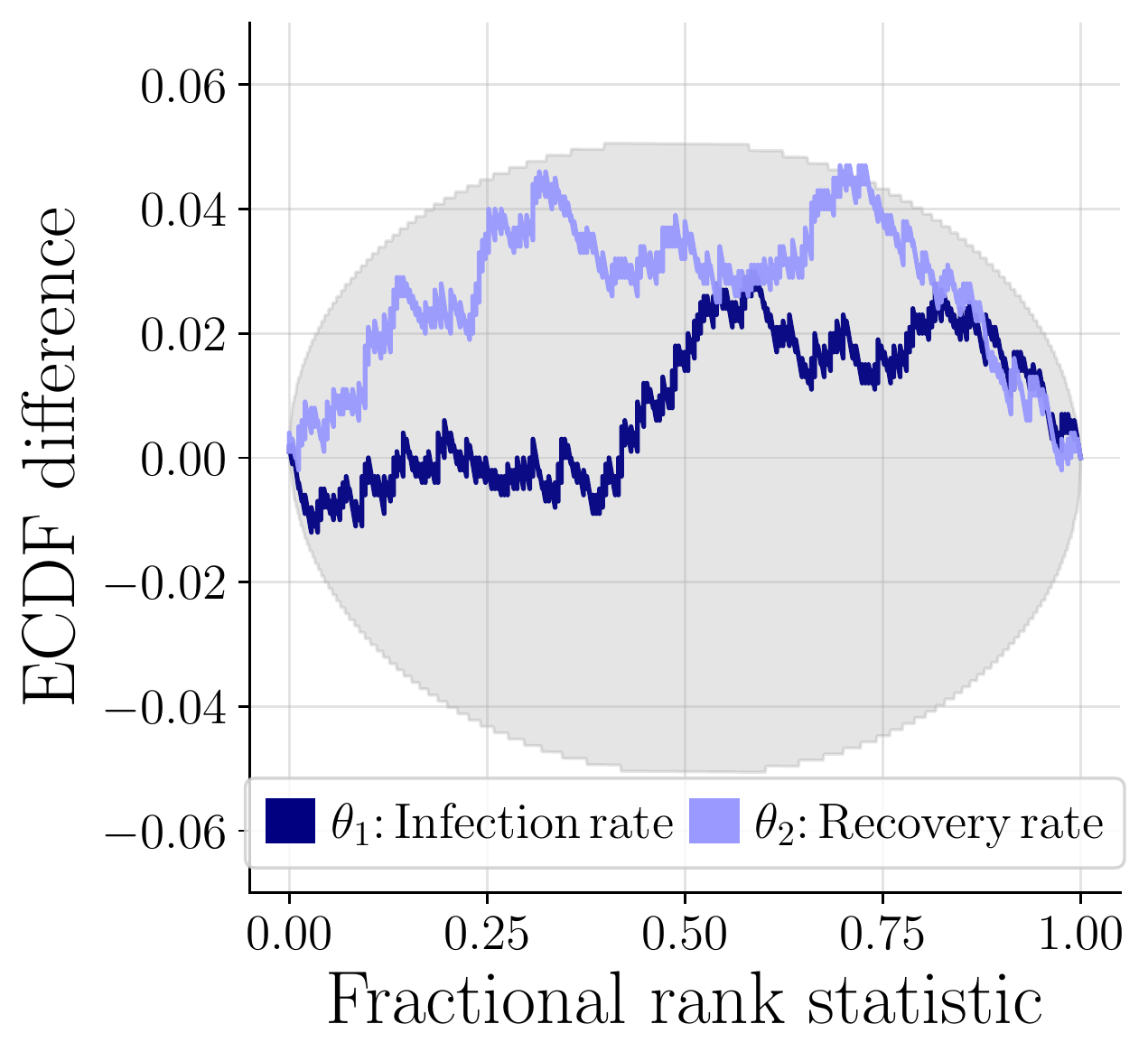}
    \end{minipage}
    \begin{minipage}[c]{0.46\linewidth}
        \includegraphics[width=\linewidth]{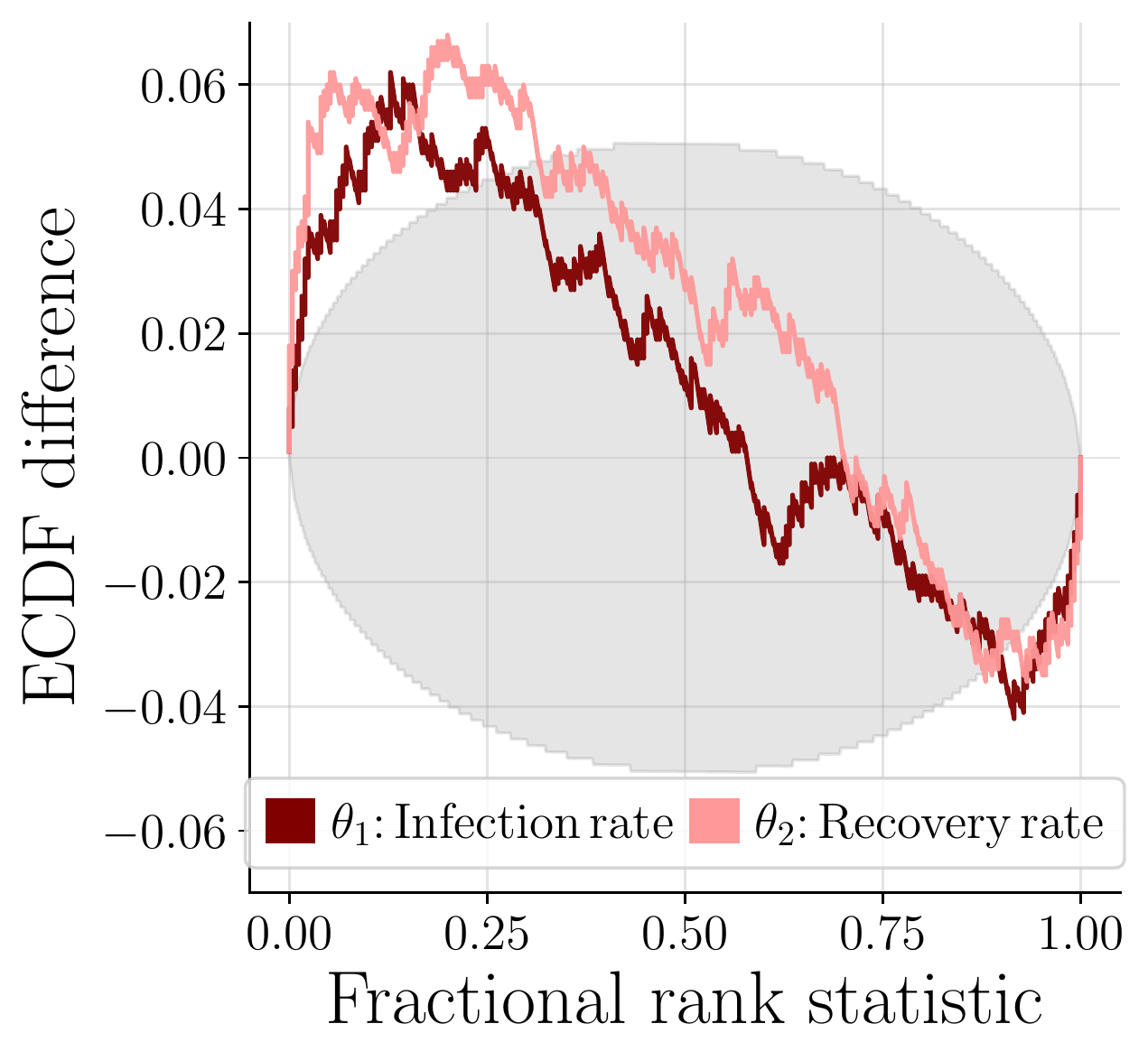}
    \end{minipage}
    \caption{\textbf{Experiment 1.} Example calibration tests for 2 of the more challenging benchmarks. \textit{Top row:} Good posterior and joint calibration of JANA for the Gaussian mixture model. \textit{Bottom row:} Posterior and joint calibration can be used in tandem to detect an underperforming likelihood network in the SIR model. The posterior network alone induces no systematic deviations when applied to simulator outputs (bottom left), but overestimates the parameters given the outputs of the surrogate network (bottom right).}
    \label{fig:benchmarks}
\end{figure}

\paragraph{Neural Posterior Estimation}

Methods for neural posterior estimation either specialize a neural approximator for inference on a single observation\footnote{The term \textit{observation} may refer to an entire data set, depending on how the data is used to update the posterior. \new{For instance, typical toy models (e.g., two moons) use a single data point, whereas realistic model applications typically use a set of data points for inference.}} \autocite{papamakarios2016fast, lueckmann2017flexible, greenberg2019automatic, durkan2020contrastive, deistler2022truncated}, or inference across arbitrarily many observations \autocite{ardizzone2018analyzing, gonccalves2020training, radev2020bayesflow, pacchiardi2022score, avecilla2022neural}.
The former methods perform \textit{sequential estimation} by iteratively refining the prior to generate simulations in the vicinity of the observation.
Thus, they are \textit{not amortized}, as each new observation necessitates a costly re-training of the neural approximator.
In contrast, the latter methods can perform \textit{amortized inference}, as the neural approximator is trained to generalize over the entire prior predictive distribution and can be queried for any observation assumed to arise from the Bayesian model.
Importantly, amortization can be performed over any aspect of the model, including data sets \autocite{gonccalves2020training} or other contextual factors, such as the number of observations in a data set or the number of time points in a time series \autocite{radev2020bayesflow}.

\paragraph{Neural Likelihood Estimation}
A related family of neural methods directly targets the intractable likelihood function instead of the posterior \autocite{papamakarios2019sequential, lueckmann2019likelihood, hermans2020likelihood, fengler2021likelihood, boelts2022flexible, munk2022probabilistic}.
The endpoint of these methods is an \textit{amortized likelihood approximator} which can mimic a complex simulator or be used in tandem with non-amortized MCMC samplers for posterior estimation.
The latter can be prohibitively time-consuming, since it not only requires expensive simulation-based training, but also integrating likelihood approximators into MCMC.
This makes validating the posteriors \autocite[e.g., via simulation-based-calibration; SBC;][]{talts2018validating, sailynoja2022graphical} challenging or even impossible in practice.
Nevertheless, likelihood approximators have certain advantages over posterior approximators, for instance, they do not need to be retrained for different priors and can emulate the behavior of large-scale simulators \autocite{lavin2021simulation}.

\paragraph{Neural Posterior and Likelihood Estimation}

\new{In a pioneering work, \textcite{wiqvist2021sequential} attempt to embody the best of both worlds by training together two networks for sequential neural posterior and likelihood approximation (SNPLA).
A potential shortcoming of SNPLA is that it optimizes the reverse Kullback-Leibler (rKL) divergence, which is prone to mode collapse and instabilities \autocite{arjovsky2017wasserstein}.
Sequential neural variational inference \autocite[SNVI;][]{glockler2022snvi} improves on SNPLA by targeting the forward KL (fKL) and using an importance sampling correction of the posterior estimates.
JANA also optimizes the mode-covering fKL by approximating an expectation over the Bayesian joint model (Eq.~\ref{eq:cost}).
In addition, JANA operates in a fully amortized manner, such that the posterior network can be applied to any set of observations (i.e., data sets; potentially with different sizes) and the likelihood network can produce instantaneous surrogate simulations given any parameter configuration.
This enables us to amortize some of the most costly procedures in Bayesian analysis, such as simulation-based calibration and leave-one-out cross-validation. 
In contrast, both SNPLA and SNVI focus on sequential (non-amortized) inference and employ a likelihood network only to support posterior estimation.
}





\section{Experiments}
\label{sec:experiment}

In the following, we will illustrate the utility of JANA in thirteen Bayesian models across five experiments.
For \textbf{Experiments 1--3}, we train the networks without the Maximum Mean Discrepancy (MMD) criterion in Eq.~\ref{eq:cost} (i.e., $\lambda = 0$), because our validations feature no model misspecification.
The code for running and reproducing all experiments is available at \url{https://github.com/bayesflow-org/JANA-Paper}.
JANA is implemented in the BayesFlow library.

\subsection{Ten Benchmark Experiments}
\begin{figure}[t]
    \centering
    \includegraphics[width=\linewidth]{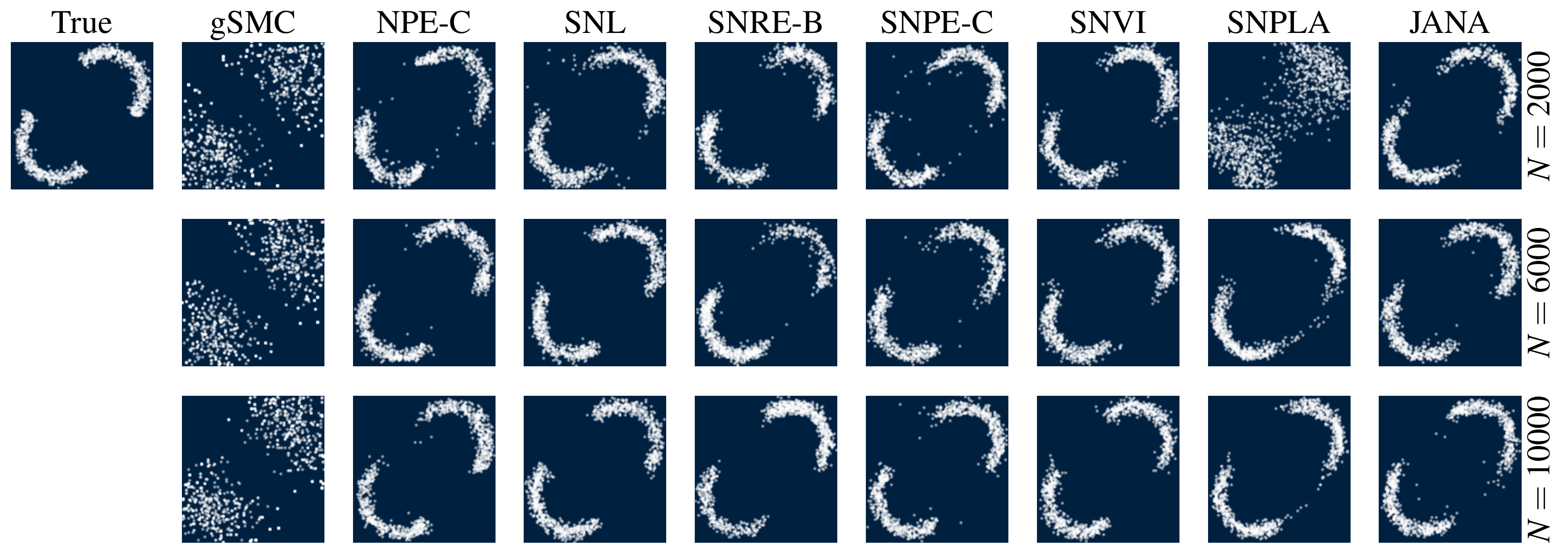} 
    \caption{\textbf{Experiment 2}. Samples from the approximate posterior distribution (Two Moons, repetition \#1). \new{No evident advantage of non-amortized over amortized neural methods (i.e., NPE-C and JANA) .}}
    \label{fig:two-moons:posterior}
\end{figure}

\paragraph{Setup} 
This experiment demonstrates the fidelity of our proposed architecture as well as the utility of our calibration checks to diagnose approximation faults on a set of ten benchmark simulation models proposed by \textcite{lueckmann2021benchmarking}.
Since these benchmarks were originally designed for gauging the performance of (non-amortized) posterior estimation, we deviate from the original problem setting by (i) approximating both posterior and likelihood; and (ii) validating our results on a much larger held-out set of $1\,000$ simulations (as compared to just 10).

For each benchmark, we train our networks with a fixed budget of $10\,000$ simulations, as we consider this to be a challenging practical setup with low-to-medium training data availability.
Importantly, our goal here is \textit{not to propose a better method for posterior estimation}, but to demonstrate the feasibility of joint amortization and the utility of our joint calibration diagnostic on a set of popular and rather diverse models.
See the \textbf{Appendix} and the accompanying code for more details and diagnostics.

\paragraph{Results} 
Overall, we observe stable training and good calibration across the ten benchmarks models, with the SIR model exhibiting systematic joint miscalibration due to likelihood approximations errors.
\autoref{fig:benchmarks} illustrates the utility of our calibration diagnostic to reveal both good calibration (i.e., ECDF trajectories completely contained in the confidence ellipsis for the Gaussian Mixture benchmark) as well as systematic deviations owing to the likelihood network (i.e., ECDF trajectories partially outside the confidence ellipsis for SIR).
Moreover, due to the inherent interpretability of the ECDF calibration plots, we can pinpoint the reasons for joint miscalibration of the SIR model:
The likelihood network tends to generate more rapid synthetic outbreaks than the actual model, which leads to the posterior network overestimating the parameters of surrogate simulations.
\begin{figure}[t]
	\centering
	\begin{subfigure}[t]{0.62\linewidth}
		\includegraphics[width=\linewidth]{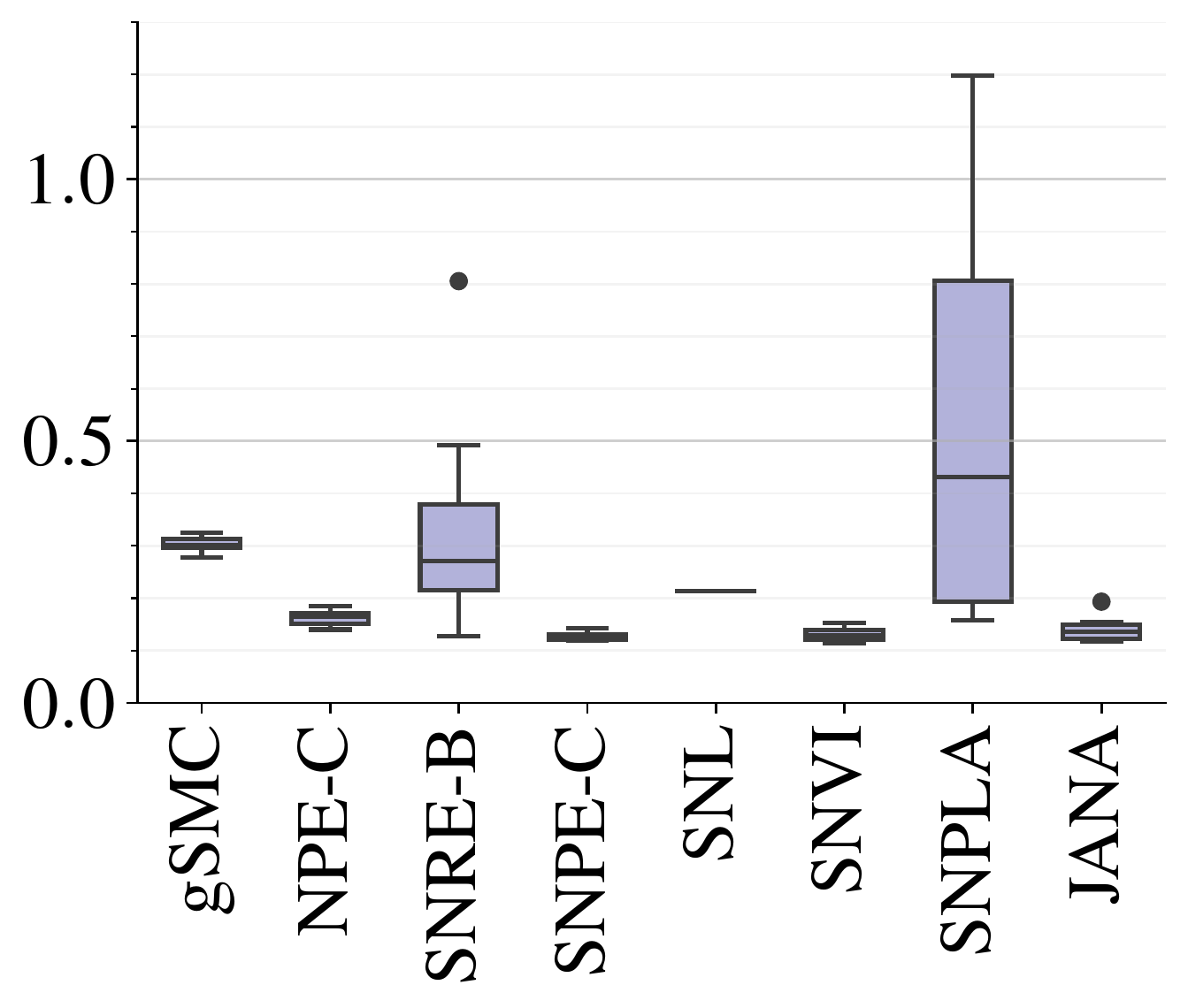}
		\caption{Posterior MMD}
		\label{fig:two-moons:boxplots:posterior}
	\end{subfigure}
	\hfill
	\begin{subfigure}[t]{0.37\linewidth}
		\includegraphics[width=\linewidth]{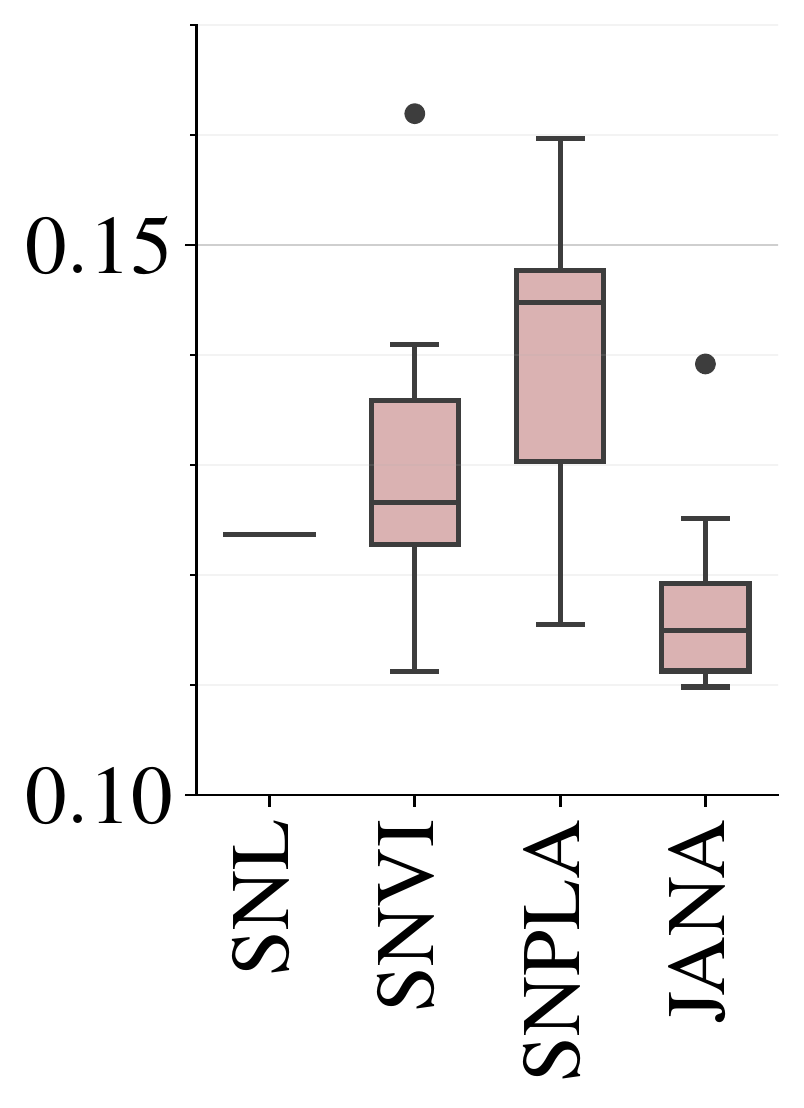}
		\caption{Joint MMD}
		\label{fig:two-moons:boxplots:posterior-predictive}
	\end{subfigure}
	\caption{\textbf{Experiment 2}. Performance with $N{=}10\,000$ training simulations, as indexed by the empirical Maximum Mean Discrepancy (MMD) estimate (lower is better).}
	\label{fig:two-moons:boxplots}
\end{figure}
\begin{figure*}[t]
	\centering
	\begin{subfigure}[t]{0.22\linewidth}
		\includegraphics[width=\linewidth]{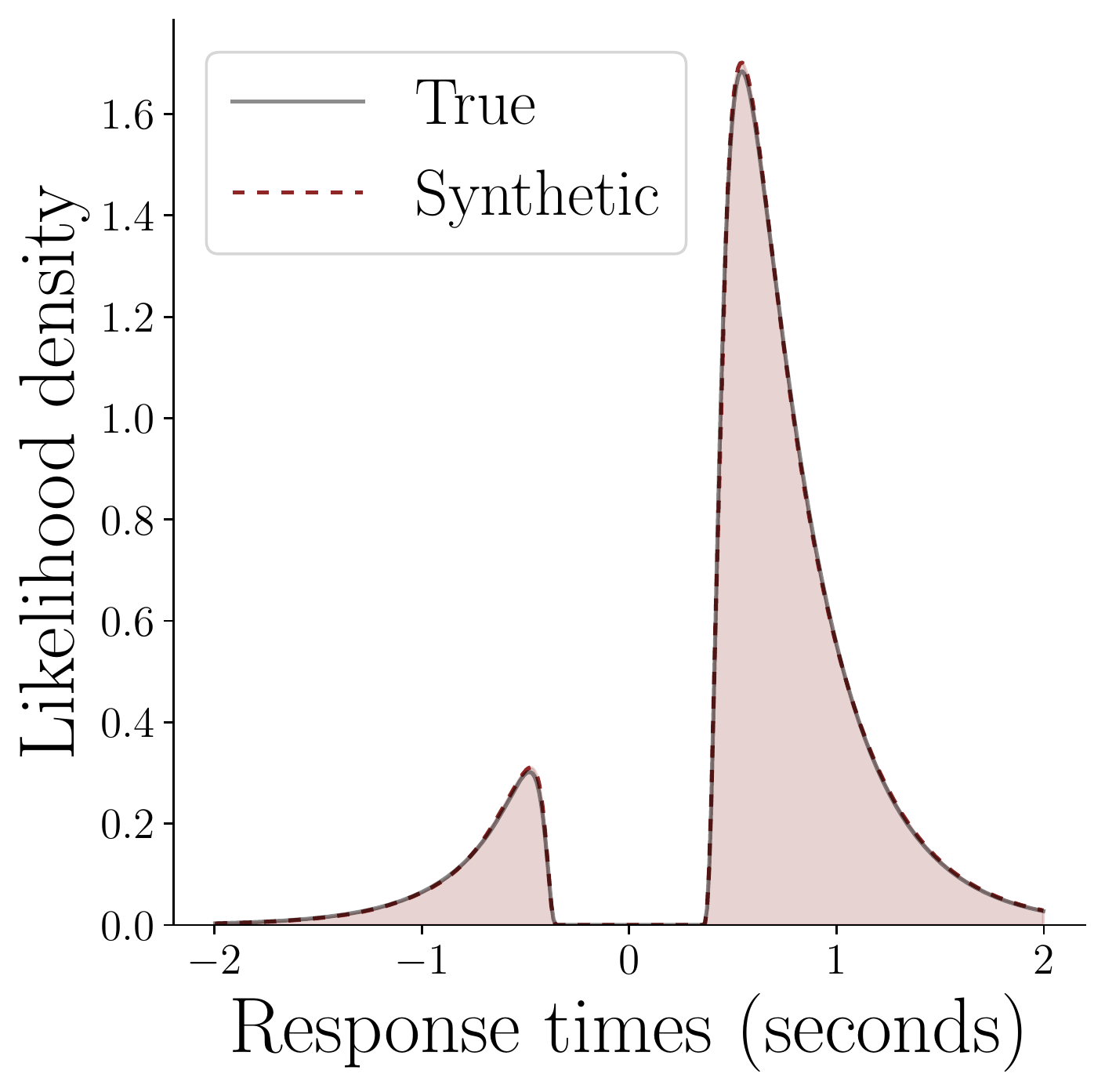}
		\caption{Likelihood emulation}
		\label{fig:diffusion-model:likelihood}
	\end{subfigure}
	\hspace*{0.25cm}
	\begin{subfigure}[t]{0.22\linewidth}
		\includegraphics[width=\linewidth]{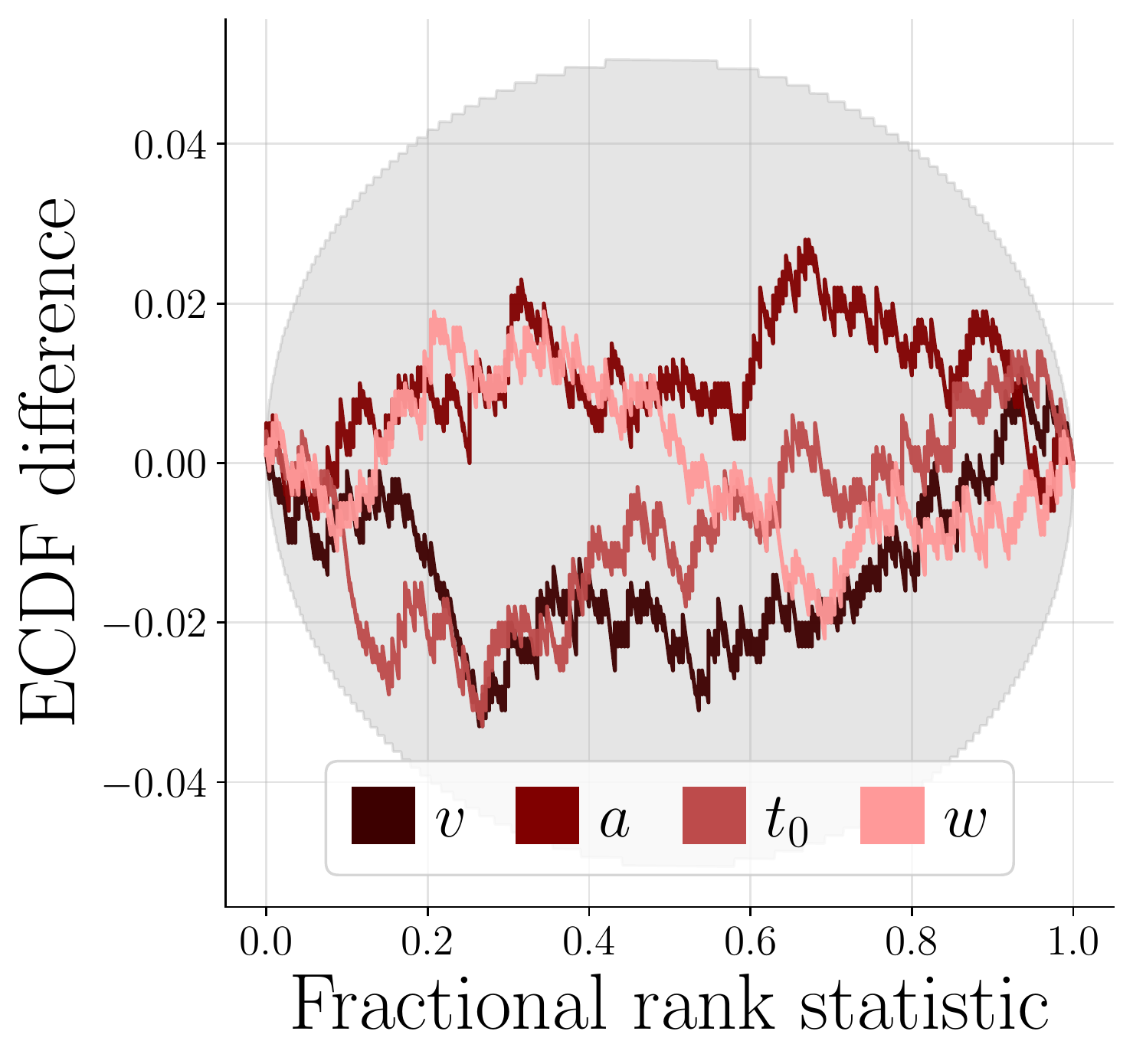}
		\caption{Joint calibration}
		\label{fig:diffusion-model:calibration}
	\end{subfigure}
	\hspace*{0.25cm}
	\begin{subfigure}[t]{0.22\linewidth}
		\includegraphics[width=\linewidth]{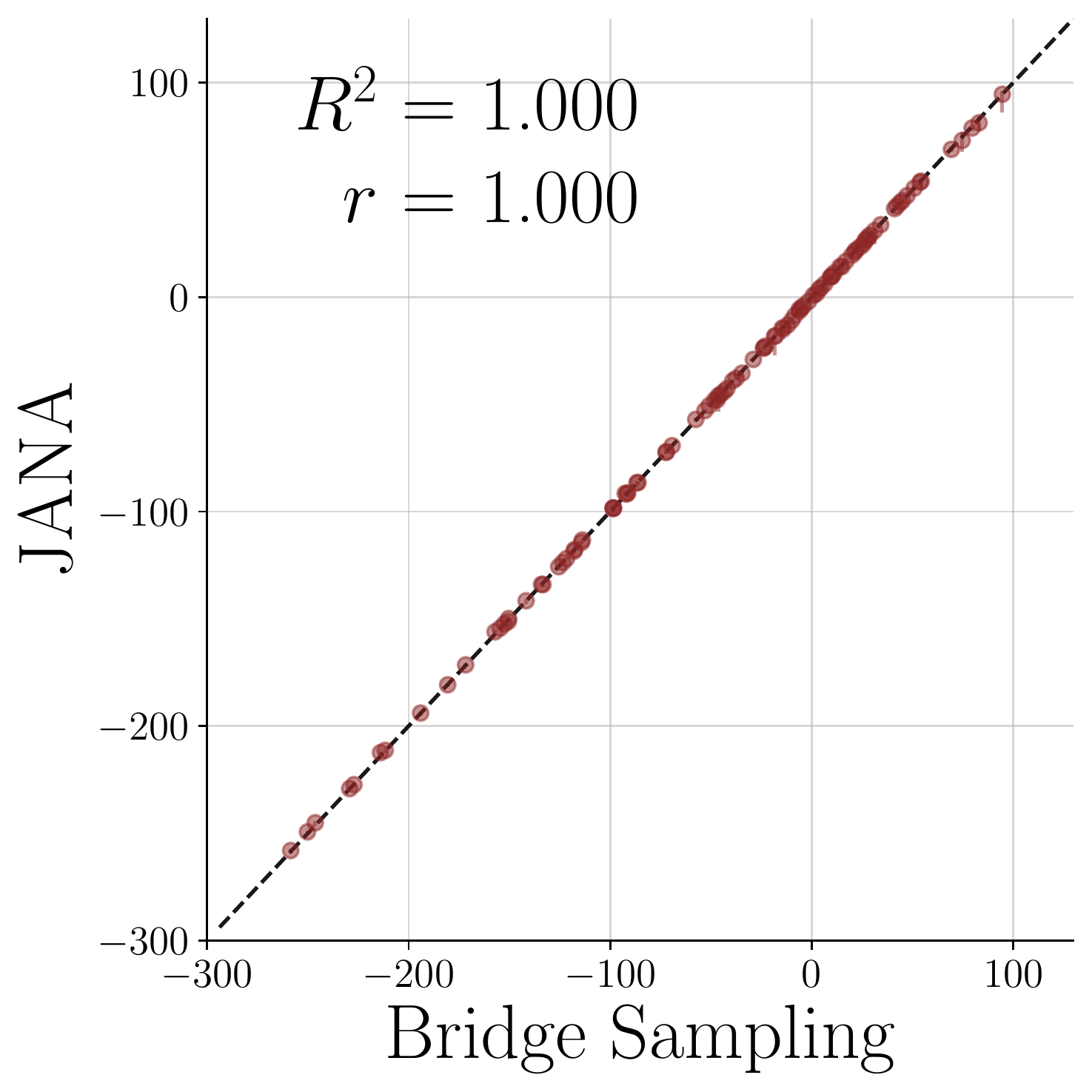}
		\caption{Prior predictive (LML)}
		\label{fig:diffusion-model:lml}
	\end{subfigure}
	\hspace*{0.25cm}
	\begin{subfigure}[t]{0.22\linewidth}
		\includegraphics[width=\linewidth]{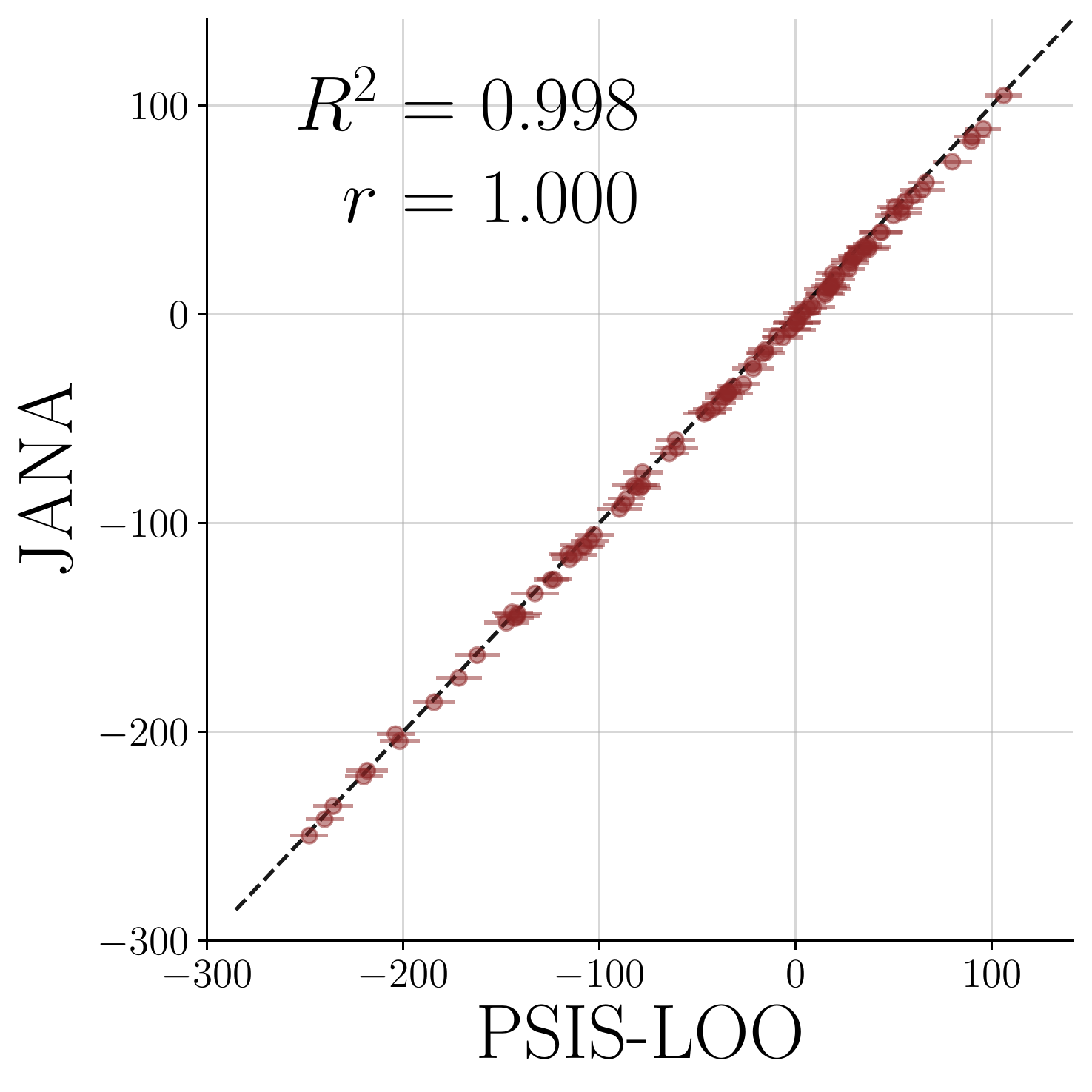}
		\caption{Post. predictive (ELPD)}
		\label{fig:diffusion-model:loo}
	\end{subfigure}
	\caption{\textbf{Experiment 3.}
		The true and synthetic likelihood align almost perfectly (\subref{fig:diffusion-model:likelihood}).
		The joint approximation of all parameters is well calibrated (\subref{fig:diffusion-model:calibration}).
		Both prior predictive (\subref{fig:diffusion-model:lml}) and posterior predictive (\subref{fig:diffusion-model:loo}) estimates of JANA closely approximate those obtained via gold standard bridge sampling and Pareto smoothed importance sampling (PSIS). Each point in (\subref{fig:diffusion-model:lml}) and (\subref{fig:diffusion-model:loo}) represents one out of 100 held-out simulations.}
\end{figure*}
\subsection{Two Moons: Method Comparison}
\paragraph{Setup}
Here, we focus specifically on the Two Moons benchmark \autocite{greenberg2019automatic, lueckmann2021benchmarking} and use the code from \textcite{wiqvist2021sequential} to compare JANA with the popular sequential methods SNL \autocite{papamakarios2019sequential}, SNPE-C \autocite{greenberg2019automatic}, SNRE-B \autocite{durkan2020contrastive}, SNPLA \autocite{wiqvist2021sequential}, SNVI \autocite{glockler2022snvi}, and a recent ABC-SMC algorithm with ``guided particles'' \autocite[here abbreviated with g-SMC, which is the method called ``hybrid'' in ][]{picchini2022guided}.
The model is characterized by a bimodal posterior with two separated crescent moons for the observed point $\x_{\text{new}} = (0, 0)^\top$ which a posterior approximator needs to recover.
We train SNL, SNPE-C, SNRE-B, SNVI, SNPLA, g-SMC, and JANA following the same setup from \textcite{wiqvist2021sequential}.\footnote{For comparability with \textcite{wiqvist2021sequential}, the setup differs from \textcite{lueckmann2021benchmarking} in terms of location and size of the moons. The results of \textbf{Experiment 2} with the implementation of \textcite{lueckmann2021benchmarking} are comparable, see the \textbf{Appendix}.}
For each method, we repeat the experiment ten times using a fixed budget of $2\,000$, $6\,000$, and $10\,000$ simulations and subsequently obtain $1\,000$ posterior draws from the converged methods.
For a numerical evaluation, we apply MMD between the approximate and analytical distributions.

\paragraph{Results}
JANA consistently explores both crescent moons throughout all repetitions and already captures the local patterns of the posterior after $2\,000$ training samples (see \autoref{fig:two-moons:posterior}).
\new{With respect to posterior performance, JANA is on par with all sequential methods which are tailored to one observed data set (see \autoref{fig:two-moons:boxplots:posterior}).
In terms of joint (posterior predictive) performance, JANA outperforms non-amortized sequential methods, see \autoref{fig:two-moons:boxplots:posterior-predictive}.}
In light of these and previous results, \textit{amortization across data sets} seems to be a reasonable choice even with limited simulation budgets, especially since sequential (non-amortized) methods may be infeasible for large data \autocite{hermans2021}.
\new{The \textbf{Appendix} contains wall-clock times and further details for training and inference.}



\subsection{Exchangeable Diffusion Model}
\paragraph{Setup} 
This example demonstrates amortized log marginal likelihood (LML) and expected log predictive density (ELPD) estimation based on a mechanistic model of decision making: the diffusion model \autocite{ratcliff2008}.
We benchmark our results against state-of-the-art likelihood-based methods.
First, we compare our marginal likelihood estimates with those obtained with bridge sampling \autocite{gronau2017bridgesampling}.
Second, we compare our leave-one-out (LOO)-ELPD estimates (Eq.~\ref{eq:elpd}) with those obtained using Pareto smoothed importance sampling \autocite{vehtari2017practical}.
Both methods use random draws obtained via MCMC, as implemented in Stan \autocite{carpenter2017stan}.


\paragraph{Results}
Our results indicate well-calibrated joint approximation (see~\autoref{fig:diffusion-model:calibration}) as well as accurate posterior and likelihood estimation (see~\autoref{fig:diffusion-model:lml} and \ref{fig:diffusion-model:loo}).
For the approximation of marginal likelihoods, we first perform amortized posterior sampling on the 100 held-out data sets. 
We then evaluate the approximate likelihood on these samples, and finally apply Eq.~\ref{eq:log_marg} to compute the LML.
Our numerical results reveal a very close correspondence between our neural log marginal likelihoods and those obtained via MCMC-based bridge sampling (see \autoref{fig:diffusion-model:lml}). 
Furthermore, our amortized LOO-CV estimates align very closely with the estimates obtained via PSIS-LOO (see \autoref{fig:diffusion-model:loo}).

\paragraph{MCMC Integration}
Surrogate likelihoods provide all information that is needed for MCMC sampling. 
We provide an interface to PyMC \autocite{salvatier_probabilistic_2016} to allow for easy model building and use of existing samplers. 
Note, that the performance of gradient-based samplers, such as Hamiltonian Monte Carlo, critically depends on the precision of partial log-likelihood derivatives. 
Using PyMC's No-U-Turn sampler (NUTS) with our neural likelihood, we obtained results similar to those using Stan. 
If gradient-based sampling methods fail, we advise to use gradient-free sampling methods, such as slice sampling. 
For detailed information, see the \textbf{Appendix}.

\begin{figure*}[h]
    \centering
    \includegraphics[width=0.99\textwidth]{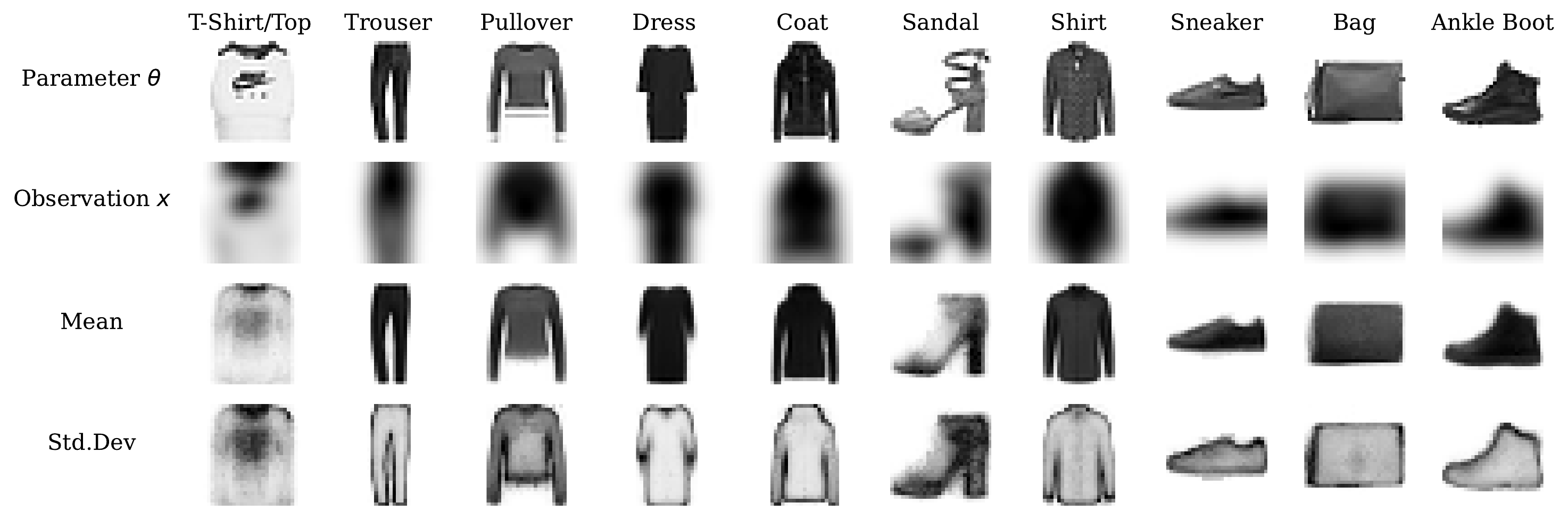} 
    \caption{\textbf{Experiment 5}. Example denoising results from each class of Fashion MNIST. \textit{First row:} Original image acting as the ``parameters'' of the noisy camera simulator. \textit{Second row:} Blurred image, acting as the output of the camera simulator. \textit{Third and fourth row:} Means and standard deviations of the posteriors estimated from the corresponding blurry ``observations''. \textit{Note:} For standard deviations, darker regions indicate larger variability in the outputs.}
    \label{fig:denoising}
\end{figure*}

\subsection{Markovian Compartmental Model}

\paragraph{Setup} This experiment demonstrates surrogate simulations of a complex non-exchangeable model of infectious diseases.
The model features 34 parameters and thus repre\-sents a considerable extension of the two-parameter toy SIR model \autocite{lueckmann2021benchmarking, radev2020bayesflow}.
We use the model specification and posterior network from \textcite{radev2021outbreakflow}.
We implement the likelihood network as a recurrent cINN (see Section~\ref{sec:lik_net}) to test its ability to emulate raw and noisy time series.
Further, we train the summary network with the MMD criterion (Eq.~\ref{eq:cost}) with $\lambda = 1$ to judge the quality of the surrogate simulations numerically.

\paragraph{Results} Upon convergence, we use the likelihood network to generate synthetic outbreak trajectories and compare them visually with the outputs of the original simulator.
We observe good emulation across a variety of different parameter configurations, each leading to a qualitatively different simulated scenario (see~\autoref{fig:covid} for an example and the \textbf{Appendix} for detailed results).
Moreover, it seems that the surrogate network is not only able to accurately approximate the median trajectory, but also the variability (i.e., \textit{aleatoric uncertainty}) in simulated trajectories.
 
Beyond purely visual comparisons, we also compute the posterior and joint calibration of the two networks using joint SBC on $1\,000$ held-out simulations.
We confirm the good posterior calibration observed by Radev et al.\ (\citeyear{radev2021outbreakflow}).
In addition, the joint calibration results help us highlight some subtle deficiencies of the likelihood network.
For instance, it tends to overestimate the variability of simulated time series, thus ``tricking'' the posterior network into estimating higher values for the noise parameters (see \textbf{Appendix}).
We attribute this deficiency to the extremely wide magnitude range of the simulated data (incidence in the order of millions) which is not captured by our simple input standardization procedure.

\subsection{High-Dimensional Bayesian Denoising}
\paragraph{Setup} The last experiment demonstrates the feasibility of JANA for tackling high-dimensional Bayesian models with relatively low simulation budgets. Similarly to \textcite{ramesh2022gatsbi}, we consider a Bayesian denoising setup on the Fashion MNIST data set, where the ``parameter vector'' $\thetab \in \mathbb{R}^{784}$ represents the original image and the ``observation'' $\x \in \mathbb{R}^{784}$ is a blurry version of the image generated by a simulated noisy camera. 

We train a JANA architecture comprising two fully connected affine coupling architectures operating on the flattened images (as they would, if the Bayesian model were a scientific simulator with $784$ parameters). 
Since both ``parameters'' and ``data'' in this unusual example are images, we use two simple convolutional networks as summary networks for both the posterior and likelihood networks. 

\paragraph{Results} We evaluate the performance of the networks on the official Fashion MNIST test set. 
To summarize their calibration, we report the average expected calibration error \autocite{radev2020bayesflow} for the posterior ($\approx 0.03 \pm 0.02$) and joint samples ($\approx 0.04 \pm 0.03$), indicating reasonable approximation fidelity and slightly increased joint miscalibration.
We also inspect the visual quality of random samples generated from the posterior and the synthetic likelihood (see~\autoref{fig:denoising} for an example of posterior estimation). 
These results suggest that the networks have captured the basic structure of the problem, with ``core features'' being easier to reconstruct than ``details''.
An extended description and more results are provided in the \textbf{Appendix}.

\section{Conclusion}
\label{sec:conclusion}

We investigated the utility of JANA for Bayesian surrogate modeling and simulation-based inference within the BayesFlow framework.
We believe that JANA can greatly enrich applications of amortized Bayesian inference.
Future work should investigate weight sharing schemes for the various network components and advance a framework-independent benchmark database for joint estimation of non-trivial scientific models.

\begin{acknowledgements}
    We thank Samuel Wiqvist for the fruitful discussions and his help with running the SNPLA experiments.
    We thank Manuel Gloeckler for the forthcoming assistance with the SNVI implementation of Experiment 2.
    We thank the reviewers for their thought-provoking feedback which has enabled us to improve the manuscript considerably.
    STR was supported by the Deutsche Forschungsgemeinschaft (DFG, German Research Foundation) under Germany’s Excellence Strategy -– EXC-2181 - 390900948 (the Heidelberg Cluster of Excellence STRUCTURES) and Google Cloud through the Academic Research Grants program.
    MS was supported by the Cyber Valley Research Fund (grant number: CyVy-RF-2021-16).
    MS and PCB were supported by the Deutsche Forschungsgemeinschaft (DFG, German Research Foundation) under Germany’s Excellence Strategy -- EXC-2075 - 390740016 (the Stuttgart Cluster of Excellence SimTech).
    VP was supported by the state of Baden-Württemberg through bwHPC and the German Research Foundation (DFG) through grant INST 35/1597-1 FUGG. 
    UP was supported by the Swedish National Research Council (Vetenskapsrådet 2019-03924) and the Chalmers AI Research Centre.
    UK was supported by the the Informatics for Life initiative funded by the Klaus Tschira Foundation.
    The authors gratefully acknowledge the support and funding.
\end{acknowledgements}

\subsubsection*{References}
\printbibliography[heading=none]

@inproceedings{greenberg2019automatic,
  title={Automatic posterior transformation for likelihood-free inference},
  author={Greenberg, David and Nonnenmacher, Marcel and Macke, Jakob},
  booktitle={International Conference on Machine Learning},
  cpages={2404--2414},
  year={2019},
  corganization={PMLR}
}

@book{mackay2003information,
  title={Information theory, inference and learning algorithms},
  author={MacKay, David},
  year={2003},
  publisher={Cambridge University Press}
}

@article{vehtari2017practical,
  title={Practical Bayesian model evaluation using leave-one-out cross-validation and WAIC},
  author={Vehtari, Aki and Gelman, Andrew and Gabry, Jonah},
  journal={Statistics and Computing},
  cvolume={27},
  cnumber={5},
  cpages={1413--1432},
  year={2017},
  publisher={Springer}
}

@article{burkner_nfloo_2021,
	title = {Efficient leave-one-out cross-validation for {Bayesian} non-factorized normal and {Student}-t models},
	volume = {36},
	number = {2},
	journal = {Computational Statistics},
	author = {Bürkner, Paul-Christian and Gabry, Jonah and Vehtari, Aki},
	year = {2021},
	note = {Publisher: Springer},
	pages = {1243--1261},
}

@article{radev2020bayesflow,
  title={BayesFlow: Learning complex stochastic models with invertible neural networks},
  author={Radev, Stefan T and Mertens, Ulf K and Voss, Andreas and Ardizzone, Lynton and K{\"o}the, Ullrich},
  journal={IEEE transactions on neural networks and learning systems},
  year={2020},
  publisher={IEEE}
}

@article{lavin2021simulation,
  title={Simulation intelligence: Towards a new generation of scientific methods},
  author={Lavin, Alexander and Zenil, Hector and Paige, Brooks and others},
  cauthor={Lavin, Alexander and Zenil, Hector and Paige, Brooks and Krakauer, David and Gottschlich, Justin and Mattson, Tim and Anandkumar, Anima and Choudry, Sanjay and Rocki, Kamil and Baydin, At{\i}l{\i}m G{\"u}ne{\c{s}} and others},
  journal={arXiv preprint},
  year={2021}
}

@article{von2022mental,
  title={Mental speed is high until age 60 as revealed by analysis of over a million participants},
  author={von Krause, Mischa and Radev, Stefan T and Voss, Andreas},
  journal={Nature Human Behaviour},
  volume={6},
  number={5},
  pages={700--708},
  year={2022},
  publisher={Nature Publishing Group}
}

@article{ivanova2021implicit,
  title={Implicit deep adaptive design: policy-based experimental design without likelihoods},
  author={Ivanova, Desi R and Foster, Adam and Kleinegesse, Steven and Gutmann, Michael U and Rainforth, Thomas},
  journal={Advances in Neural Information Processing Systems},
  volume={34},
  pages={25785--25798},
  year={2021}
}

@article{radev2021amortized,
  title={Amortized Bayesian model comparison with evidential deep learning},
  author={Radev, Stefan T and D'Alessandro, Marco and Mertens, Ulf K and Voss, Andreas and K{\"o}the, Ullrich and B{\"u}rkner, Paul-Christian},
  journal={IEEE Transactions on Neural Networks and Learning Systems},
  year={2021},
  publisher={IEEE}
}

@article{ardizzone2019guided,
  title={Guided image generation with conditional invertible neural networks},
  author={Ardizzone, Lynton and L{\"u}th, Carsten and Kruse, Jakob and Rother, Carsten and K{\"o}the, Ullrich},
  journal={arXiv preprint},
  year={2019}
}

@article{carpenter2017stan,
  title={Stan: A probabilistic programming language},
  author={Carpenter, Bob and Gelman, Andrew and Hoffman, Matthew D and Lee, Daniel and Goodrich, Ben and Betancourt, Michael and Brubaker, Marcus and Guo, Jiqiang and Li, Peter and Riddell, Allen},
  journal={Journal of statistical software},
  volume={76},
  number={1},
  year={2017},
  publisher={Columbia Univ., New York, NY (United States); Harvard Univ., Cambridge, MA~…}
}

@article{pacchiardi2022score,
  title={Score Matched Neural Exponential Families for Likelihood-Free Inference.},
  author={Pacchiardi, Lorenzo and Dutta, Ritabrata},
  journal={J. Mach. Learn. Res.},
  volume={23},
  pages={38--1},
  year={2022}
}

@article{hermans2021,
Author = {Joeri Hermans and Arnaud Delaunoy and François Rozet and Antoine Wehenkel and Volodimir Begy and Gilles Louppe},
Title = {A Trust Crisis In Simulation-Based Inference? {Y}our Posterior Approximations Can Be Unfaithful},
Year = {2021},
journal = {arXiv preprint},
}

@article{gelman2020bayesian,
  title={Bayesian workflow},
  author={Gelman, Andrew and Vehtari, Aki and Simpson, Daniel and others},
  cauthor={Gelman, Andrew and Vehtari, Aki and Simpson, Daniel and Margossian, Charles C and Carpenter, Bob and Yao, Yuling and Kennedy, Lauren and Gabry, Jonah and B{\"u}rkner, Paul-Christian and Modr{\'a}k, Martin},
  journal={arXiv preprint},
  year={2020}
}

@article{schmitt2022bayesflow,
  title={Detecting Model Misspecification in Amortized Bayesian Inference with Neural Networks},
  author={Schmitt, Marvin and B{\"u}rkner, Paul-Christian and K{\"o}the, Ullrich and Radev, Stefan T},
  journal={arXiv preprint},
  year={2021}
}

@article{alexanderson2020robust,
  title={Robust model training and generalisation with Studentising flows},
  author={Alexanderson, Simon and Henter, Gustav Eje},
  journal={arXiv preprint arXiv:2006.06599},
  year={2020}
}

@article{draxler2022whitening,
  title={Whitening Convergence Rate of Coupling-based Normalizing Flows},
  author={Draxler, Felix and Schn{\"o}rr, Christoph and K{\"o}the, Ullrich},
  journal={arXiv preprint arXiv:2210.14032},
  year={2022}
}

@article{ramesh2022gatsbi,
  title={GATSBI: Generative adversarial training for simulation-based inference},
  author={Ramesh, Poornima and Lueckmann, Jan-Matthis and Boelts, Jan and Tejero-Cantero, {\'A}lvaro and Greenberg, David S and Gon{\c{c}}alves, Pedro J and Macke, Jakob H},
  journal={arXiv preprint arXiv:2203.06481},
  year={2022}
}

@inproceedings{munk2022probabilistic,
  title={Probabilistic surrogate networks for simulators with unbounded randomness},
  author={Munk, Andreas and Zwartsenberg, Berend and {\'S}cibior, Adam and Baydin, At{\i}l{\i}m G{\"u}ne{\c{s}} G and Stewart, Andrew and Fernlund, Goran and Poursartip, Anoush and Wood, Frank},
  booktitle={Uncertainty in Artificial Intelligence},
  pages={1423--1433},
  year={2022},
  organization={PMLR}
}

@article{avecilla2022neural,
  title={Neural networks enable efficient and accurate simulation-based inference of evolutionary parameters from adaptation dynamics},
  author={Avecilla, Grace and Chuong, Julie N and Li, Fangfei and Sherlock, Gavin and Gresham, David and Ram, Yoav},
  journal={PLoS Biology},
  volume={20},
  number={5},
  pages={e3001633},
  year={2022},
  publisher={Public Library of Science San Francisco, CA USA}
}

@article{gonccalves2020training,
  title={Training deep neural density estimators to identify mechanistic models of neural dynamics},
  author={Gon{\c{c}}alves, Pedro J and Lueckmann, Jan-Matthis and Deistler, Michael and others},
  cauthor={Gon{\c{c}}alves, Pedro J and Lueckmann, Jan-Matthis and Deistler, Michael and Nonnenmacher, Marcel and {\"O}cal, Kaan and Bassetto, Giacomo and Chintaluri, Chaitanya and Podlaski, William F and Haddad, Sara A and Vogels, Tim P and others},
  journal={Elife},
  cvolume={9},
  cpages={e56261},
  year={2020},
  publisher={eLife Sciences Publications Limited}
}

@article{radev2021outbreakflow,
  title={OutbreakFlow: Model-based Bayesian inference of disease outbreak dynamics with invertible neural networks and its application to the COVID-19 pandemics in Germany},
  author={Radev, Stefan T and Graw, Frederik and Chen, Simiao and Mutters, Nico T and Eichel, Vanessa M and B{\"a}rnighausen, Till and K{\"o}the, Ullrich},
  journal={PLoS computational biology},
  cvolume={17},
  cnumber={10},
  cpages={e1009472},
  year={2021},
  publisher={Public Library of Science San Francisco, CA USA}
}

@inproceedings{hermans2020likelihood,
  title={Likelihood-free mcmc with amortized approximate ratio estimators},
  author={Hermans, Joeri and Begy, Volodimir and Louppe, Gilles},
  booktitle={International Conference on Machine Learning},
  pages={4239--4248},
  year={2020},
  organization={PMLR}
}

@article{gronau2017bridgesampling,
  title={bridgesampling: An R package for estimating normalizing constants},
  author={Gronau, Quentin F and Singmann, Henrik and Wagenmakers, Eric-Jan},
  journal={arXiv preprint},
  year={2017}
}

@article{meng_simulating_1996,
	title = {Simulating ratios of normalizing constants via a simple identity: a theoretical exploration},
	cvolume = {6},
	cnumber = {4},
	journal = {Statistica Sinica},
	author = {Meng, Xiao-Li and Wong, Wing Hung},
	year = {1996},
	cpages = {831--860},
}

@article{fengler2021likelihood,
  title={Likelihood approximation networks (LANs) for fast inference of simulation models in cognitive neuroscience},
  author={Fengler, Alexander and Govindarajan, Lakshmi N and Chen, Tony and Frank, Michael J},
  journal={Elife},
  volume={10},
  pages={e65074},
  year={2021},
  publisher={eLife Sciences Publications Limited}
}

@article{vehtari_survey_2012,
	title = {A survey of {Bayesian} predictive methods for model assessment, selection and comparison},
	cvolume = {6},
	journal = {Statistics Surveys},
	author = {Vehtari, Aki and Ojanen, Janne},
	year = {2012},
	cpages = {142--228},
}

@article{dinh2016density,
  title={Density estimation using real nvp},
  author={Dinh, Laurent and Sohl-Dickstein, Jascha and Bengio, Samy},
  journal={arXiv preprint arXiv:1605.08803},
  year={2016}
}

@article{Gretton2012,
    author = {Gretton, A and Borgwardt, K. and Rasch, Malte and Schölkopf, Bernhard and Smola, AJ},
    year = {2012},
    pages = {723-773},
    title = {{A Kernel Two-Sample Test}},
    volume = {13},
    journal = {The Journal of Machine Learning Research}
}

@article{bieringer2021measuring,
  title={Measuring QCD splittings with invertible networks},
  author={Bieringer, Sebastian and Butter, Anja and Heimel, Theo and H{\"o}che, Stefan and K{\"o}the, Ullrich and Plehn, Tilman and Radev, Stefan T},
  journal={SciPost Physics},
  volume={10},
  number={6},
  pages={126},
  year={2021}
}

@article{kingma2018glow,
  title={Glow: Generative flow with invertible 1x1 convolutions},
  author={Kingma, Durk P and Dhariwal, Prafulla},
  journal={Advances in neural information processing systems},
  volume={31},
  year={2018}
}

@inproceedings{lueckmann2019likelihood,
  title={Likelihood-free inference with emulator networks},
  author={Lueckmann, Jan-Matthis and Bassetto, Giacomo and Karaletsos, Theofanis and Macke, Jakob H},
  booktitle={Symposium on Advances in Approximate Bayesian Inference},
  cpages={32--53},
  year={2019},
  corganization={PMLR}
}

@inproceedings{papamakarios2019sequential,
  title={Sequential neural likelihood: Fast likelihood-free inference with autoregressive flows},
  author={Papamakarios, George and Sterratt, David and Murray, Iain},
  cbooktitle={22nd AISTATS Conference Proceedings},
  cbooktitle={The 22nd International Conference on Artificial Intelligence and Statistics},
  cpages={837--848},
  year={2019},
  corganization={PMLR}
}

@inproceedings{lueckmann2017flexible,
  title={Flexible statistical inference for mechanistic models of neural dynamics},
  author={Lueckmann, Jan-Matthis and Gon{\c{c}}alves, Pedro J and Bassetto, Giacomo and {\"O}cal, Kaan and Nonnenmacher, Marcel and Macke, Jakob H},
  booktitle={31st NeurIPS Conference Proceedings},
  cbooktitle={Proceedings of the 31st International Conference on Neural Information Processing Systems},
  cpages={1289--1299},
  year={2017}
}

@article{papamakarios2016fast,
  title={Fast $\varepsilon$-free inference of simulation models with bayesian conditional density estimation},
  author={Papamakarios, George and Murray, Iain},
  journal={Advances in neural information processing systems},
  volume={29},
  year={2016}
}

@inproceedings{durkan2020contrastive,
  title={On contrastive learning for likelihood-free inference},
  author={Durkan, Conor and Murray, Iain and Papamakarios, George},
  booktitle={International Conference on Machine Learning},
  cpages={2771--2781},
  year={2020},
  organization={PMLR}
}

@article{deistler2022truncated,
  title={Truncated proposals for scalable and hassle-free simulation-based inference},
  author={Deistler, Michael and Goncalves, Pedro J and Macke, Jakob H},
  journal={arXiv preprint},
  year={2022}
}

@article{boelts2022flexible,
  title={Flexible and efficient simulation-based inference for models of decision-making},
  author={Boelts, Jan and Lueckmann, Jan-Matthis and Gao, Richard and Macke, Jakob H},
  journal={Elife},
  cvolume={11},
  cpages={e77220},
  year={2022},
  publisher={eLife Sciences Publications Limited}
}

@inproceedings{arjovsky2017wasserstein,
  title={Wasserstein generative adversarial networks},
  author={Arjovsky, Martin and Chintala, Soumith and Bottou, L{\'e}on},
  booktitle={International conference on machine learning},
  pages={214--223},
  year={2017},
  organization={PMLR}
}

@article{bloem2020probabilistic,
  title={Probabilistic symmetries and invariant neural networks},
  author={Bloem-Reddy, Benjamin and Teh, Yee Whye},
  journal={The Journal of Machine Learning Research},
  volume={21},
  number={1},
  pages={3535--3595},
  year={2020},
  publisher={JMLRORG}
}

@article{abadi2016tensorflow,
  title={Tensorflow: Large-scale machine learning on heterogeneous distributed systems},
  author={Abadi, Mart{\'\i}n and Agarwal, Ashish and Barham, Paul and Brevdo, Eugene and Chen, Zhifeng and Citro, Craig and Corrado, Greg S and Davis, Andy and Dean, Jeffrey and Devin, Matthieu and others},
  journal={arXiv preprint arXiv:1603.04467},
  year={2016}
}

@article{kingma2014adam,
  title={Adam: A method for stochastic optimization},
  author={Kingma, Diederik P and Ba, Jimmy},
  journal={arXiv preprint arXiv:1412.6980},
  year={2014}
}

@article{kim2020softflow,
  title={Softflow: Probabilistic framework for normalizing flow on manifolds},
  author={Kim, Hyeongju and Lee, Hyeonseung and Kang, Woo Hyun and Lee, Joun Yeop and Kim, Nam Soo},
  journal={Advances in Neural Information Processing Systems},
  volume={33},
  pages={16388--16397},
  year={2020}
}

@article{voss2007fast,
  title={Fast-dm: A free program for efficient diffusion model analysis},
  author={Voss, Andreas and Voss, Jochen},
  journal={Behavior research methods},
  volume={39},
  number={4},
  pages={767--775},
  year={2007},
  publisher={Springer}
}

@book{marin2018likelihood,
  title={Likelihood-free model choice},
  author={Marin, Jean-Michel and Pudlo, Pierre and Estoup, Arnaud and Robert, Christian},
  year={2018},
  publisher={Chapman and Hall/CRC}
}

@article{jiang2017learning,
  title={Learning summary statistic for approximate Bayesian computation via deep neural network},
  author={Jiang, Bai and Wu, Tung-yu and Zheng, Charles and Wong, Wing H},
  journal={Statistica Sinica},
  pages={1595--1618},
  year={2017},
  publisher={JSTOR}
}

@article{sailynoja2022graphical,
  title={Graphical test for discrete uniformity and its applications in goodness-of-fit evaluation and multiple sample comparison},
  author={S{\"a}ilynoja, Teemu and B{\"u}rkner, Paul-Christian and Vehtari, Aki},
  journal={Statistics and Computing},
  volume={32},
  number={2},
  pages={1--21},
  year={2022},
  publisher={Springer}
}

@article{smc,
  title={{ABC} {SMC} for parameter estimation and model selection with applications in systems biology},
  author={Toni, Tina},
  journal={Nature Precedings},
  cpages={1--1},
  year={2011},
  publisher={Nature Publishing Group}
}

@article{cranmer2020frontier,
  title={The frontier of simulation-based inference},
  author={Cranmer, Kyle and Brehmer, Johann and Louppe, Gilles},
  journal={Proceedings of the National Academy of Sciences},
  year={2020},
  publisher={National Acad Sciences}
}

@article{durkan2019neural,
  title={Neural spline flows},
  author={Durkan, Conor and Bekasov, Artur and Murray, Iain and Papamakarios, George},
  journal={Advances in neural information processing systems},
  volume={32},
  year={2019}
}

@article{bellagente2022understanding,
  title={Understanding event-generation networks via uncertainties},
  author={Bellagente, Marco and Hau{\ss}mann, Manuel and Luchmann, Michel and Plehn, Tilman},
  journal={SciPost Physics},
  volume={13},
  number={1},
  pages={003},
  year={2022}
}

@inproceedings{ardizzone2018analyzing,
  title={Analyzing inverse problems with invertible neural networks},
  author={Ardizzone, Lynton and Kruse, Jakob and Wirkert, Sebastian and Rahner, Daniel and Pellegrini, Eric W and Klessen, Ralf S and Maier-Hein, Lena and Rother, Carsten and K{\"o}the, Ullrich},
  booktitle={Intl. Conf. on Learning Representations},
  year={2019}
}

@article{gutmann2016bayesian,
  title={Bayesian optimization for likelihood-free inference of simulator-based statistical models},
  author={Gutmann, Michael U and Corander, Jukka},
  journal={Journal of Machine Learning Research},
  year={2016}
}

@article{talts2018validating,
  title={Validating Bayesian inference algorithms with simulation-based calibration},
  author={Talts, Sean and Betancourt, Michael and Simpson, Daniel and Vehtari, Aki and Gelman, Andrew},
  journal={arXiv preprint},
  year={2018}
}

@article{marin2012approximate,
  title={Approximate {B}ayesian computational methods},
  author={Marin, Jean-Michel and Pudlo, Pierre and Robert, Christian P and Ryder, Robin J},
  journal={Statistics and Computing},
  cvolume={22},
  cnumber={6},
  cpages={1167--1180},
  year={2012},
  publisher={Springer}
}

@book{sisson2018handbook,
  title={Handbook of approximate {B}ayesian computation},
  author={Sisson, Scott A and Fan, Yanan and Beaumont, Mark},
  year={2018},
  publisher={CRC Press}
}

@inproceedings{wiqvist2019partially,
  title={Partially exchangeable networks and architectures for learning summary statistics in approximate Bayesian computation},
  author={Wiqvist, Samuel and Mattei, Pierre-Alexandre and Picchini, Umberto and Frellsen, Jes},
  booktitle={International Conference on Machine Learning},
  cpages={6798--6807},
  year={2019},
  corganization={PMLR}
}

@article{chen2020neural,
  title={Neural approximate sufficient statistics for implicit models},
  author={Chen, Yanzhi and Zhang, Dinghuai and Gutmann, Michael and Courville, Aaron and Zhu, Zhanxing},
  journal={International Conference on Learning Representations},
  year={2021}
}

@article{beaumont2009adaptive,
  title={Adaptive approximate {B}ayesian computation},
  author={Beaumont, Mark A and Cornuet, Jean-Marie and Marin, Jean-Michel and Robert, Christian P},
  journal={Biometrika},
  volume={96},
  number={4},
  pages={983--990},
  year={2009},
  publisher={Oxford University Press}
}

@article{del2012adaptive,
  title={An adaptive sequential {Monte Carlo method for approximate Bayesian computation}},
  author={Del Moral, Pierre and Doucet, Arnaud and Jasra, Ajay},
  journal={Statistics and Computing},
  cvolume={22},
  cnumber={5},
  cpages={1009--1020},
  year={2012},
  publisher={Springer}
}

@article{marjoram2003markov,
  title={Markov chain {M}onte {C}arlo without likelihoods},
  author={Marjoram, Paul and Molitor, John and Plagnol, Vincent and Tavar{\'e}, Simon},
  journal={Proceedings of the National Academy of Sciences},
  cvolume={100},
  cnumber={26},
  cpages={15324--15328},
  year={2003},
  publisher={National Acad Sciences}
}

@article{picchini2014inference,
  title={Inference for {SDE} models via approximate Bayesian computation},
  author={Picchini, Umberto},
  journal={Journal of Computational and Graphical Statistics},
  volume={23},
  number={4},
  pages={1080--1100},
  year={2014},
  publisher={Taylor \& Francis}
}

@article{wood2010statistical,
  title={Statistical inference for noisy nonlinear ecological dynamic systems},
  author={Wood, Simon N},
  journal={Nature},
  cvolume={466},
  cnumber={7310},
  cpages={1102--1104},
  year={2010},
  publisher={Nature Publishing Group}
}

@article{price2018bayesian,
  title={Bayesian synthetic likelihood},
  author={Price, Leah F and Drovandi, Christopher C and Lee, Anthony and Nott, David J},
  journal={Journal of Computational and Graphical Statistics},
  cvolume={27},
  cnumber={1},
  cpages={1--11},
  year={2018},
  publisher={Taylor \& Francis}
}

@article{andrieu2010particle,
  title={Particle Markov chain Monte Carlo methods},
  author={Andrieu, Christophe and Doucet, Arnaud and Holenstein, Roman},
  journal={Journal of the Royal Statistical Society: Series B},
  cjournal={Journal of the Royal Statistical Society: Series B (Statistical Methodology)},
  cvolume={72},
  cnumber={3},
  cpages={269--342},
  year={2010},
  publisher={Wiley Online Library}
}

@article{lueckmann2021benchmarking,
Author = {Jan-Matthis Lueckmann and Jan Boelts and David S. Greenberg and Pedro J. Gonçalves and Jakob H. Macke},
Title = {Benchmarking Simulation-Based Inference},
Year = {2021},
journal = {arXiv preprint},
}

@article{wiqvist2021sequential,
Author = {Samuel Wiqvist and Jes Frellsen and Umberto Picchini},
Title = {Sequential Neural Posterior and Likelihood Approximation},
Year = {2021},
journal = {arXiv preprint},
}

@article{ratcliff2008,
  cdoi = {10.1162/neco.2008.12-06-420},
  year = {2008},
  publisher = {{MIT} Press - Journals},
  volume = {20},
  number = {4},
  pages = {873--922},
  author = {Roger Ratcliff and Gail McKoon},
  title = {The Diffusion Decision Model: Theory and Data for Two-Choice Decision Tasks},
  journal = {Neural Computation}
}

@article{salvatier_probabilistic_2016,
	title = {Probabilistic programming in {Python} using {PyMC3}},
	cvolume = {2},
	cissn = {2376-5992},
	curl = {https://peerj.com/articles/cs-55},
	cdoi = {10.7717/peerj-cs.55},
	urldate = {2022-12-02},
	journal = {PeerJ Computer Science},
	author = {Salvatier, John and Wiecki, Thomas V. and Fonnesbeck, Christopher},
	year = {2016},
	cpages = {e55},
}

@article{picchini2022guided,
  title={Guided sequential schemes for intractable {B}ayesian models},
  author={Picchini, Umberto and Tamborrino, Massimiliano},
  journal={arXiv preprint arXiv:2206.12235},
  year={2022}
}

@inproceedings{
glockler2022snvi,
title={Variational methods for simulation-based inference},
author={Manuel Gl{\"o}ckler and Michael Deistler and Jakob H. Macke},
booktitle={International Conference on Learning Representations},
year={2022},
}

\clearpage
\appendix
\onecolumn 
\begin{center}
	\Large\bfseries\textsc{Appendix}
\end{center}

\section{Frequently Asked Questions (FAQ)}

\textbf{Q: How can I reproduce the results?}\\[3pt]
Code to reproduce all results is available in the repository at \url{https://github.com/bayesflow-org/JANA-Paper}.

\vspace*{6pt}\textbf{Q: How can I apply JANA to my own Bayesian models?}\\[3pt]
Simulation-based algorithms for jointly amortized inference are implemented in the \texttt{BayesFlow} library. 
Take a look at the code and tutorials, available at:
\url{https://github.com/stefanradev93/BayesFlow}.

\vspace*{6pt}\textbf{Q: When should I use amortized inference instead of sequential methods?}\\[3pt]
Whenever you want to follow a principled Bayesian workflow and you have lots of data sets on which a Bayesian model needs to be applied independently.

\vspace*{6pt}\textbf{Q: Does amortization come at the cost of wasteful simulations?}\\[3pt]
Some previous papers assume that this is generally the case. 
On the contrary, we believe that wasteful simulations are primarily the consequence of poorly chosen priors, whereas modern neural networks actually profit from broader simulation scopes, as long as the priors are informative. Moreover, amortization makes a principled Bayesian workflow much easier than case-based inference.
Still, specifying sensible joint priors is not always easy.

\vspace*{6pt}\textbf{Q: Can you somehow combine the three networks and utilize weight sharing?}\\[3pt]
Finding a suitable weight sharing approach which is applicable to various model structures---such as exchangeable or Markovian---proves challenging.
Since JANA is an attempt at a universal method, we refrain from customizing the overall architecture to suit a particular model structure.
Instead of weight sharing, we exploit the \text{probabilistic symmetries} of joint Bayesian learning, which is universal across all model structures (see Figure 1 of the main paper).
Although it remains a possible area for further investigation, we are uncertain whether weight sharing in our context is even desirable.

\vspace*{6pt}\textbf{Q: Can I use a different type of generative network for the posterior or likelihood networks?}\\[3pt]
JANA can operate with arbitrary conditional density approximators. 
However, it is important that these approximators are able to efficiently compute \textit{normalized densities} for the purpose of marginal likelihood and posterior predictive estimation.

\vspace*{6pt}\textbf{Q: Why do you need a summary network?}\\[3pt]
Because most real world data comes in various sizes and shapes. 
Thus, we need an interface between the Bayesian model and the posterior network which renders the latter applicable to various sizes and shapes.

\vspace*{6pt}\textbf{Q: Can you also use a summary network for the likelihood network?}\\[3pt]
It is possible and can be helpful if the parameter space of the reference Bayesian model requires some form of compression.
Indeed, in the second iteration of the paper, we included a Bayesian denoising experiment (\textbf{Experiment 5}) which equips the surrogate likelihood with a convolutional summary network.

\vspace*{6pt}\textbf{Q: Is it necessary to have normalized likelihood estimates or would a standard feedforward neural network suffice?}\\[3pt]
A normalized likelihood is necessary to estimate the expected log predictive density (ELPD) for approximating out-of-sample predictive performance via cross validation or log marginal likelihoods (LMLs) for approximating Bayes factors. Normalization is also needed to compare likelihoods obtained from different models (which might otherwise report unnormalized likelihoods at different scales). If none of these (log) likelihood metrics is needed for a particular analysis, normalization of the likelihood network is not strictly required.

\FloatBarrier
\clearpage
\section{Code}

The code and instructions for running and reproducing all experiments are available at the project's repository, hosted at \url{https://github.com/bayesflow-org/JANA-Paper}.
We use fixed seeds for the random number generators of test (held-out) sets. 
Training uses no seeds, as we believe the methods to be stable enough to converge on any run.

\section{Method Details}
\subsection{Pseudocode}

\begin{algorithm}[H]
	\caption{Jointly amortized neural approximation: offline training using a pre-simulated training set}
	\label{alg:jana}
	\begin{algorithmic}[1]
		\REQUIRE{Bayesian model $\joint$; summary network $\mathcal{H}_{\psib}$; posterior network $\mathcal{P}_{\phib}$; likelihood network $\mathcal{L}_{\etab}$; number of simulations $N$ (budget); batch size $B$}
		\STATE{Initialize $\mathcal{D} = \{\}$.}
		\FOR{$n=1,\ldots,N$}
		\STATE{Sample from prior: $\thetab_n \sim \prior$}
		\STATE{Sample from (implicit) likelihood: $\x_n \sim p(\x \given \thetab_n)$}
		\STATE{Add simulations to training data: $\mathcal{D} := \mathcal{D} \cup \{(\thetab_n, \x_n)\}$ \hfill}
		\ENDFOR
		\WHILE{not converged}
		\STATE{Sample batch from training data: $\{(\thetab_b, \x_b)\}_{b=1}^B \sim \mathcal{D}$}
		\STATE{Compute Monte Carlo estimate of loss function over batch (Eq.~12).}
		\STATE{Update neural network parameters $(\psib, \phib, \etab)$ via backpropagation.}
		\ENDWHILE
		\RETURN{trained networks $\lika, \postah, \mathcal{H}_{\psib}$}
	\end{algorithmic}
\end{algorithm}

\subsection{Likelihood Networks for Exchangeable Data}

Exchangeable models generate IID data, that is, each run $\model$ with a fixed configuration $\thetab$ is independent of all other runs. 
Thus, for $N$ runs of such a (memoryless or stateless) model, the likelihood decomposes into the product of point-wise likelihoods:
\begin{equation}
	\lik = \prod_{n=1}^N p(\x_n \given \thetab)
\end{equation}

Accordingly, we can represent such data as unordered sets and simply apply the likelihood network exchangeably by concatenating each $\x_n$ with $\thetab$ in each coupling layer.
The forward pass for a single conditional affine coupling layer \autocite{ardizzone2019guided, radev2020bayesflow} of an exchangeable likelihood network is given by:
\begin{align} 
	\z_n^{\mathcal{A}} &= \x_n^{\mathcal{A}} \odot \exp(S_1(\x_n^{\mathcal{B}}; \thetab)) + T_1(\x_n^{\mathcal{B}}; \thetab)\nonumber  \\ 
	\z_n^{\mathcal{B}} &= \x_n^{\mathcal{B}} \odot \exp(S_2(\z_n^{\mathcal{A}}; \thetab)) + T_2(\z_n^{\mathcal{A}}; \thetab)\nonumber,
\end{align}
where $\x_n = (\x_n^{\mathcal{A}}, \x_n^{\mathcal{B}})$ is a disjoint partition of the input data at position $n$, $\z_n = (\z_n^{\mathcal{A}}, \z_n^{\mathcal{B}})$ is the  corresponding latent partition, and the functions $S_1$, $S_2$, $T_1$, $T_2$ are implemented as multi-headed fully connected (FC) neural networks (with trainable parameters suppressed for clarity). 
The forward pass for neural spline flows \autocite{durkan2019neural} is modified accordingly, such that the spline parameters are generated exchangeably, conditioned on the parameter vector $\thetab$.

\subsection{Likelihood Networks for Markovian Data}

The widely used family of Markovian models factorize in a way that the probability of each data point depends on previous data points:
\begin{equation}
	\lik = \prod_{n=1}^Np(\x_n\given\thetab,\x_{1:{n-1}})
\end{equation}

Such models require a slightly different coupling layer design which respects their non-IID outputs.
To this end, we augment standard coupling layers with a conditional recurrent (GRU) memory $\h_n = M(\thetab, \x_n; \h_{n-1})$ which encodes temporal dependencies into a hidden state vector $\h$.

For instance, the forward pass for a single conditional affine coupling layer of the non-exchangeable likelihood network is then given by:
\begin{align} 
	\h_n &= M(\thetab, \x_n; \h_{n-1})\nonumber\\
	\z_n^{\mathcal{A}} &= \x_n^{\mathcal{A}} \odot \exp(S_1(\x_n^{\mathcal{B}}; \thetab, \h_{n-1})) + T_1(\x_n^{\mathcal{B}}; \thetab, \h_{n-1}) \nonumber\\ 
	\z_n^{\mathcal{B}} &= \x_n^{\mathcal{B}} \odot \exp(S_2(\z_n^{\mathcal{A}}; \thetab, \h_{n-1})) + T_2(\z_n^{\mathcal{A}}; \thetab, \h_{n-1}) \nonumber,
\end{align}
where now each latent representation $\z_n$ at position $n$ depends on the preceding data points, as encoded by $\h_{n-1}$, and the functions $S_1$, $S_2$, $T_1$, $T_2$ are implemented as multi-headed fully connected (FC) neural networks.
The forward pass for neural spline flows \autocite{durkan2019neural} is modified accordingly, such that the spline parameters are generated using a recurrent memory, conditioned on the parameter vector $\thetab$.

\subsection{Correctness of Joint Simulation-Based Training}

To show that our jointly optimized criterion yields correct posterior, likelihood, and marginal likelihood inference, consider first the joint optimization of the posterior and the summary network:
\begin{align}
	(\phib^*, \psib^*) &= \argmin_{\phib, \psib}\mathbb{E}_{p^*(\x)}\left[\mathbb{KL}(\post\,||\,\postah) \label{eq:kl_full}
	\right] \\
	&= \argmin_{\phib, \psib}\mathbb{E}_{p^*(\x)}\left[ \mathbb{E}_{p(\thetab \given \x)}\left[\log \post - \log \postah \right]\right] \\
	&= \argmin_{\phib, \psib}\mathbb{E}_{p^*(\x)}\left[ \mathbb{E}_{p(\thetab \given \x)}\left[-\log \postah \right]\right] \label{eq:kl_reduced}
\end{align}
The above criterion (Eq.~\ref{eq:kl_full}) states that, in order to achieve proper amortized inference, we want to minimize the Kullback-Leibler (KL) divergence between the analytic and the approximate posterior density in expectation over all possible observations from the true data-generating distribution $p^*$.
This reduces to the expected negative log posterior (Eq.~\ref{eq:kl_reduced}), since the negative entropy of the analytic posterior $-\mathbb{H}\left[ \post \right] = \mathbb{E}_{p(\thetab \given \x)}\left[\log \post \right]$ does not depend on the neural network parameters $(\phib, \psib)$.

In order to make amortized posterior inference tractable under Eq.~\ref{eq:kl_full}, we need to assume that the true data-generating distribution $p^*$ and the model-implied (i.e., prior predictive) distribution $p(\x) = \mathbb{E}_{\prior}\left[\lik\right]$ match, that is, $p^*(\x) = p(\x)$ for any $\x$.
In other words, we invoke the so-called \textit{closed-world assumption}, which states the Bayesian model is a correct representation of the true data-generating distribution.
In that case, we can simply replace $p^*(\x)$ with $p(\x)$ and write our criterion as:
\begin{align}
	(\phib^*, \psib^*) &= \argmin_{\phib, \psib}\mathbb{E}_{p(\x)}\left[ \mathbb{E}_{p(\thetab \given \x)}\left[-\log \postah \right]\right] \\
	&= \argmin_{\phib, \psib}\mathbb{E}_{p( \thetab, \x)}\left[-\log \postah \right] \label{eq:kl_final}
\end{align}
We can now readily approximate the expectation with its empirical mean over a data set of simulations $(\thetab, \x) \sim \mathcal{D}$ generated from the Bayesian joint model $p(\thetab, \x)$. 
Thereby, we leverage the fact that we can directly evaluate $-\log \postah$ (and not a lower bound) due to the use of a normalizing flow (NF) for the approximate posterior.
Moreover, as shown by \textcite{radev2020bayesflow}, perfect convergence under Eq.~\ref{eq:kl_final} ensures that the summary network learns maximally informative (ideally sufficient) summary statistics and the posterior network samples from the analytic posterior.
Note, however, that if the key assumption of $p^*(\x) = p(\x)$ is violated for some $\x$, then the approximate posterior may no longer be a faithful representation of the analytic posterior in general. 
This situation motives the introduction of the summary space distribution $p(\mathcal{H}_{\psib}(\x))$ (to be discussed shortly).

As for the likelihood network, we aim to minimize the KL divergence between the analytic and the approximate posterior density in expectation over all possible parameter configurations from the prior:
\begin{align}
	\etab^* &= \argmin_{\etab}\mathbb{E}_{\prior}\left[\mathbb{KL}(\lik\,||\,\lika) \label{eq:kl_full_lik}
	\right]
\end{align}
Following the same reasoning as for the posterior KL and leveraging the fact that the expectation runs over a model-implied quantity (i.e., the prior), the above criterion directly reduces to:
\begin{align}
	\etab^* = \argmin_{\etab}\mathbb{E}_{p( \thetab, \x)}\left[-\log \lika \right] \label{eq:kl_final_lik}
\end{align}
Observing that both optimization criteria (Eq.~\ref{eq:kl_final} and Eq.~\ref{eq:kl_final_lik}) include an expectation over the Bayesian joint $p(\thetab, \x)$, we arrive at our combined loss function:
\begin{align}
	\mathcal{L}_{\text{JANA}} := -\mathbb{E}_{p( \thetab, \x)}\big[\log \lika + \log \postah \big] \label{eq:kl_final_joint}
\end{align}
Thus, under the closed-world assumption, proper minimization of this loss ensures correct posterior and likelihood approximation.
However, in practice, we want to obtain some measure of the mismatch between $p^*(\x)$ and $p(\x)$.
Moreover, since $\x$ is typically a high dimensional data set (e.g., a data set of multivariate IID observations) and the posterior network only ``sees'' $\x$ through the lens of the summary network, it makes sense to measure the potential mismatch in the reduced summary space given by $\mathcal{H}_{\psib}(\x)$.
To make the detection task even easier, we want to re-structure the unrestricted $p(\mathcal{H}_{\psib}(\x))$ into a simple distribution (e.g., Gaussian) with a well-defined notion of an outlier.
Accordingly, we utilize the Maximum Mean Discrepancy \autocite[MMD;][]{Gretton2012}:
\begin{align}
	\mathbb{MMD}^2\big[p^*(\x)\,||\,p(\x)\big] =
	\mathbb{E}_{
		p^*(\x)}\big[\kappa(\x, \x')\big]
	+ \mathbb{E}_{
		p(\x)}\big[\kappa(\x, \x')\big]
	- 2 \mathbb{E}_{
		\x \sim p^*(\x), \x' \sim  p(\x)}\big[\kappa(\x, \x')\big],
\end{align}
where $\kappa(\cdot, \cdot)$ is any reproducing kernel and we simply replace $\x$ with $\mathcal{H}_{\etab}(\x)$.
The MMD is a suitable alternative to the KL whenever we want to measure the distance between two distributions from which we can obtain samples but cannot evaluate explicitly.
Our augmented loss function then becomes:
\begin{align}
	\mathcal{L}_{\text{JANA-MMD}} := -\mathbb{E}_{p( \thetab, \x)}\big[\log \lika + \log \postah \big] + \lambda \cdot \mathbb{MMD}^2\big[p(\mathcal{H}_{\psib}(\x))\,||\,\mathcal{N}(\boldsymbol{0}, \mathbb{I})\big]\label{eq:kl_final_joint_mmd},
\end{align}
where $\mathcal{N}(\boldsymbol{0}, \mathbb{I})$ denotes a spherical multivariate Gaussian distribution.
Note, that, in theory, proper minimization of the MMD term does not trade off the performance of the posterior network, but simply implies a reparameterization $\phib \rightarrow \phib', \psib \rightarrow \psib'$, such that:
\begin{align}
	p(\thetab) = \int \postah\,p(\x)\,\diff\x = \int p_{\phib'}(\thetab \given \mathcal{H}_{\psib'}(\x))\, \mathcal{N}(\mathcal{H}_{\psib'}(\x)\,\given \,\boldsymbol{0}, \mathbb{I})\,\diff\x
\end{align}

In a particular empirical setting, neural network parameters $(\phib, \psib)$ may be more easily reachable by a given optimizer than corresponding parameters $(\phib', \psib')$, resulting in a practical trade-off. 
However, \textcite{schmitt2022bayesflow} did not observe a diminished performance of amortized posterior approximators trained with a structured summary space, warranting promising results and further investigation into latent summary spaces.

Finally, the correctness of the posterior and likelihood networks trivially implies a correct marginal likelihood (i.e., model evidence) due to the probabilistic change-of-variable resulting from Bayes' rule:
\begin{equation}
	\post = \frac{\lik\,\prior}{p(\x)} \Longleftrightarrow p(\x) = \lik\,\frac{\prior}{\post}
\end{equation}
Thus, assuming perfect convergence of the posterior and the likelihood network under either $\mathcal{L}_{\text{JANA}}$ (Eq.~\ref{eq:kl_final_joint}) or $\mathcal{L}_{\text{JANA-MMD}}$ (Eq.~\ref{eq:kl_final_joint_mmd}), we can compute the log marginal likelihood (LML) of $\x$ by using any single $\thetab \sim p(\thetab)$ as:
\begin{equation}
	\log p(\x) = \log \lika + \log \prior - \log \postah \label{eq:lml} 
\end{equation}
Moreover, it follows, that we can use any violation of Eq.~\ref{eq:lml} to diagnose non-convergence and measure the joint approximation error incurred by the networks.

\FloatBarrier
\clearpage
\section{Implementation Details and Additional Results}

All experiments are implemented using the BayesFlow library \url{https://github.com/stefanradev93/BayesFlow} built on top of TensorFlow \autocite{abadi2016tensorflow}.
Throughout, we use an Adam optimizer \autocite{kingma2014adam} with an initial learning rate between $0.0005$ and $0.001$, default hyperparameters, and a cosine learning rate decay schedule.
All networks are trained on a single machine equipped with an NVIDIA\textsuperscript{\textregistered} T4 graphics accelerator with 16GB of GPU memory. 

\FloatBarrier
\subsection{Experiment 1: Ten Benchmarks}

We follow the model specifications from \textcite{lueckmann2021benchmarking}.
Model implementations are directly imported from the BayesFlow library under MIT license because this implementation has no dependencies on a particular deep learning framework.
For inspecting the software code for the benchmark implementations, we kindly refer the reader to the BayesFlow repository \url{https://github.com/stefanradev93/BayesFlow/tree/master/bayesflow/benchmarks}.
\autoref{tab:app:benchmarks:overview} contains an overview of the benchmarks and core network settings. The full network configurations can be inspected in the code section of the \textbf{Appendix}.

\begin{table}[h]
	\begin{center}
		\caption{Overview of the model and training configurations for \textbf{Experiment 1}.}
		\label{tab:app:benchmarks:overview}
		\begin{tabular}{r|l|c|c|c|c|c|l}
			\textbf{\#} & \textbf{Benchmark name} & \textbf{\# Dimensions}$^1$ & \textbf{Epochs} & \textbf{Batch size} & \textbf{LR} & \textbf{\# Coupling}$^2$ & \textbf{Results} \\
			\hline
			1 & Gaussian Linear & (10, 10) & 50 & 64  & 0.001 & (5, 5) & \autoref{fig:app:benchmark:gaussian_linear}\\
			2 & Gaussian Linear Uniform & (10, 10) & 50 & 64  & 0.001  & (5, 5) & \autoref{fig:app:benchmark:gaussian_linear_uniform}\\
			3 & SLCP$^3$ & (8, 5) & 100 & 32  & 0.0005 & (4, 6) & \autoref{fig:app:benchmark:slcp}\\
			4 & SLCP$^3$ with Distractors & (100, 5) & 60 & 32 & 0.001 & (6, 8) & \autoref{fig:app:benchmark:slcp_distractors}\\
			5 & Bernoulli GLM & (10, 10) & 50 & 32  & 0.0001 & (5, 8) & \autoref{fig:app:benchmark:bernoulli_glm}\\
			6 & Bernoulli GLM Raw & (100, 10) & 50 & 32 & 0.0001 & (8, 8) & \autoref{fig:app:benchmark:bernoulli_glm_raw}\\
			7 & Gaussian Mixture & (2, 2) & 150 & 64 & 0.0005  & (6, 6) & \autoref{fig:app:benchmark:gaussian_mixture}\\
			8 & Two Moons & (2, 2) & 50 & 32  & 0.0005 & (6,6)  & \autoref{fig:app:benchmark:two_moons}\\
			9 & SIR & (10, 2) & 250 & 32 & 0.0001 & (6,6)  &\autoref{fig:app:benchmark:sir}\\
			10 & Lotka-Volterra & (20, 4) & 150 & 128 & 0.001 & (8,6) & \autoref{fig:app:benchmark:lotka_volterra}\\
		\end{tabular}\\
	\end{center}
	\hspace*{0.5cm}{\footnotesize $^1$ Dimensionality of the Bayesian model, denoted as a tuple for $\x$ and $\thetab$, respectively.}\\
	\hspace*{0.5cm}{\footnotesize $^1$ Number of coupling layers, denoted as a tuple for the likelihood and posterior network, respectively.}\\
	\hspace*{0.5cm}{\footnotesize $^2$ Simple Likelihood, Complex Posterior.}
\end{table}

The following figures show the loss history (training and validation) as well as detailed calibration diagnostics for the posterior and joint learning tasks.
Note that the simulation budget is fixed at $10\,000$ simulations.
However, depending on the benchmark, the \emph{number of training steps} may vary (i.e., Gaussian Linear is trivial to learn and requires a few epochs, in contrast to a more challenging benchmark, such as Lotka-Volterra).

Further, note that most of these models are \textit{not} meaningful for joint or likelihood estimation in their original formulation.
Still, we apply JANA to all benchmarks for the sake of completeness, as these experiments serve as a proof-of-concept for more advanced applications.

Special care is needed for the Bernoulli GLM Raw model, as its likelihood yields $N$ IID binary data points. 
These should neither be directly modeled as $N$-dimensional vectors (as this completely ignores the permutation-invariance of the data), nor as exchangeable inputs for coupling-based invertible networks (as the latter assumes at least two-dimensional continuous outputs).
In order to tackle the likelihood of this model, we augment each binary data point $x_n$ with an independent random variate $u_n \sim \mathcal{N}(0, 1)$ and use a SoftFlow architecture \autocite{kim2020softflow} for dequantization of the binary data.

\clearpage

\appendixbenchmark{1}{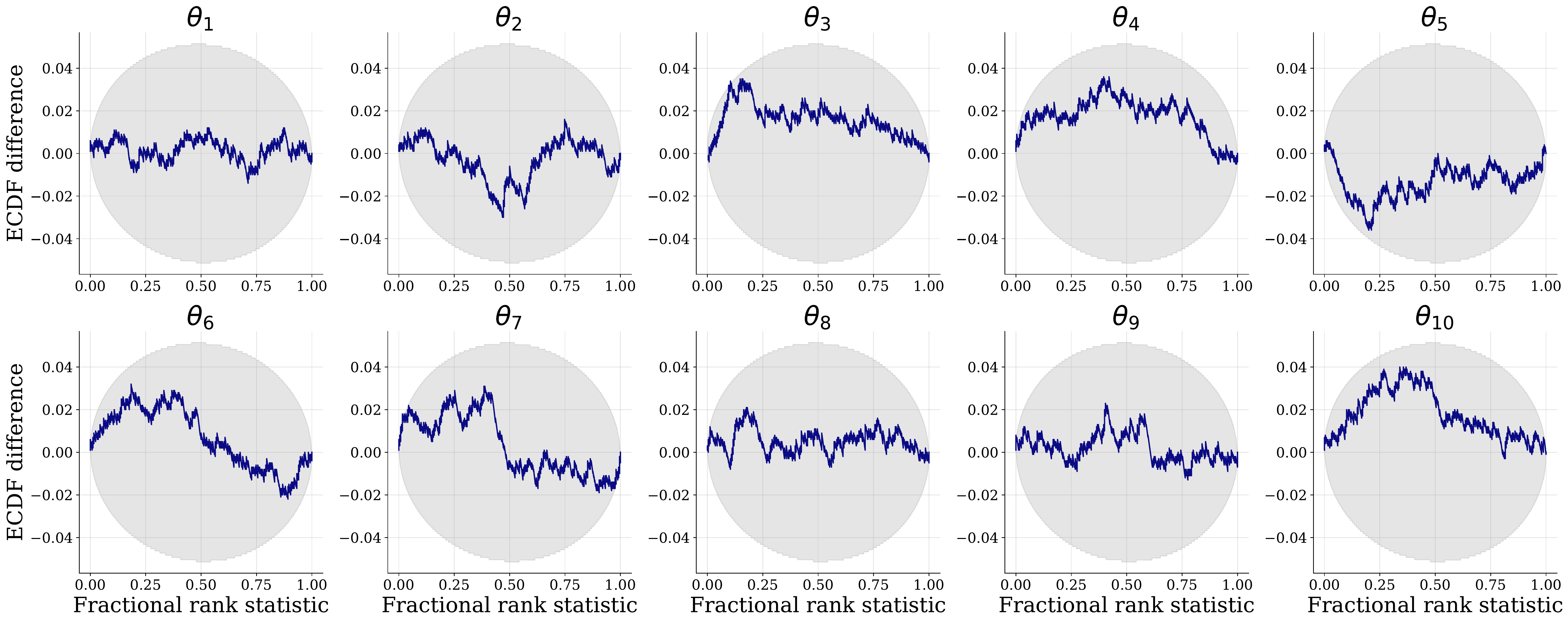}{Gaussian Linear}
\appendixbenchmark{2}{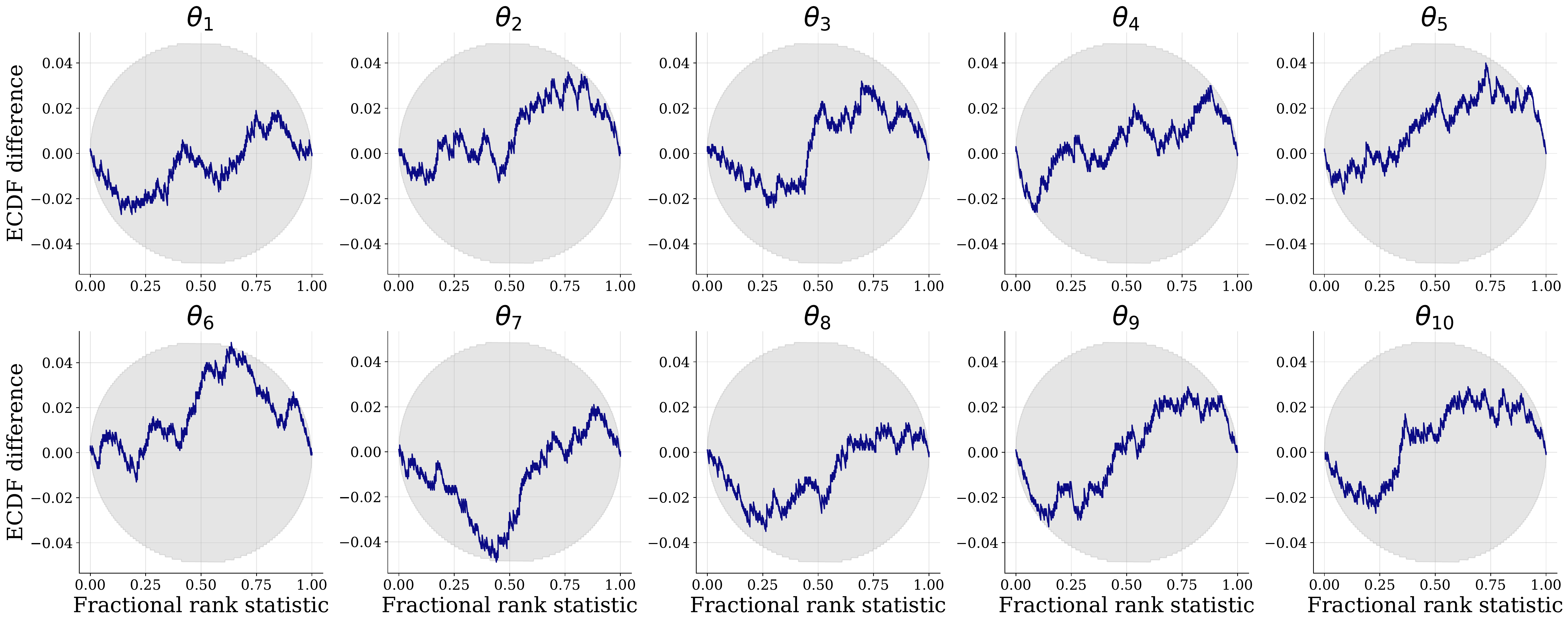}{Gaussian Linear Uniform}
\appendixbenchmark{3}{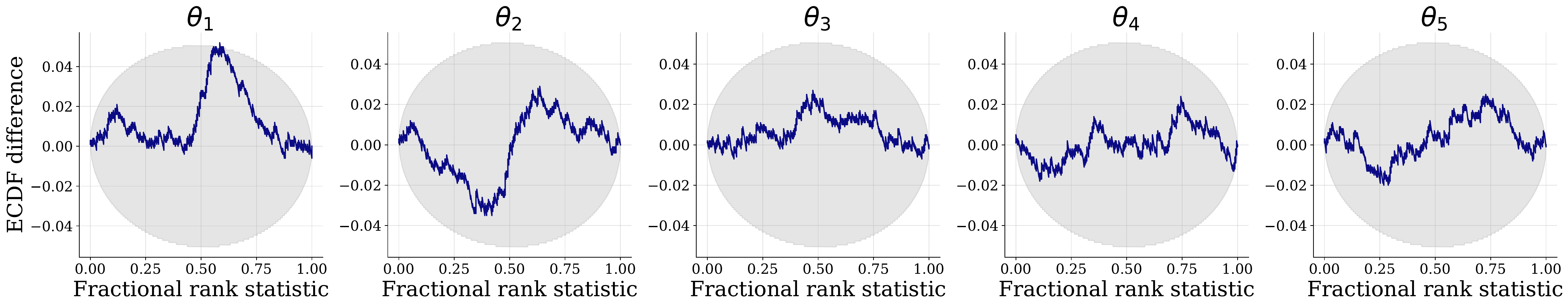}{Simple Likelihood Complex Posterior}
\appendixbenchmark{4}{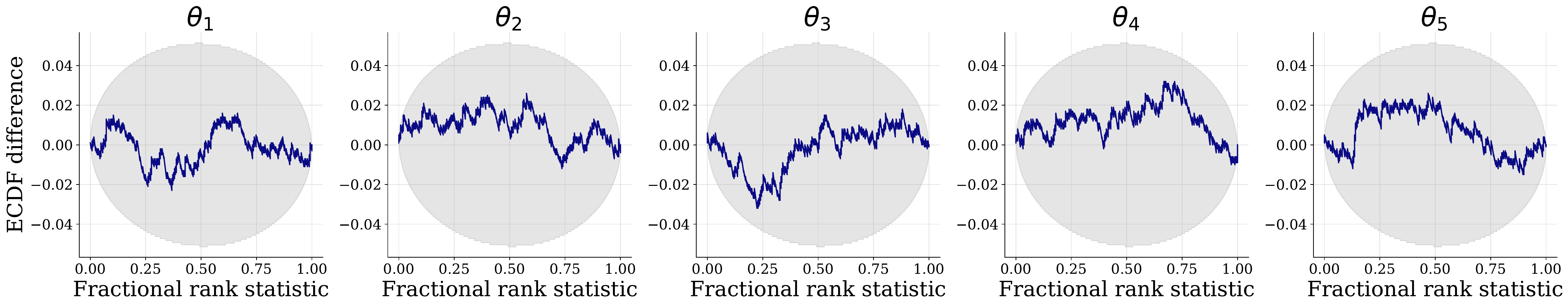}{Simple Likelihood Complex Posterior with Distractors}
\appendixbenchmark{5}{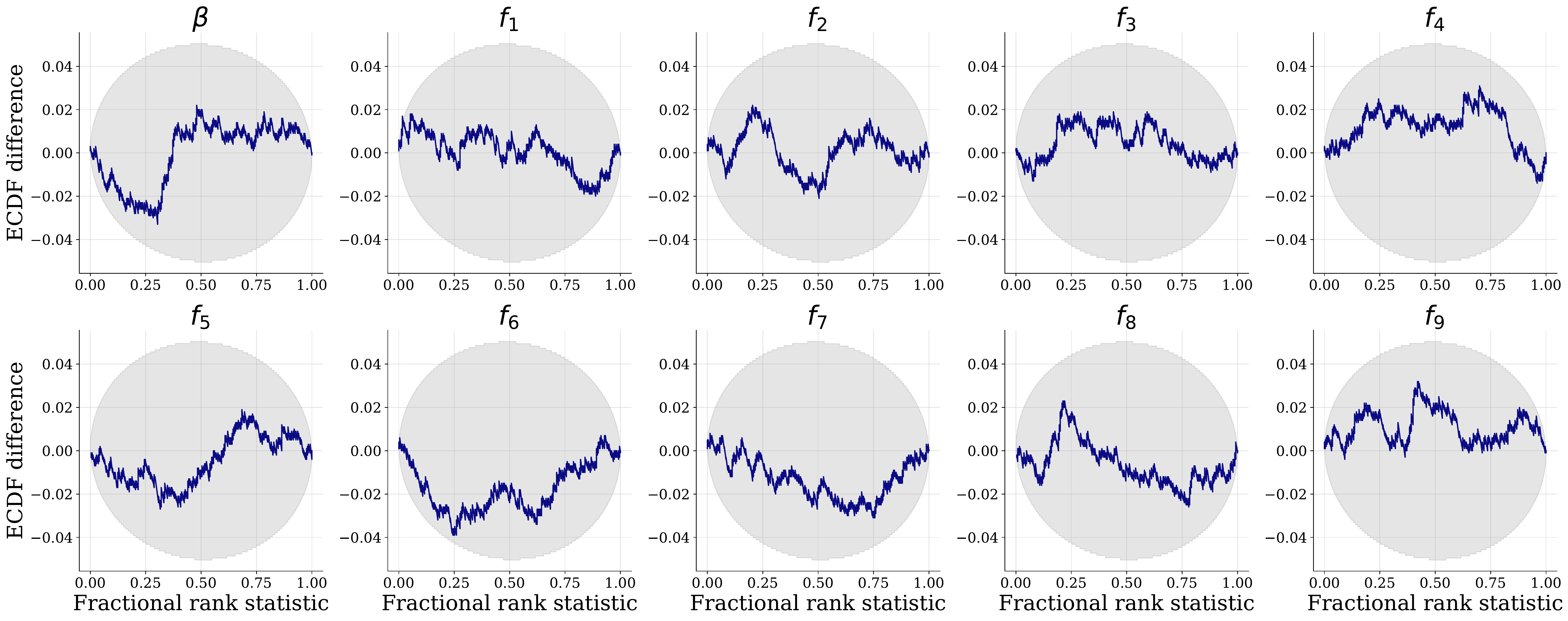}{Bernoulli GLM}
\appendixbenchmark{6}{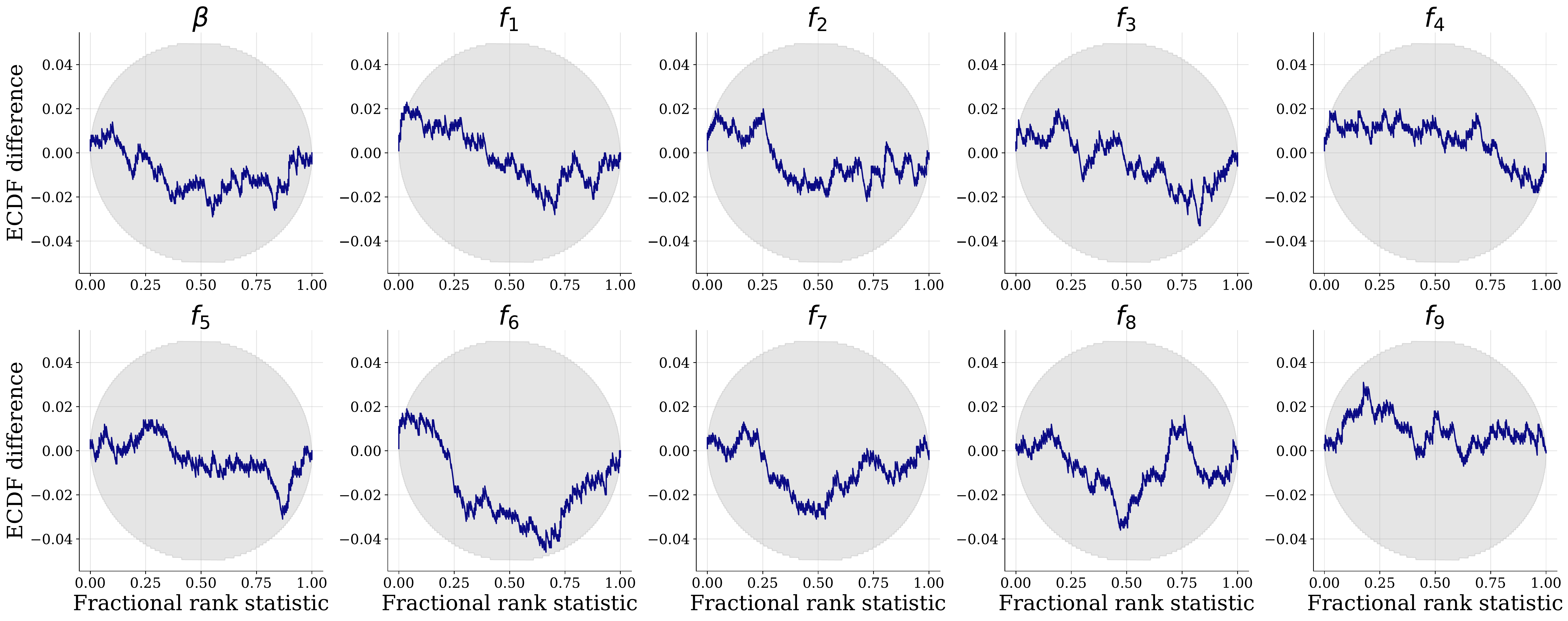}{Bernoulli GLM raw}
\appendixbenchmarkcompressed{7}{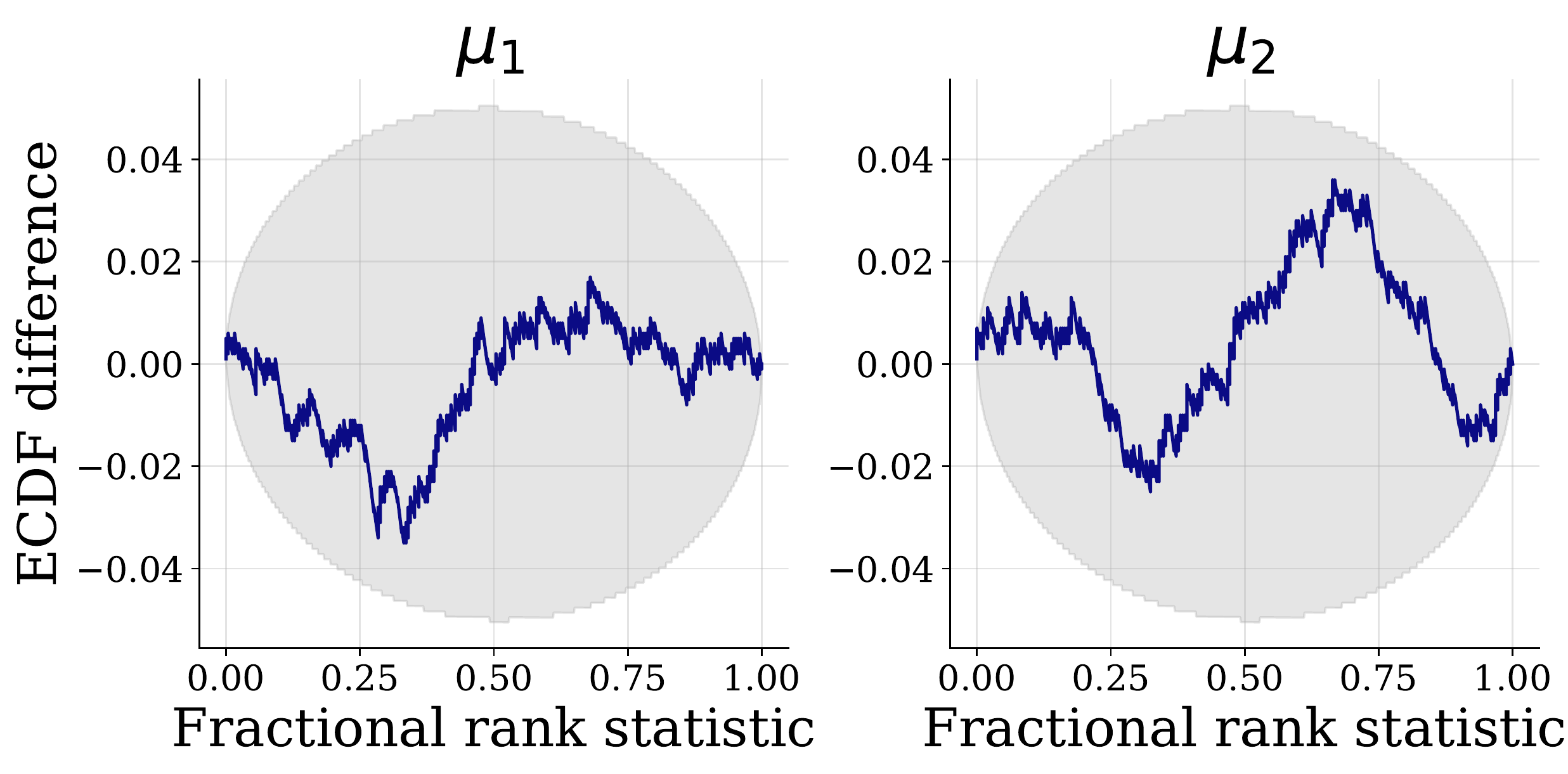}{Gaussian Mixture}
\appendixbenchmarkcompressed{8}{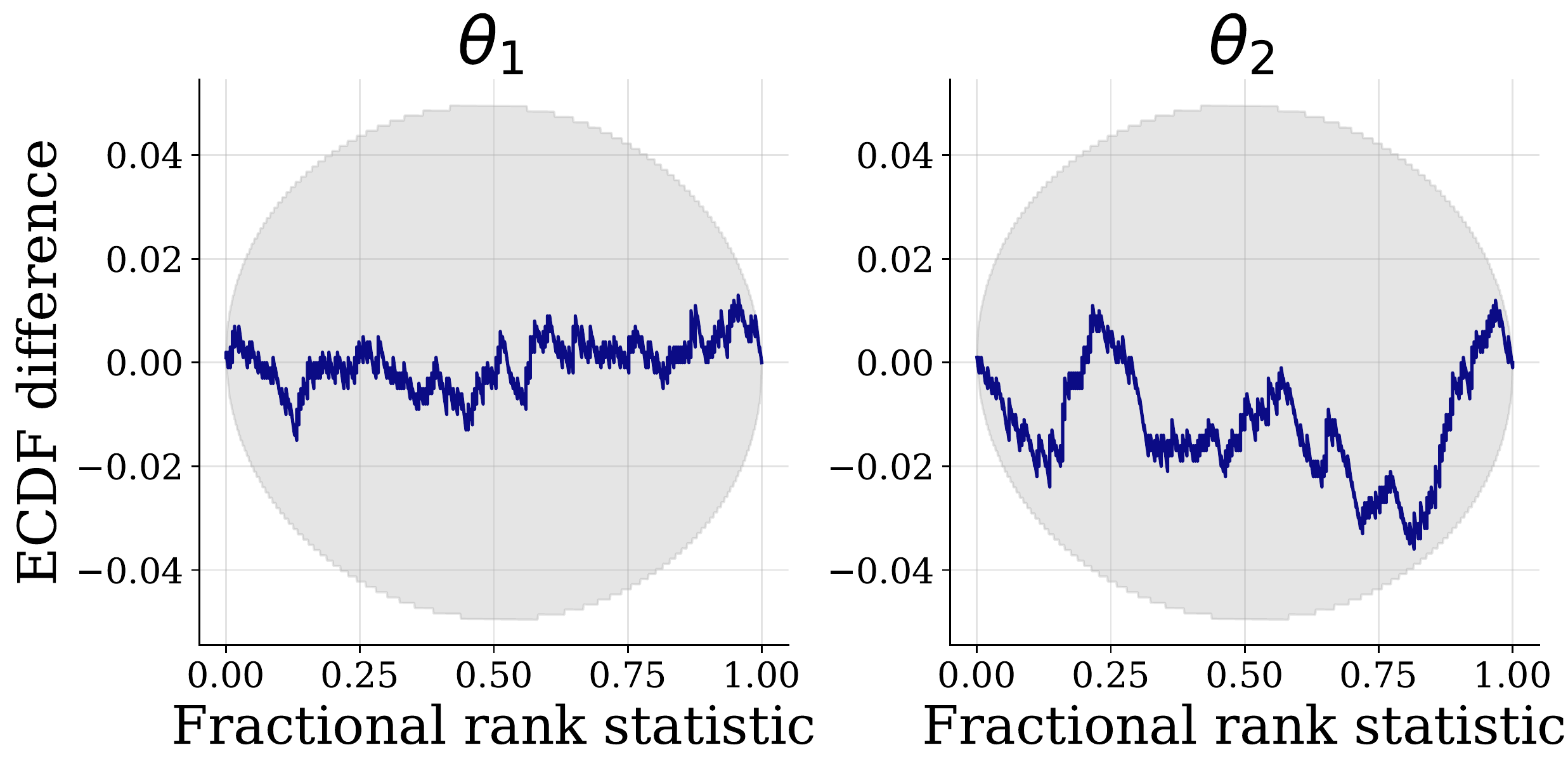}{Two Moons}
\appendixbenchmarkcompressed{9}{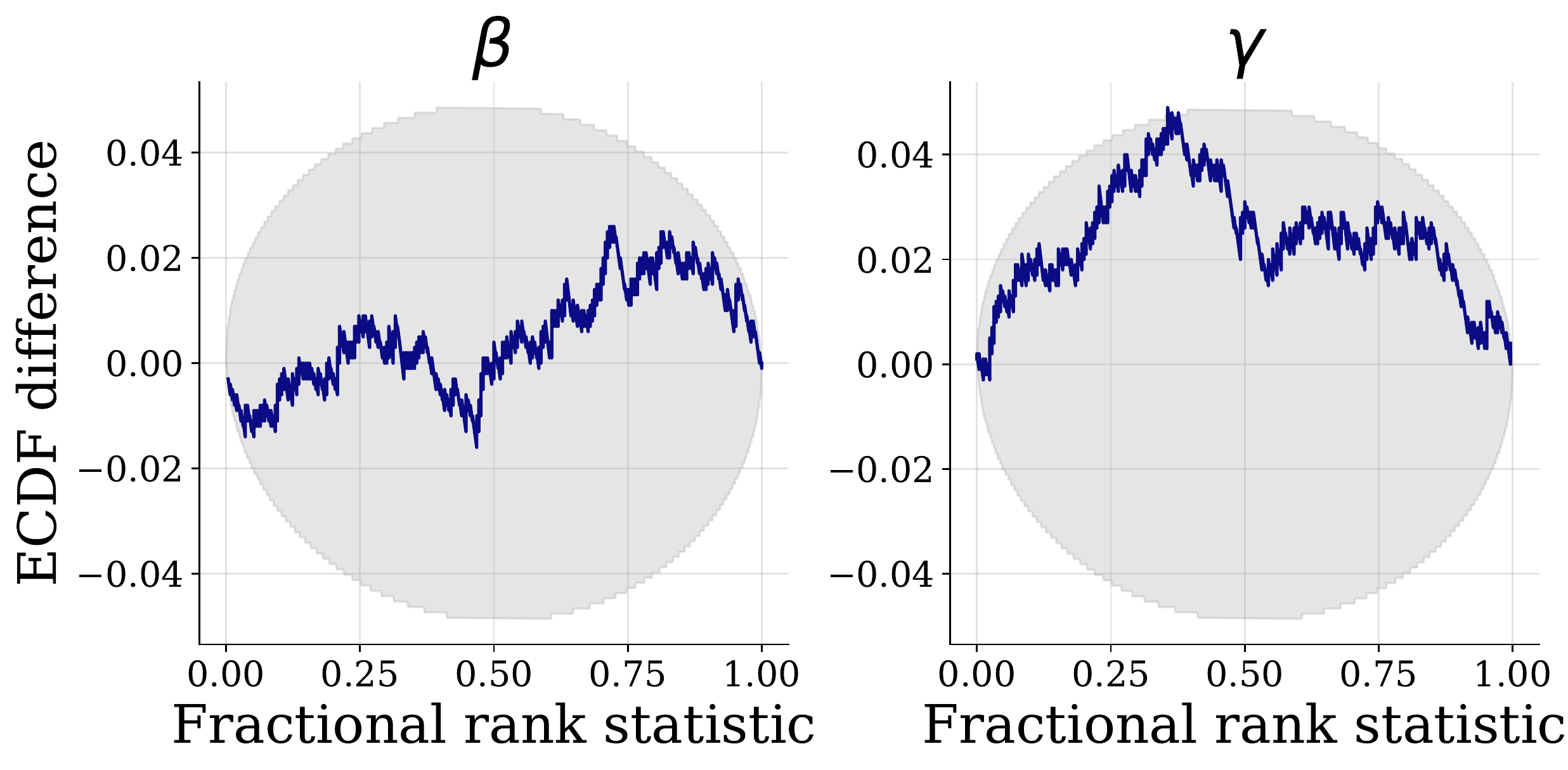}{SIR time series}
\appendixbenchmark{10}{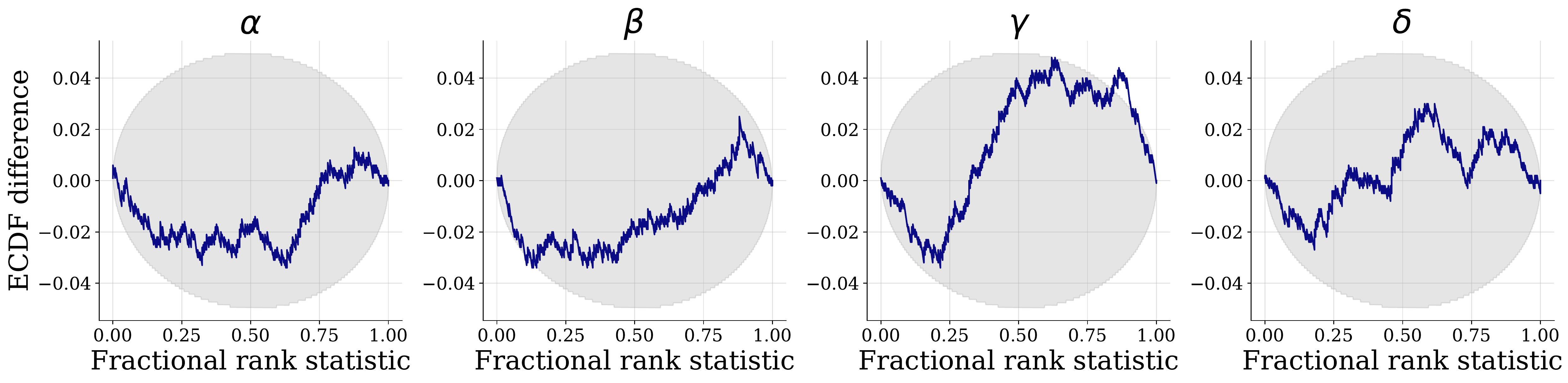}{Lotka-Volterra}

\FloatBarrier
\clearpage
\subsection{Experiment 2: Two Moons}

\paragraph{Model details}
This experiment utilizes the two moons simulator from \textcite{greenberg2019automatic} -- not to be confused with the standard two moons data set used for unconditional estimation -- with the same experimental setup as described in \textcite{wiqvist2021sequential}.

\paragraph{Network and training details}

The posterior network is a neural spline flow with $4$ coupling layers and a Gaussian latent space.
The likelihood network uses an interleaved coupling architecture with $5$ coupling layers.
We train the networks in an offline fashion on the respective simulation budget ($2\,000$, $6\,000$ and $10\,000$ simulations) for $64$ epochs with a batch size of $32$ and a learning rate of $0.0005$. 

The wall-clock times on a consumer-grade CPU are listed in \autoref{tab:tm_wallclock}. 
While the JANA implementation in the BayesFlow framework would certainly benefit from GPU acceleration, the available implementations of SNPLA and SNVI do not come with GPU support out-of-the-box due to their APIs to dependent packages (i.e., issues with Pyro for SNVI and issues with PyTorch for SNPLA).
We repeat the training phase of each method $10$ times to further investigate the reliability of the methods.
We only conducted one repetition with SNL due to the prohibitively slow run time (see \autoref{tab:tm_wallclock}).

\begin{table}[ht]
	\centering
	\begin{tabular}{l||c|c|c|c|c|c|c}
		& NPE-C & SNPE-C & SNRE-B & SNL & SNVI & SNPLA & JANA \\
		\hline
		\textbf{Training (seconds)}  & 229 & 1151 & 5533 & 17492 & 198 & 496 & 435 \\
		\textbf{Posterior Inference (seconds)}  & 0.02 & 0.02 & 592.63 & 1890 & 0.60 & 0.01 & 0.13 \\
		\textbf{Posterior Predictive Inference (seconds)}  & --- & --- & --- & 1872 & 0.61 & 0.03 & 0.27 \\
	\end{tabular}
	\caption{Average wall-clock times (seconds) on a consumer-grade CPU for different neural methods. Training time is based on offline learning with $10\,000$ simulations. Posterior inference indicates wall-clock time for obtaining $1\,000$ samples from the approximate posterior on a single observation. Posterior predictive inference indicates wall-clock time for obtaining $1\,000$ samples from the approximate posterior and evaluating the approximate likelihood of each sample. Note, that NPE-C and JANA are \textit{amortized}, so no further training is needed for applications on new observations.}
	\label{tab:tm_wallclock}
\end{table}

\begin{figure}
	\centering
	\twomoonsposterior{Repetition \#1 (main paper)}{plots/two_moons_1/tm_posterior_seed1.pdf}
	\hspace*{1cm}
	\twomoonsposterior{Repetition \#2}{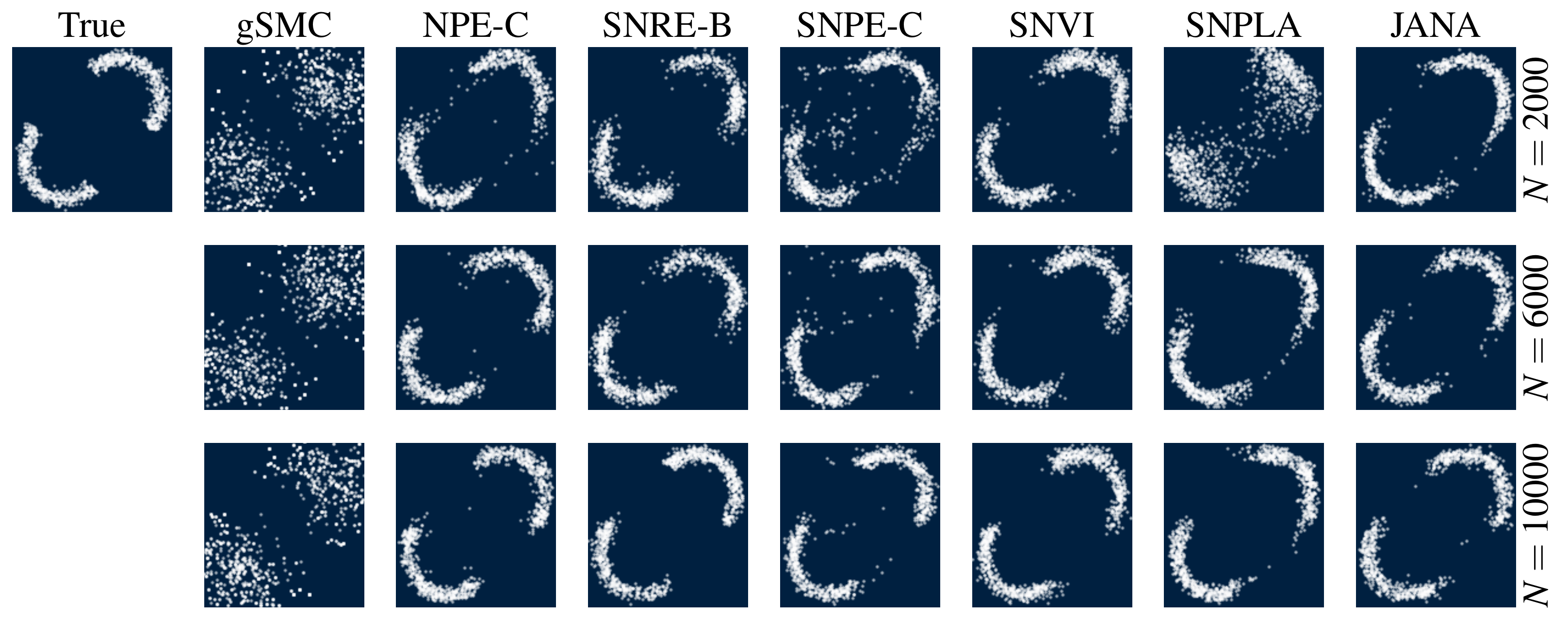}
	\\
	\twomoonsposterior{Repetition \#3}{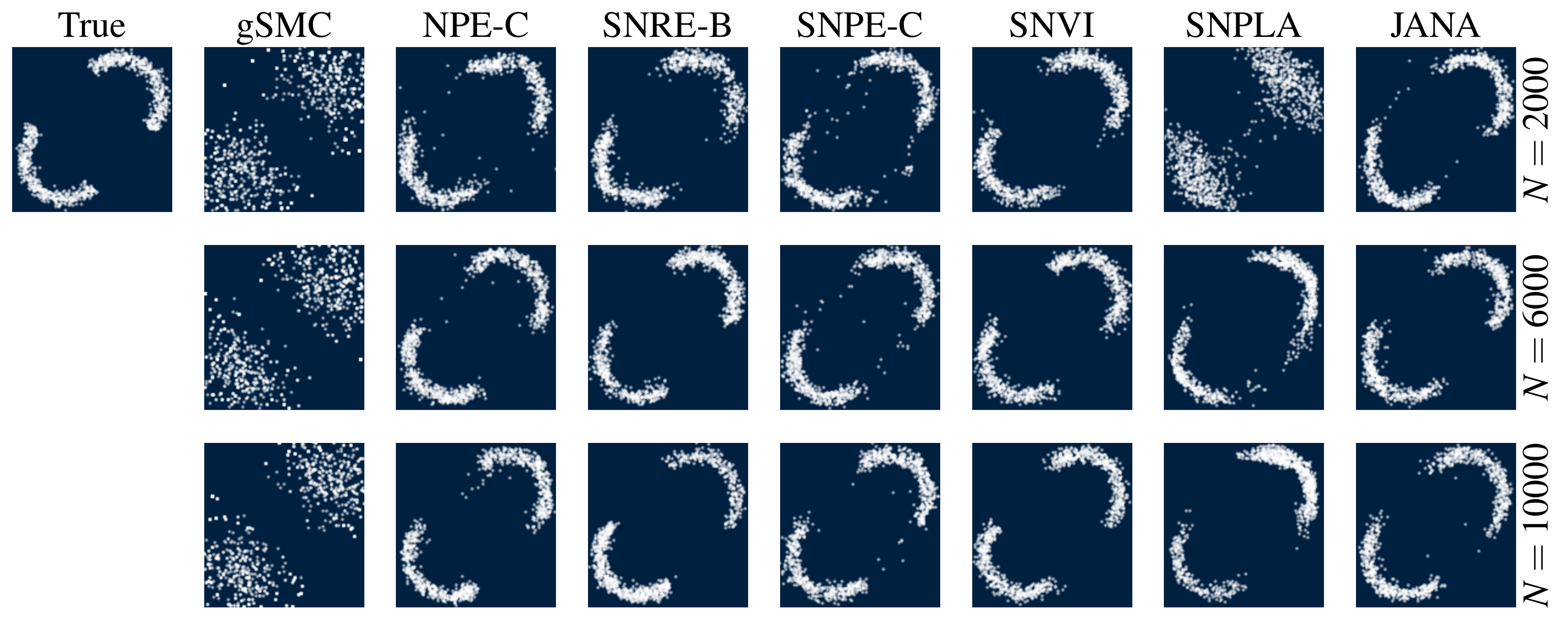}
	\hspace*{1cm}
	\twomoonsposterior{Repetition \#4}{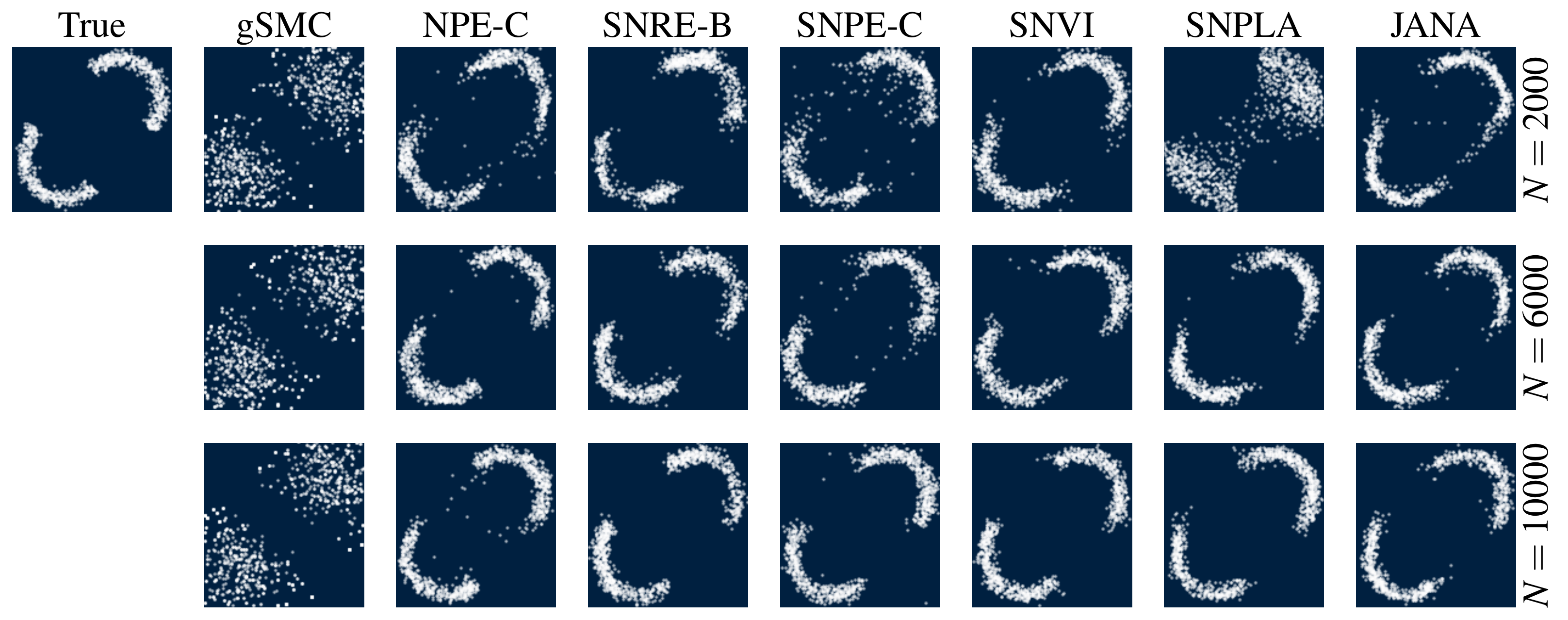}
	\\
	\twomoonsposterior{Repetition \#5}{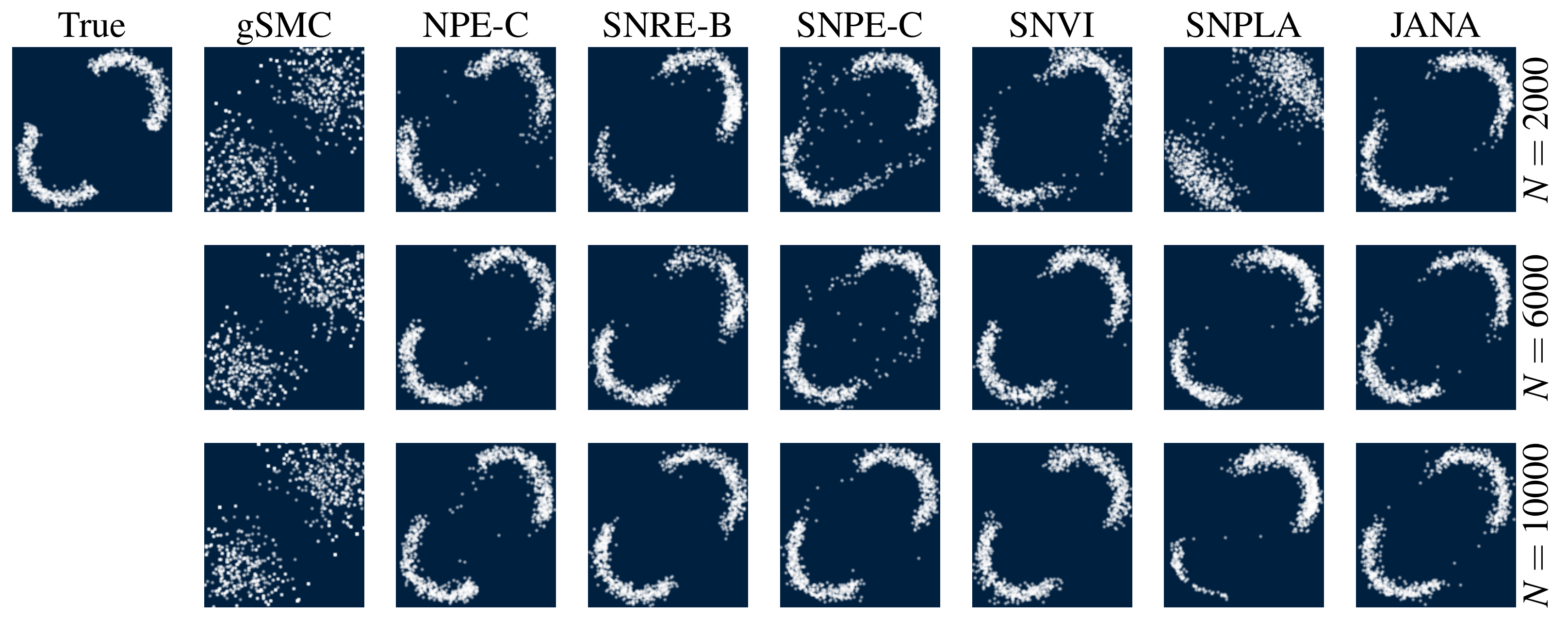}
	\hspace*{1cm}
	\twomoonsposterior{Repetition \#6}{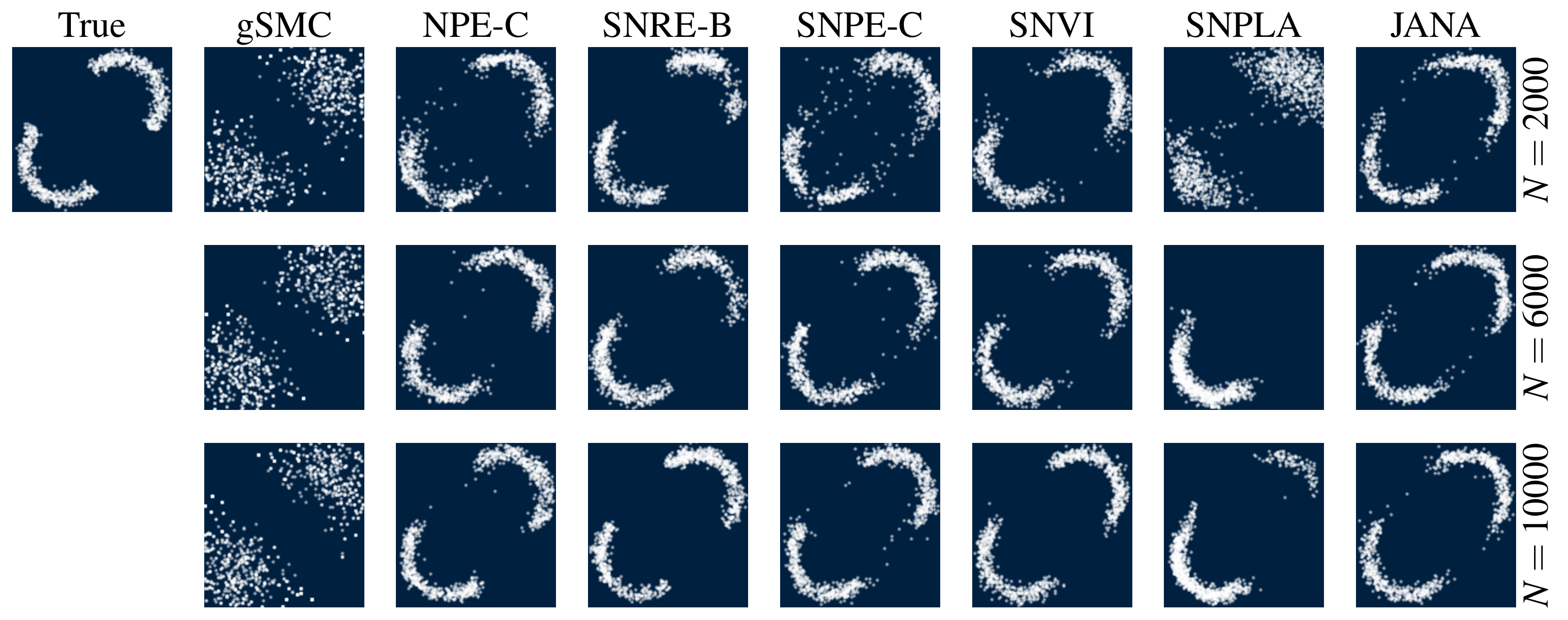}
	\\
	\twomoonsposterior{Repetition \#7}{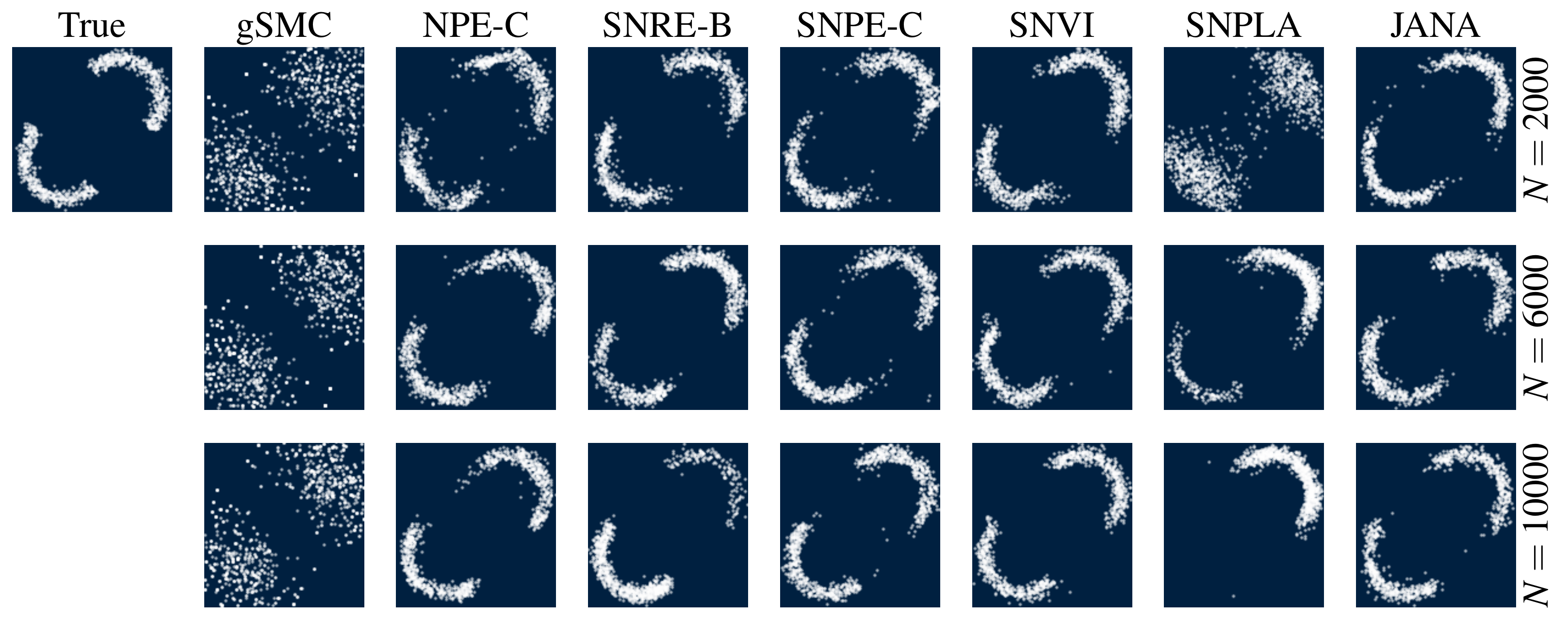}
	\hspace*{1cm}
	\twomoonsposterior{Repetition \#8}{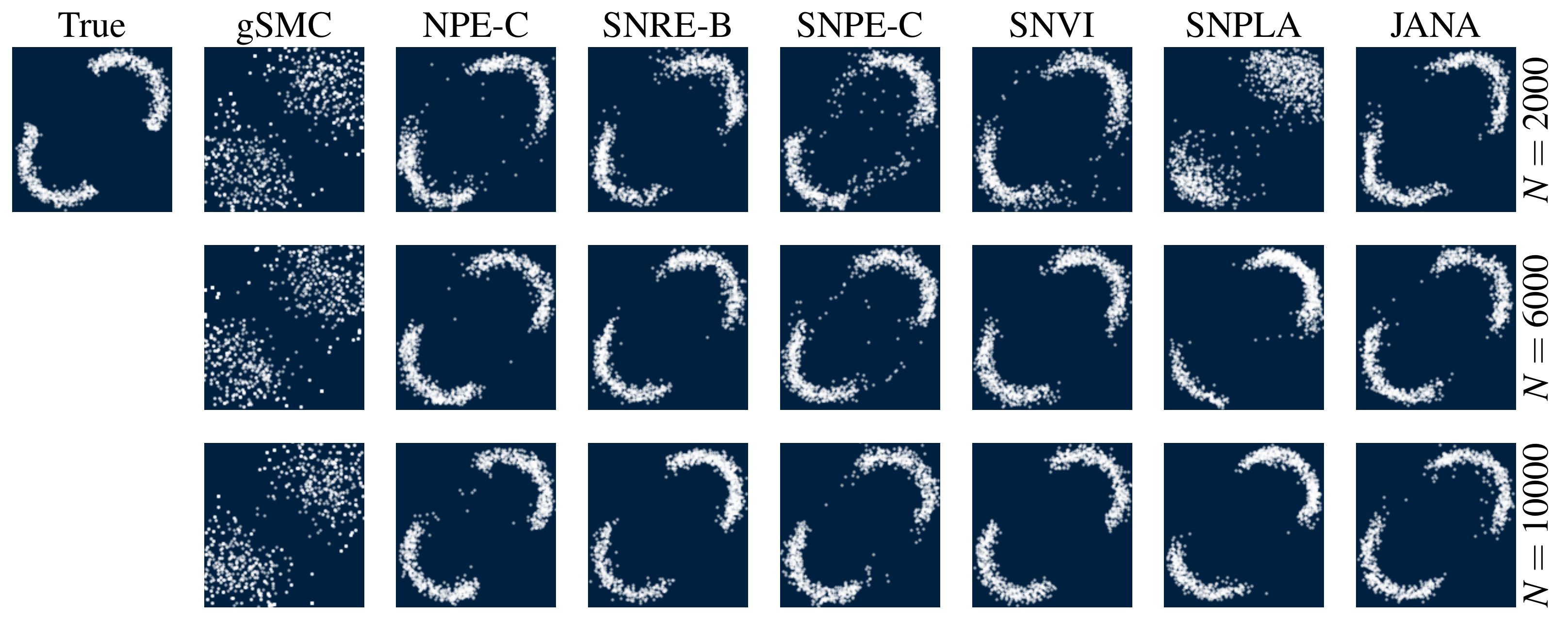}
	\\
	\twomoonsposterior{Repetition \#9}{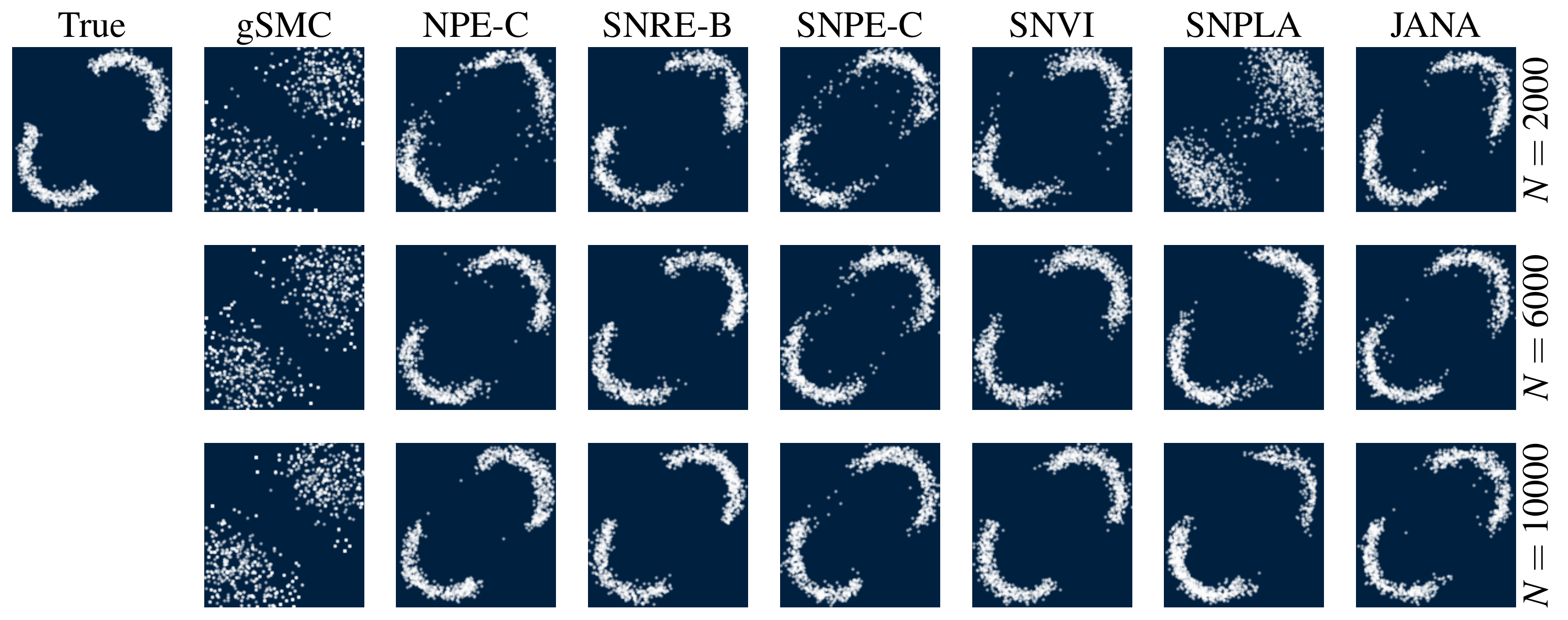}
	\hspace*{1cm}
	\twomoonsposterior{Repetition \#10}{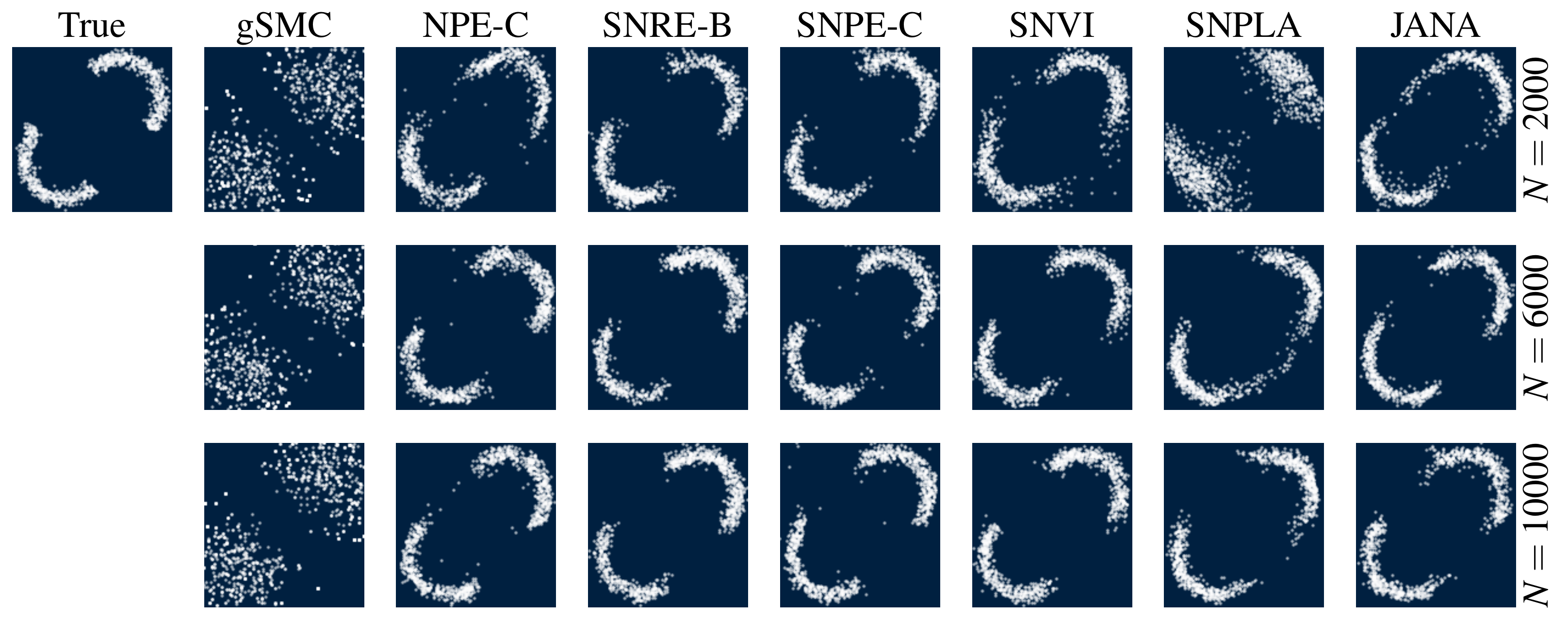}
	\caption{\textbf{Experiment 2.} $1\,000$ Posterior draws for all methods and repetitions of the experiment. The main paper shows repetition \#1, which is in line with the other runs. The axis limits represent the support of the uniform prior distribution.}
\end{figure}

\FloatBarrier
\paragraph{Different Simulator}

We repeat the experiment with the simulator from \textcite{lueckmann2021benchmarking}, which produces smaller moons with larger relative distance.
The only difference to \textcite{lueckmann2021benchmarking} is that we use a broader uniform prior with bounds $[-2, 2]$ (instead of $[-1,1]$) to further increase the difficulty of the task.
The results are largely equivalent to the ones reported in the main text.

\begin{figure}[H]
	\centering
	\twomoonsposterior{Repetition \#1 (main paper)}{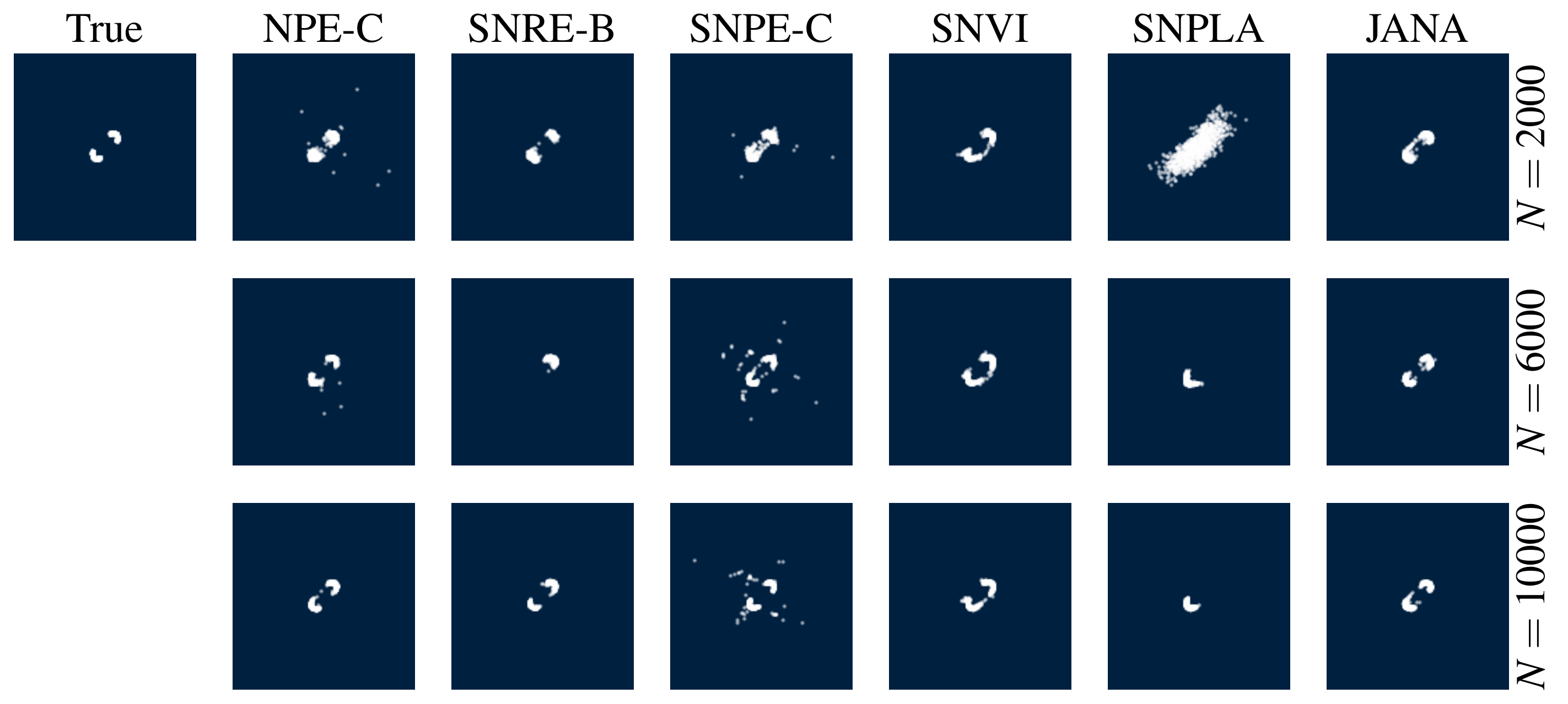}
	\hspace*{1cm}
	\twomoonsposterior{Repetition \#2}{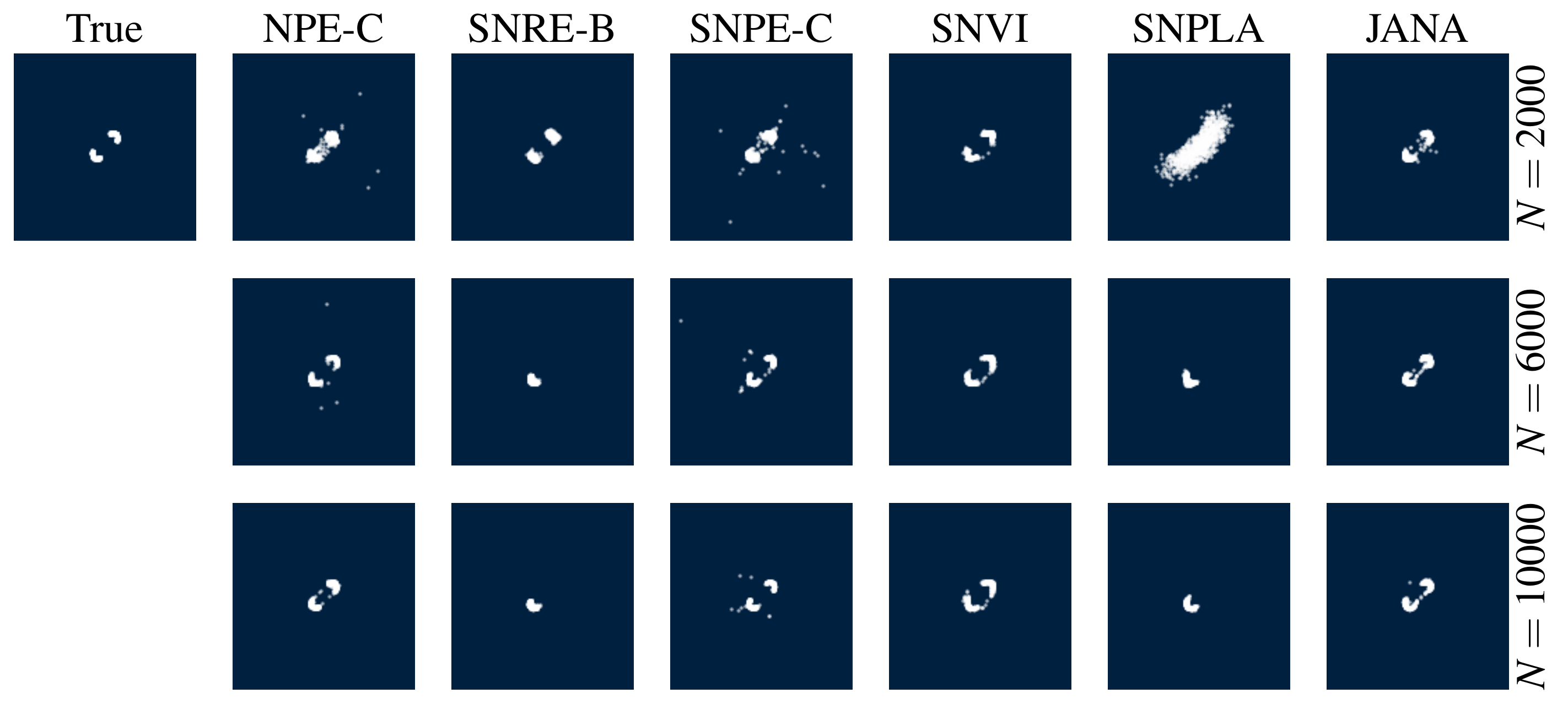}
	\\
	\twomoonsposterior{Repetition \#3}{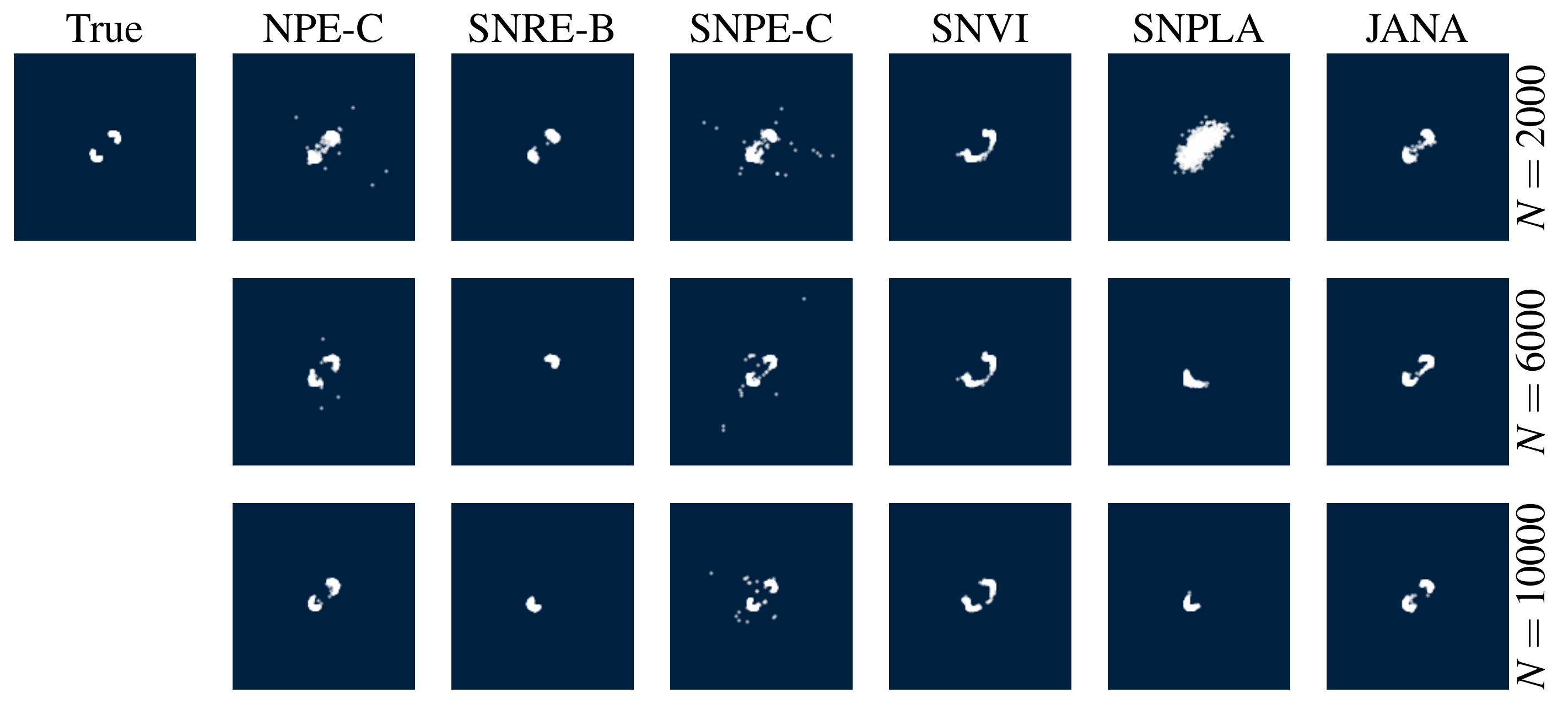}
	\hspace*{1cm}
	\twomoonsposterior{Repetition \#4}{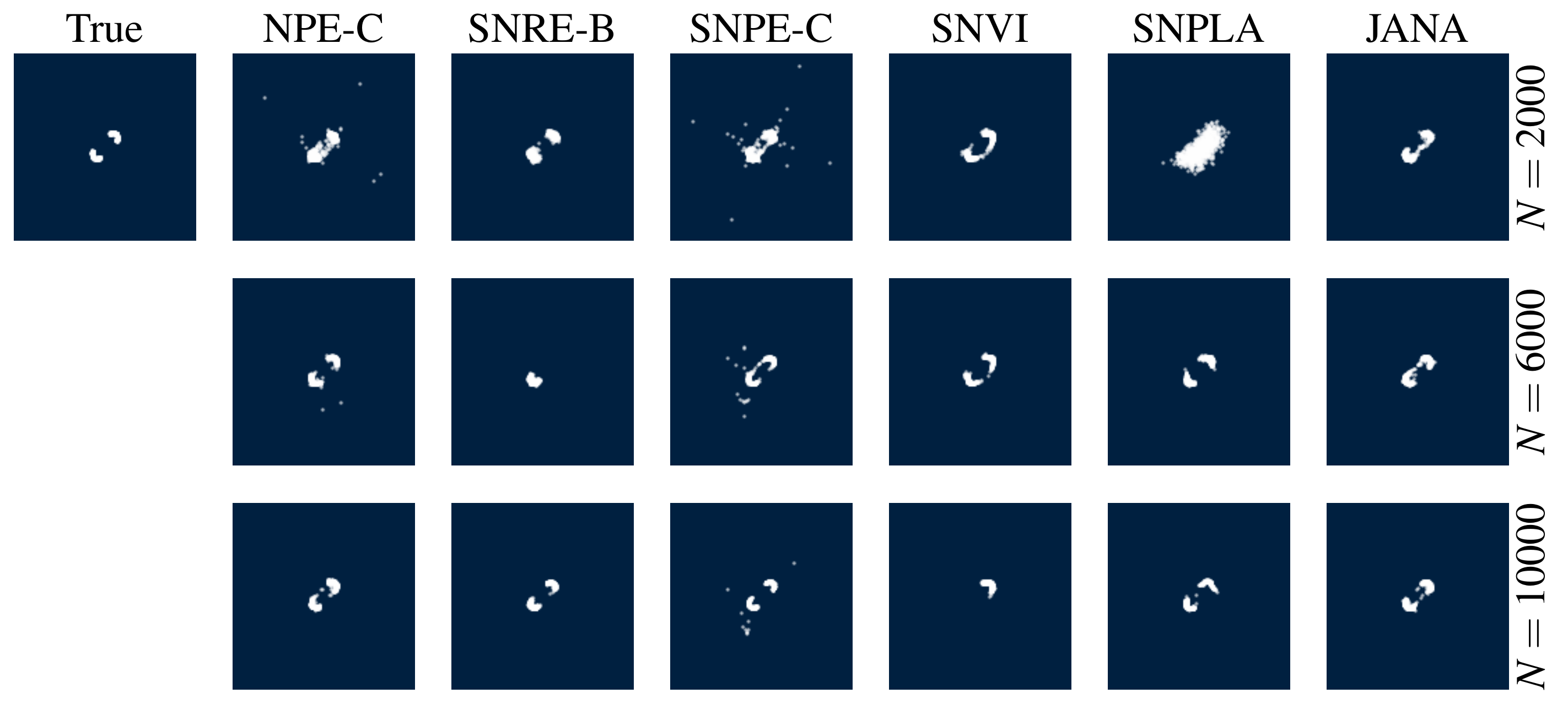}
	\\
	\twomoonsposterior{Repetition \#5}{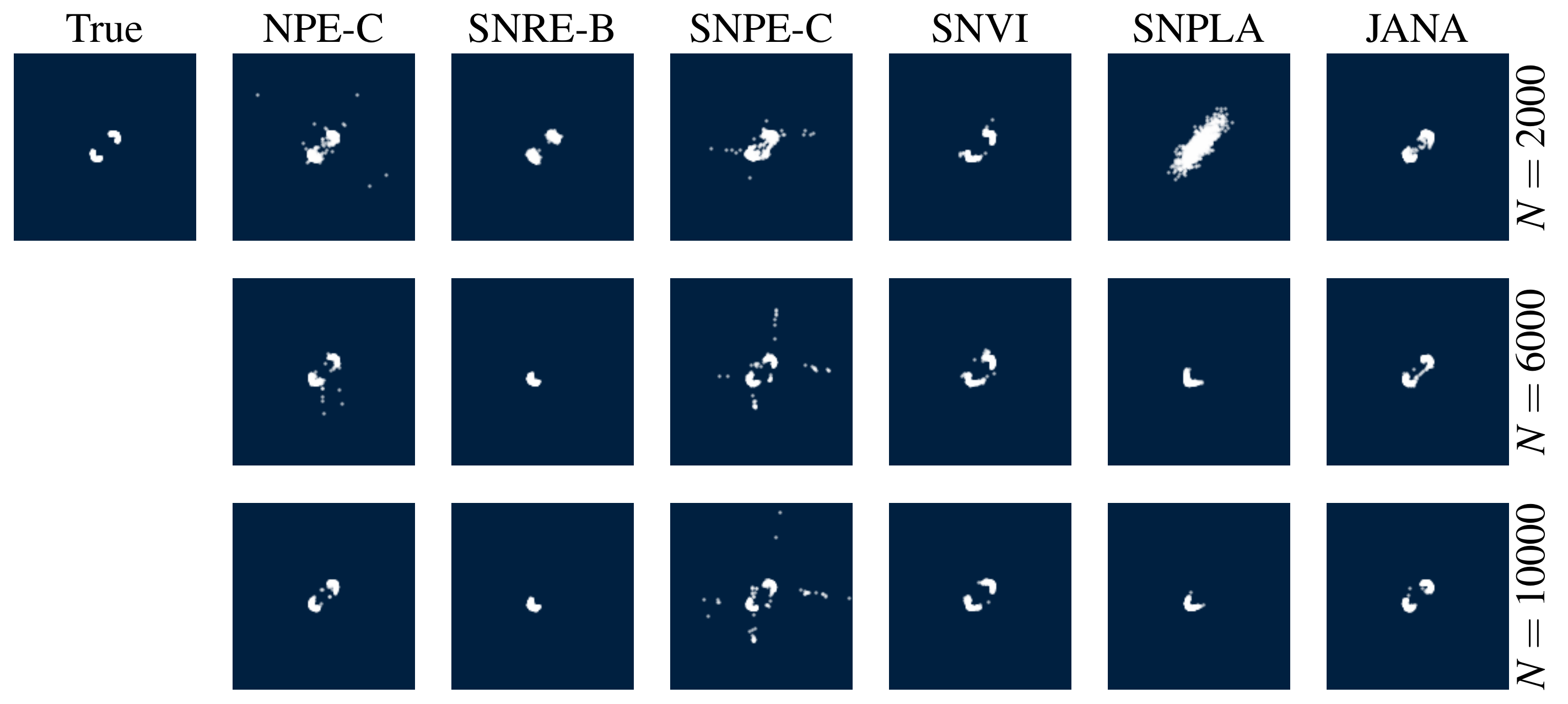}
	\hspace*{1cm}
	\twomoonsposterior{Repetition \#6}{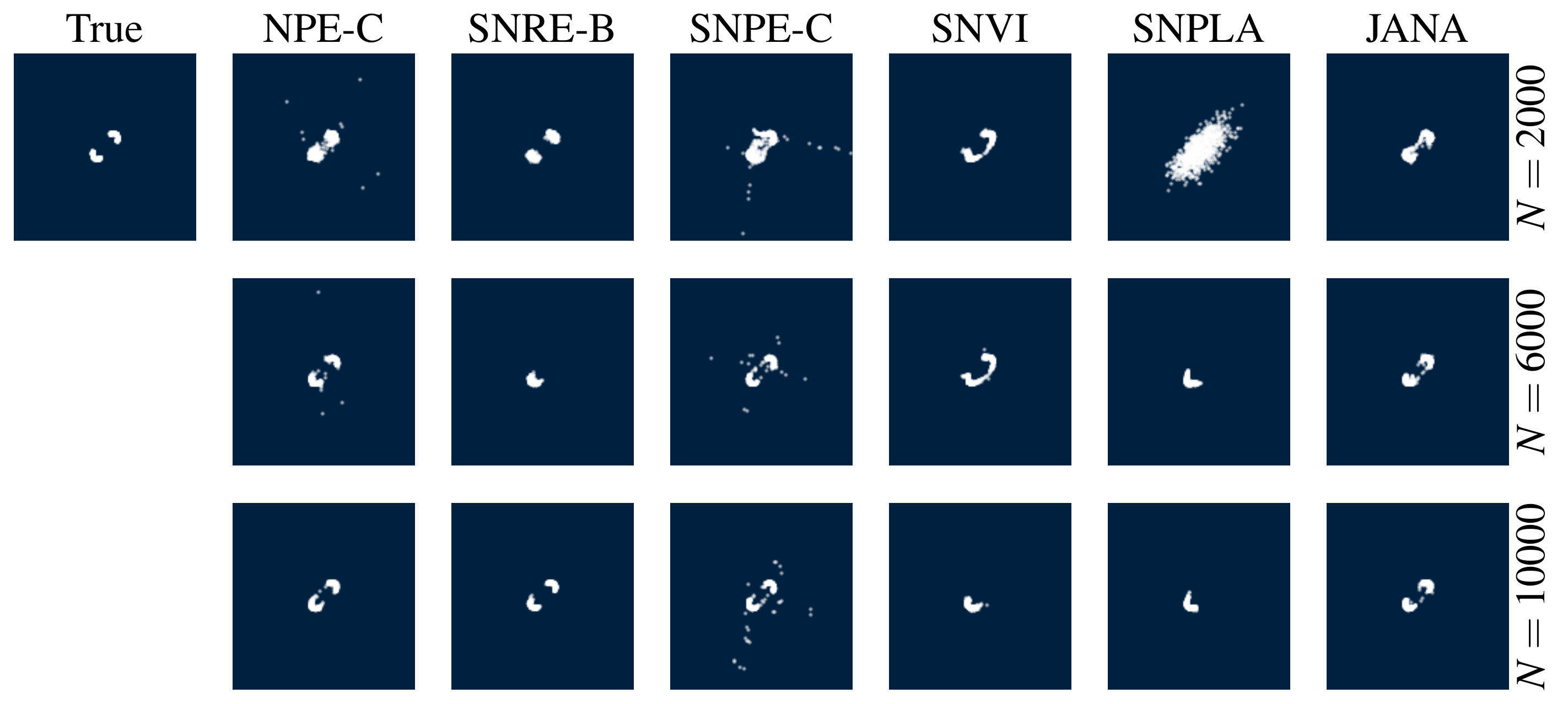}
	\\
	\twomoonsposterior{Repetition \#7}{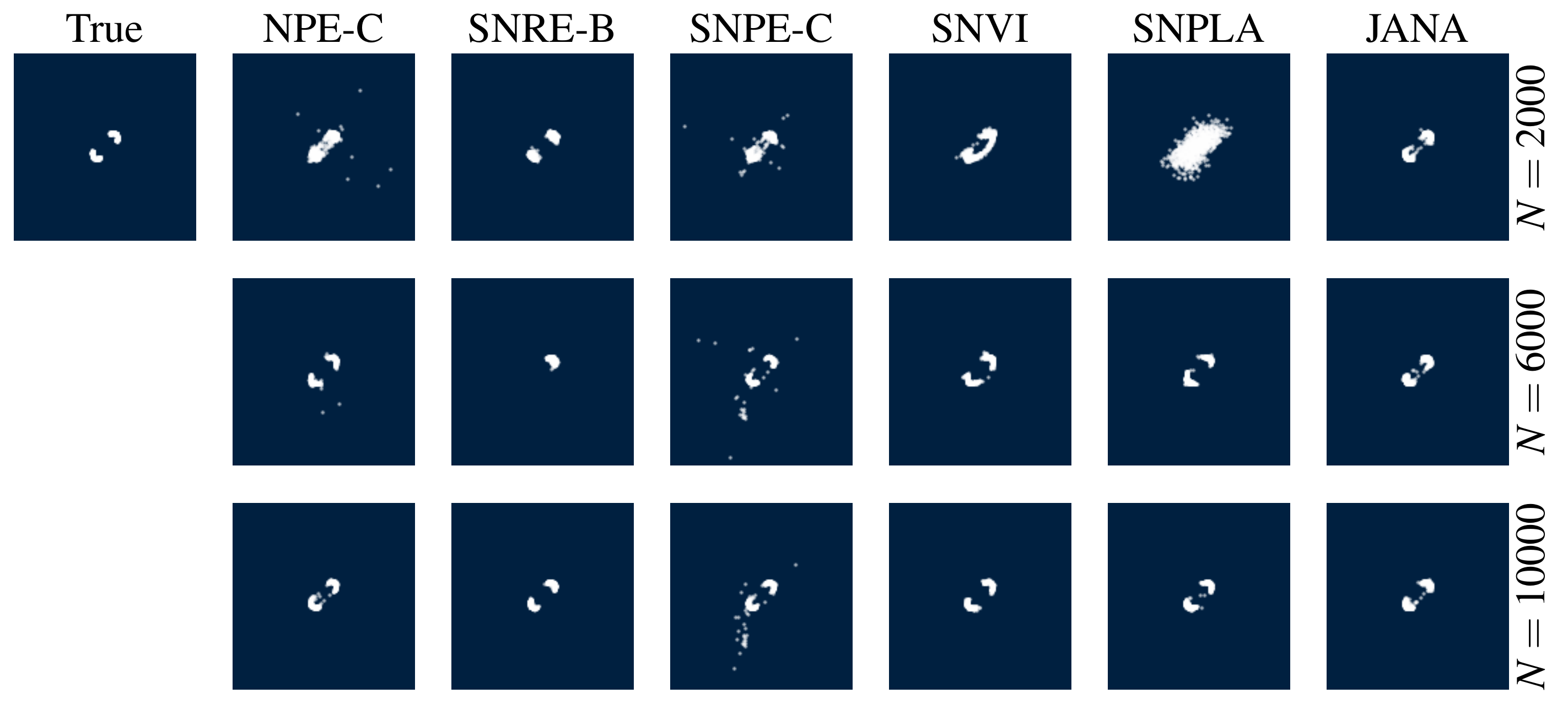}
	\hspace*{1cm}
	\twomoonsposterior{Repetition \#8}{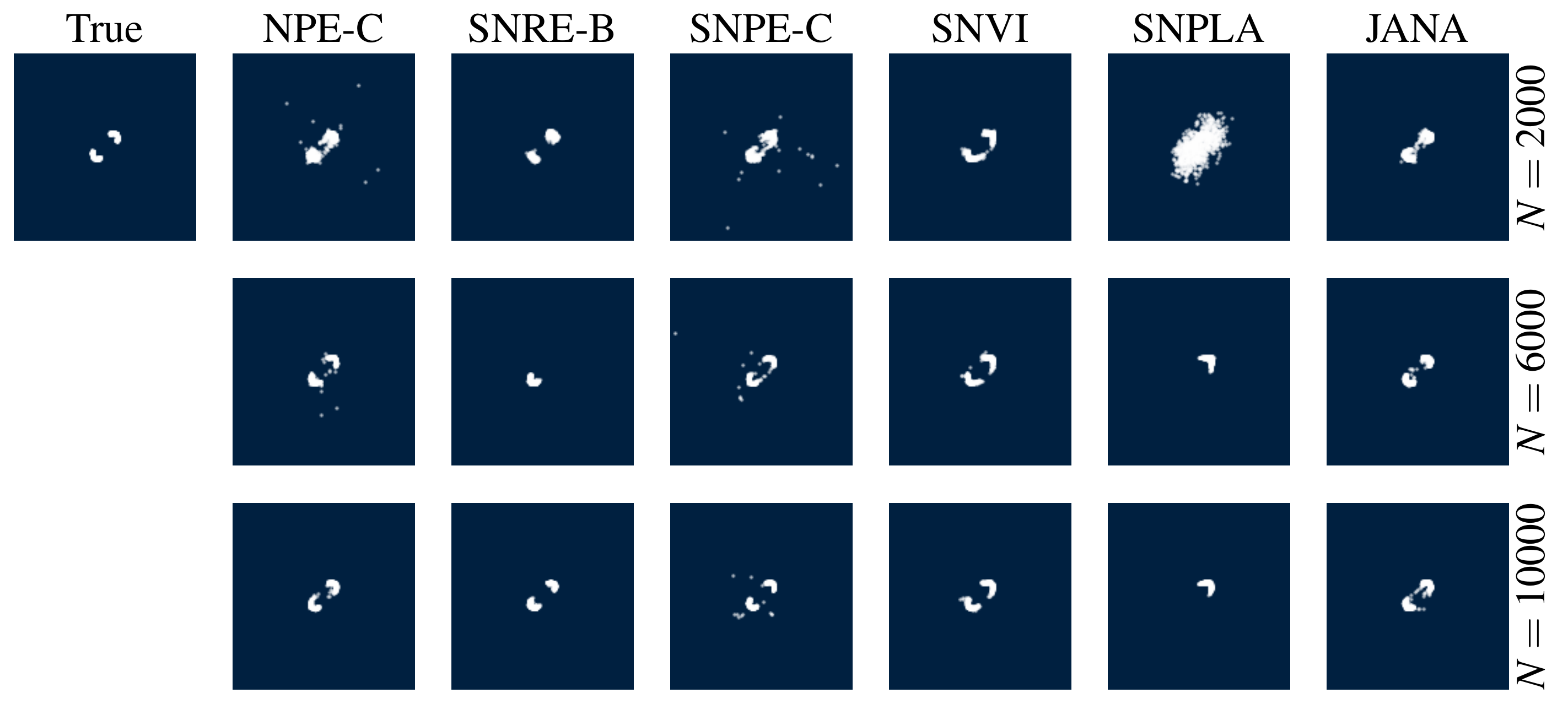}
	\\
	\twomoonsposterior{Repetition \#9}{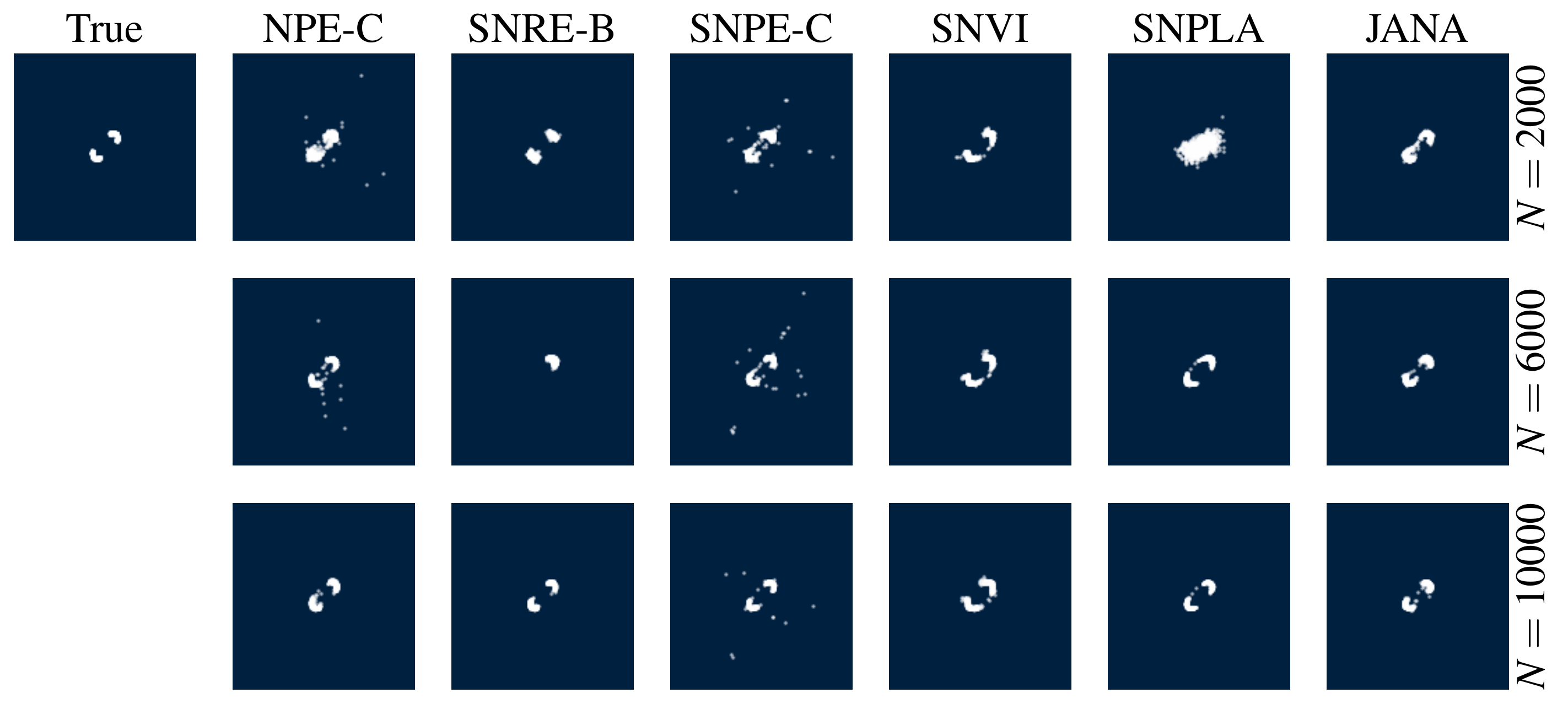}
	\hspace*{1cm}
	\twomoonsposterior{Repetition \#10}{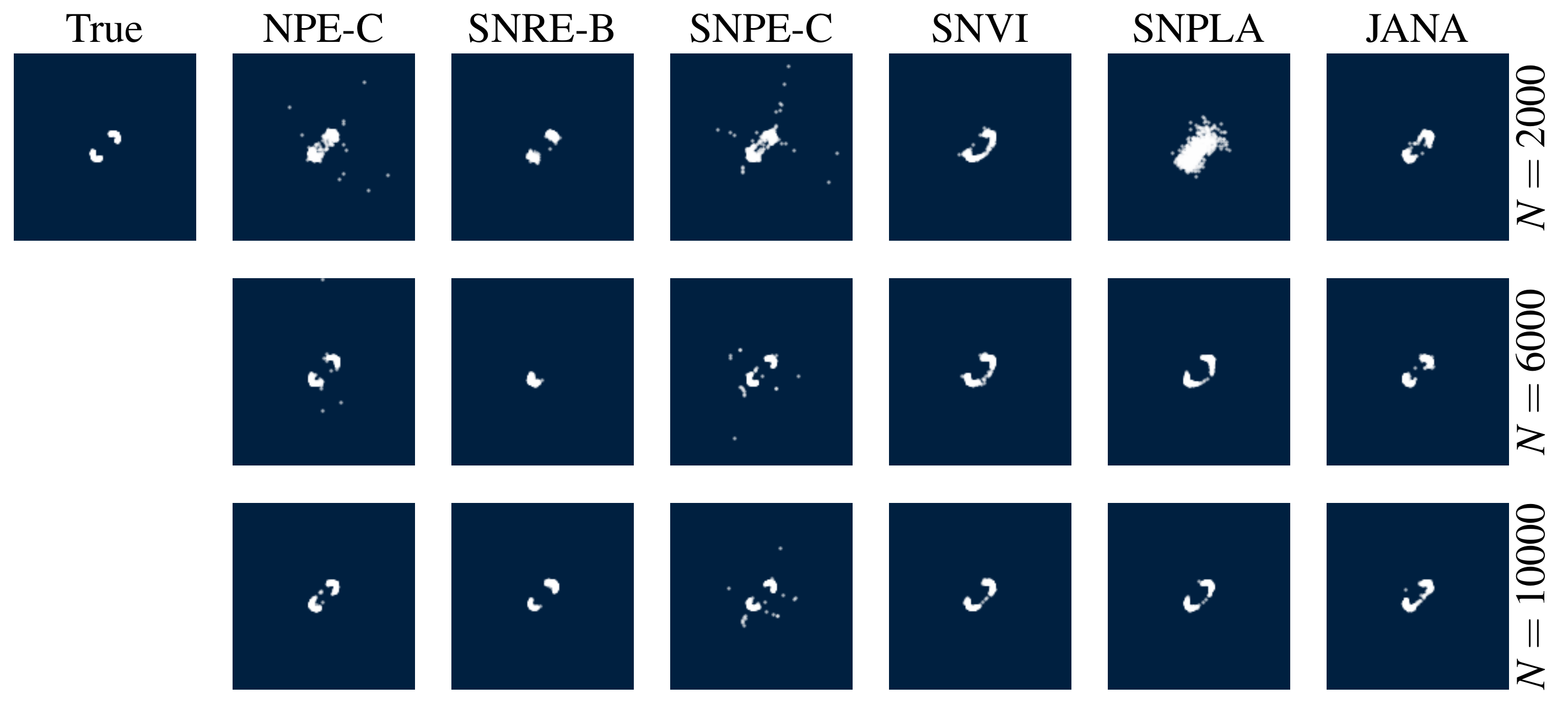}
	\caption{\textbf{Experiment 2.} $1\,000$ Posterior draws for all methods and repetitions of the experiment with a more challenging simulator. The axis limits represent the support of the uniform prior distribution.}
\end{figure}

\FloatBarrier
\clearpage
\subsection{Experiment 3: Exchangeable Diffusion Model}

\paragraph{Model details}
We focus on the drift diffusion model (DDM)---a cognitive model describing reaction times (RTs) in binary decision tasks \autocite{ratcliff2008}.
The DDM assumes that perceptual information for a choice alternative accumulates continuously according to a Wiener diffusion process. 
The change in information accumulation $\mathrm{d}x$ follows a random walk with drift and Gaussian noise:
\begin{equation}
	\mathrm{d}x = v\mathrm{d}t + \xi \sqrt{\mathrm{d}t}\quad\text{with} \quad\xi\sim\mathcal{N}(0, 1).
\end{equation}

The model consists of four parameters: drift-rate $v$, boundary separation $a$, non-decision time $t_0$ and bias (relative starting point) $w$. 
The model has the particularity of being very sensible to early outliers, as all reaction times smaller than the non-decision time are considered impossible (i.e., have a likelihood of zero).
We employ the simple DDM, as its likelihood function is tractable \autocite{voss2007fast}, and place truncated normal priors over the parameters $\thetab = (v, a, t_0, w)$,
\begin{equation}
	v \sim \mathcal{TN}_{[-5,5]}(0, 10),\quad a \sim \mathcal{TN}_{[0.5,3]}(1, 1),\quad t_0 \sim \mathcal{TN}_{[0.2,1]}(0.4, 0.2),\quad w \sim \mathcal{TN}_{[0.3,0.7]}(0.5, 0.1),
\end{equation}

where $\mathcal{TN}_{[a, b]}(\mu, \sigma)$ denotes the truncated normal distribution with location $\mu$ and standard deviation $\sigma$ truncated within the interval $[a, b]$.
The summary network is a permutation-invariant network which reduces simulated IID RT data sets to $S=10$ summary statistics \autocite{radev2020bayesflow}.

\paragraph{Network and training details}

The summary network is a deep permutation-invariant network with $2$ equivariant modules followed by an invariant module \autocite{radev2020bayesflow, bloem2020probabilistic}.
The summary network reduces the IID RT data sets into $10$-dimensional learned summary statistics.
The posterior network is a conditional invertible neural network (cINN) with $5$ conditional coupling layers and a Student-$t$ latent space ($df = 50$).
The internal networks of the coupling layers are fully connected (FC) networks with $2$ hidden layers featuring $128$ units and \texttt{tanh} activation function.

The likelihood network is a cINN with $12$ conditional coupling layers, with smaller internal FC networks of $2$ hidden layers having $32$ units each, a \texttt{tanh} activation function, and a Student-$t$ latent space.
We train the networks in an offline fashion.
The likelihood network is trained for 20 epochs with a batch size of $64$ and a learning rate of $0.001$.
The posterior network is trained for 100 epochs with a batch size of $64$ and a learning rate of $0.002$.

\begin{figure}[H]
	\centering
	\begin{subfigure}[t]{0.20\linewidth}
		\includegraphics[width=\linewidth]{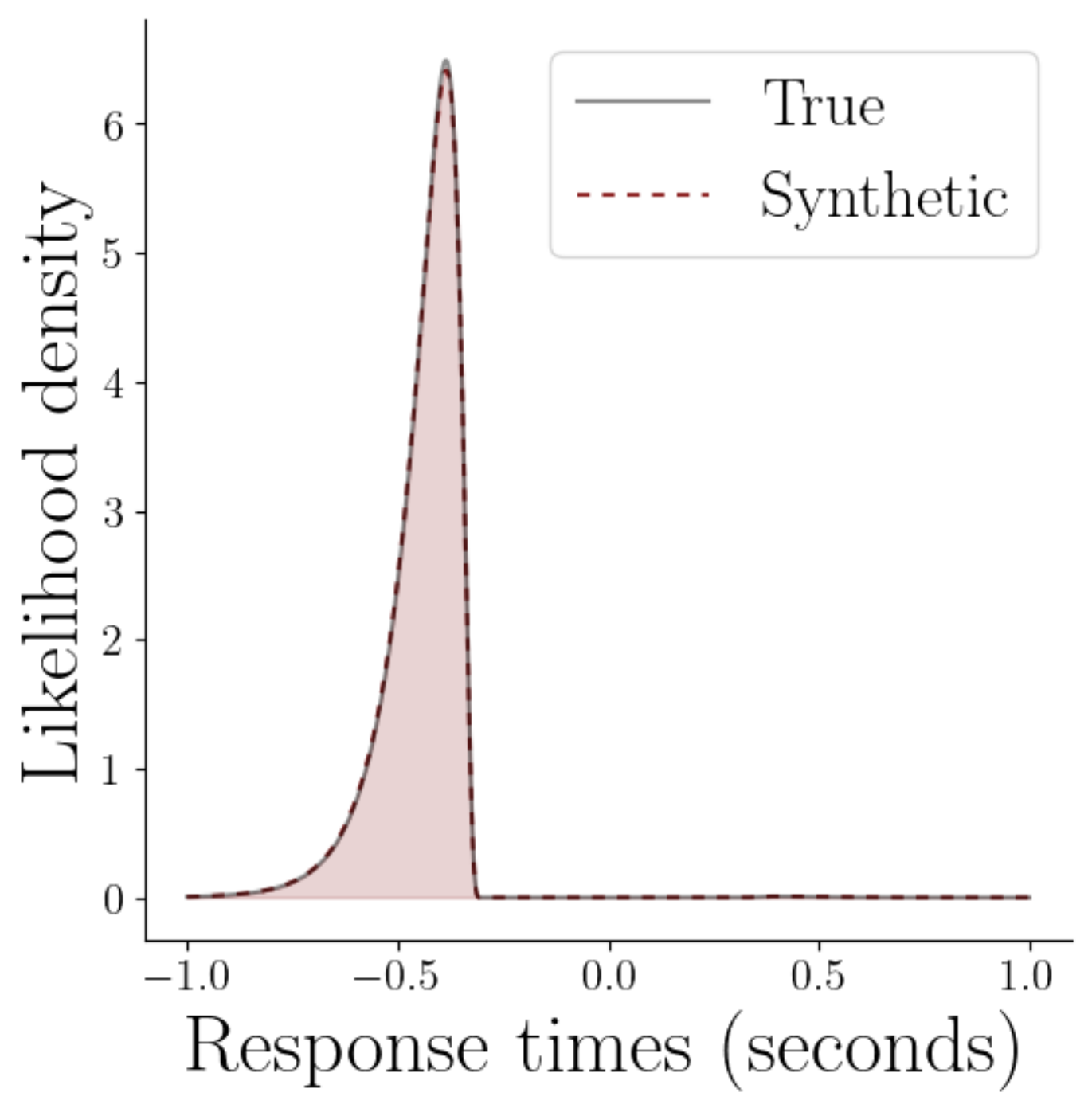}
	\end{subfigure}
	\hspace*{0.2cm}
	\begin{subfigure}[t]{0.20\linewidth}
		\includegraphics[width=\linewidth]{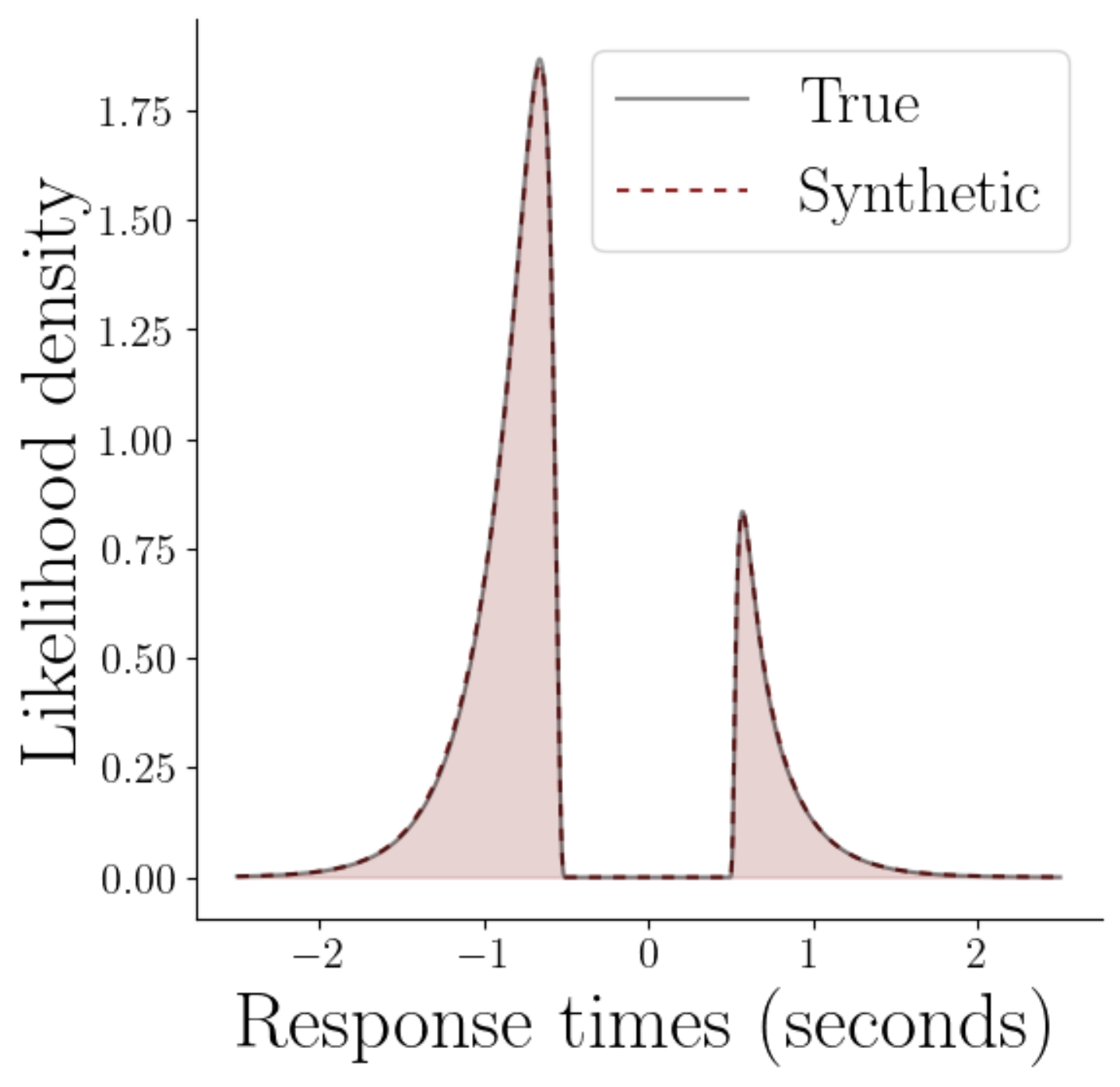}
	\end{subfigure}
	\hspace*{0.2cm}
	\begin{subfigure}[t]{0.20\linewidth}
		\includegraphics[width=\linewidth]{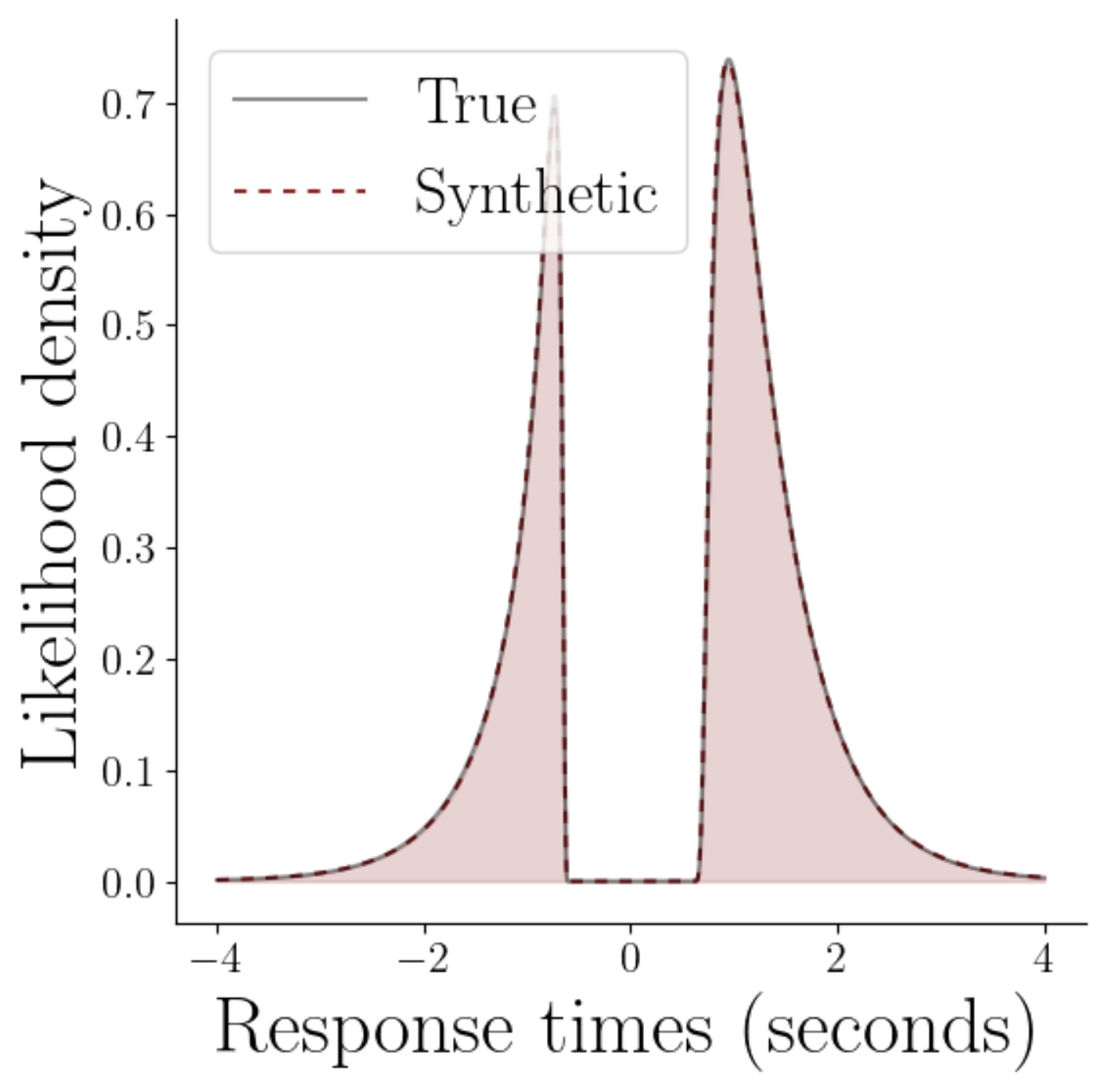}
	\end{subfigure}
	\hspace*{0.2cm}
	\begin{subfigure}[t]{0.20\linewidth}
		\includegraphics[width=\linewidth]{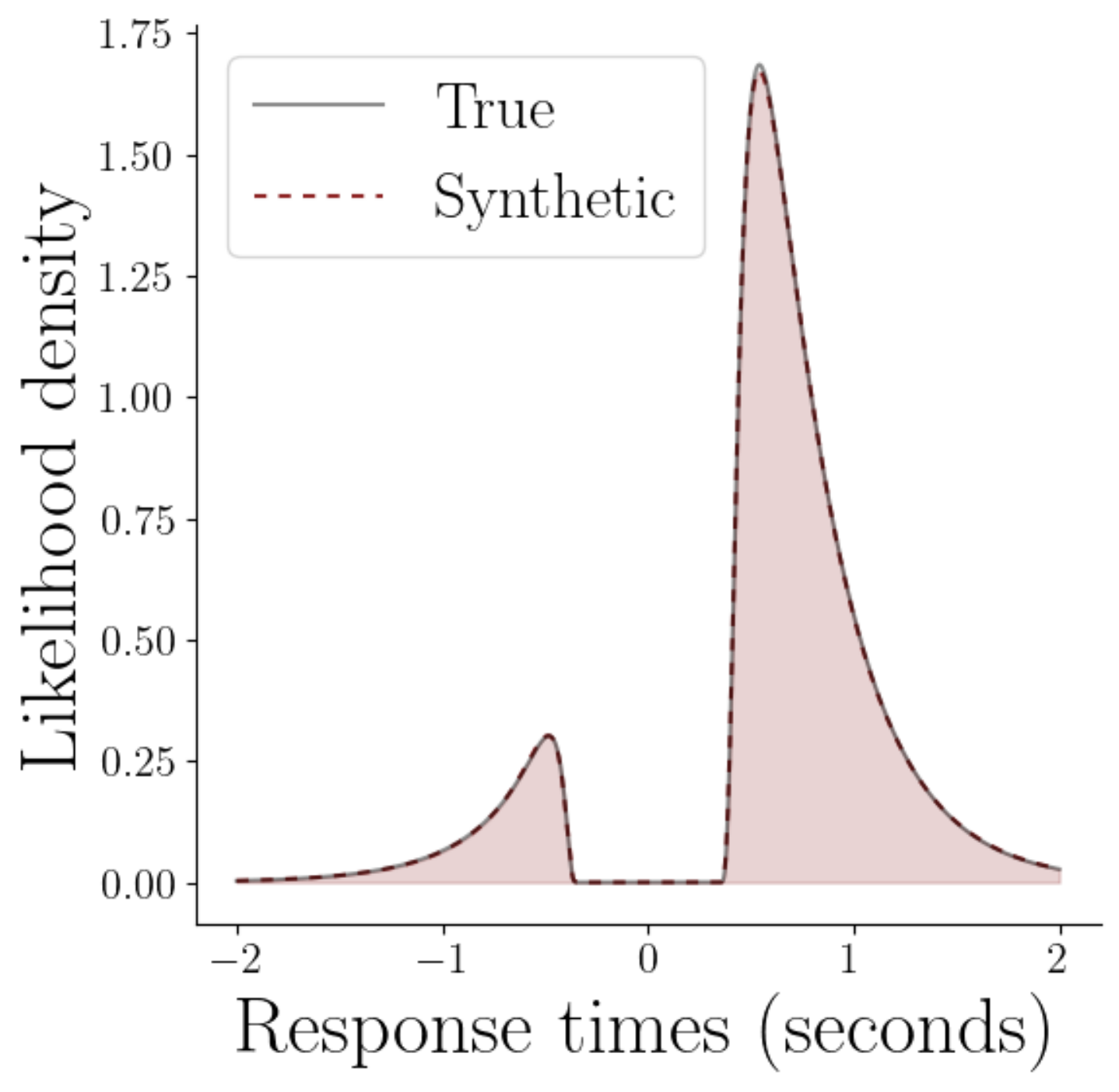}
	\end{subfigure}
	\caption{\textbf{Experiment 3.} JANA exhibits essentially perfect likelihood emulation for various parameter configurations.}
	\label{fig:app:ddm:likelihood}
\end{figure}

\begin{figure}[H]
	\centering
	\includegraphics[width=0.90\linewidth]{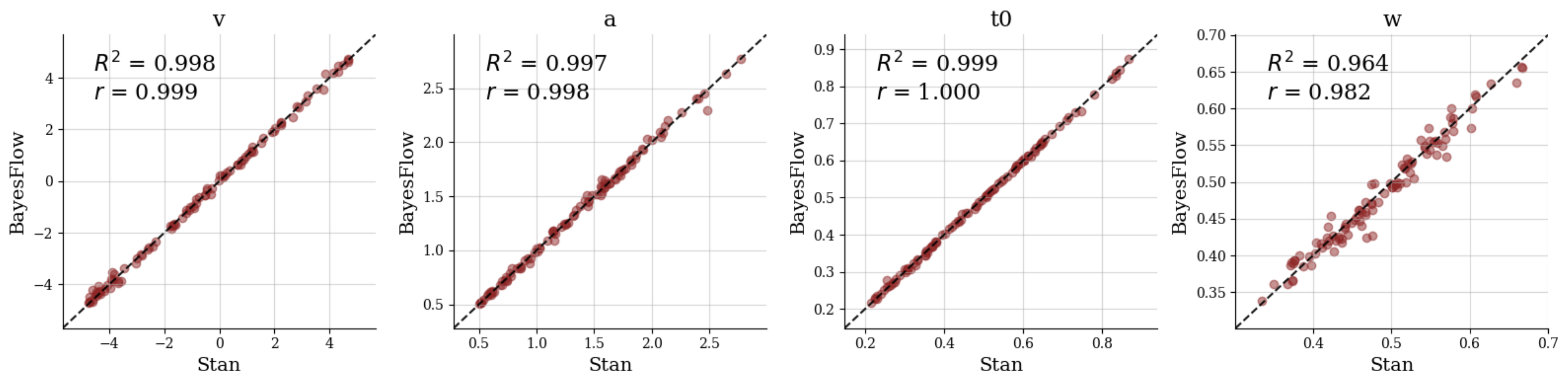}
	\caption{\textbf{Experiment 3.} The parameter recovery of JANA is largely identical to the estimates obtained via the gold-standard HMC-MCMC implementation in Stan.}
	\label{fig:app:ddm:stan-recovery}
\end{figure}


\FloatBarrier
\clearpage
\subsection{Experiment 4: Markovian Compartmental Model}

\paragraph{Model details} We use the model formulation from \textcite{radev2021amortized}, which consists of three components: 1) a latent sub-model, 2), an intervention sub-model; 3) and an observation sub-model.

First, the latent sub-model is a SIR-type system of non-linear ordinary differential equations (ODEs) with six population compartments representing the interactions between 
susceptible ($S$), exposed ($E$), infected ($I$), carrier ($C$), recovered ($R$), and dead ($D$) individuals.
The interaction dynamics are governed by:
\begin{align}
	\frac{dS}{dt} &= -\lambda (t)\,\left(\frac{C + \beta\,I}{N}\right)\,S \\
	\frac{dE}{dt} &= \lambda (t)\,\left(\frac{C + \beta\,I}{N}\right)\,S - \gamma\,E \\
	\frac{dC}{dt} &= \gamma\,E - (1 - \alpha)\,\eta\,C - \alpha\,\theta\,C\\
	\frac{dI}{dt} &= (1 - \alpha)\,\eta\,C - (1-\delta)\,\mu\,I - \delta\,d\,I \\
	\frac{dR}{dt} &= \alpha\,\theta\,C + (1-\delta)\,\mu\,I \\
	\frac{dD}{dt} &= \delta\,d\,I
\end{align}
For simulating the system, we use $dt = 1$ which corresponds to a time scale of days.

Second, an \textit{intervention sub-model} accounts for changes in the transmission rate $\lambda(t)$ due to non-pharmaceutical policies.
It defines three change points for $\lambda(t)$ encoding an assumed transmission rate reduction in response to intervention measures imposed by the German authorities in 2020.
Each change point is a piece-wise linear function with three parameters: the effect strength and the boundaries defining the time interval for the effect to take place \autocite{radev2021outbreakflow}.

The observation sub-model assumes that only compartments $I$, $R$, and $D$ are potentially observable.
Moreover, it accounts for the fact that officially reported cases might not represent the true latent numbers of an outbreak:
\begin{align}
	I^{(obs)}_t &= I^{(obs)}_{t-1} + (1 - f_I(t))\,(1 - \alpha)\,\eta\,C_{t-L_I} + \sqrt{I^{(obs)}_{t-1}} \,\sigma_I\,\xi_t \\ 
	R^{(obs)}_t &= R^{(obs)}_{t-1} + (1 - f_R(t))\,(1-\delta)\,\mu\,I_{t-L_R}  + \sqrt{R^{(obs)}_{t-1}}\,\sigma_R\,\xi_t \\
	D^{(obs)}_t &= D^{(obs)}_{t-1} + (1 - f_D(t))\, \delta\,d\,I_{t-L_D} + \sqrt{D^{(obs)}_{t-1}}\,\sigma_D\, \xi_t 
\end{align}
In the above equations, $L_I, L_R$, and $L_D$ denote the reporting delays (lags), and denote $\sigma_I, \sigma_R$, and $\sigma_D$ the scales of multiplicative reporting noise for the respective compartments. 
The noise variables $\xi_t$ follow a Student-\textit{t} distribution with 4 degrees of freedom. 
The weekly modulation of reporting coverage $f_{\mathcal{C}}(t)$ for each of the compartments $\mathcal{C} \in \{I, R, D\}$ is computed as follows:
\begin{align}
	f_{\mathcal{C}}(t) = (1 - A_{\mathcal{C}})\,\left(1 - \left| \sin \left( \frac{\pi}{7}t - 0.5\, \Phi_{\mathcal{C}} \right) \right|  \right)
\end{align}
This yields three additional unknown parameters for the weekly modulation amplitudes $A_I, A_R, A_D$, and phases $\Phi_I, \Phi_R, \Phi_D$, each. 

\paragraph{Network and training details}
The summary network is a combination of 1D convolutional and LSTM layers, which reduce the multivariate time series into a vector of $192$ learned summary statistics \autocite{radev2021outbreakflow}.
The posterior network is a conditional invertible neural network (cINN) with $6$ conditional affine coupling layers.
The internal networks of the coupling layers are fully connected (FC) networks with $2$ hidden layers of $128$ units and a \texttt{swish} activation function.
The likelihood network is a recurrent cINN with $8$ conditional coupling layers with the same structure as the coupling layers of the posterior network.
We use a gated recurrent unit (GRU) with $256$ hidden units for the internal recurrent memory.
We train the networks in an online fashion (i.e., on-the-fly simulations) for $100$ epochs with a batch size of $32$ and a learning rate of $0.0005$. 
This initial learning rate is reduced throughout the training phase following a cosine decay schedule with a minimum learning rate of $0$.

\begin{figure}
	\centering
	\includegraphics[width=0.9\linewidth]{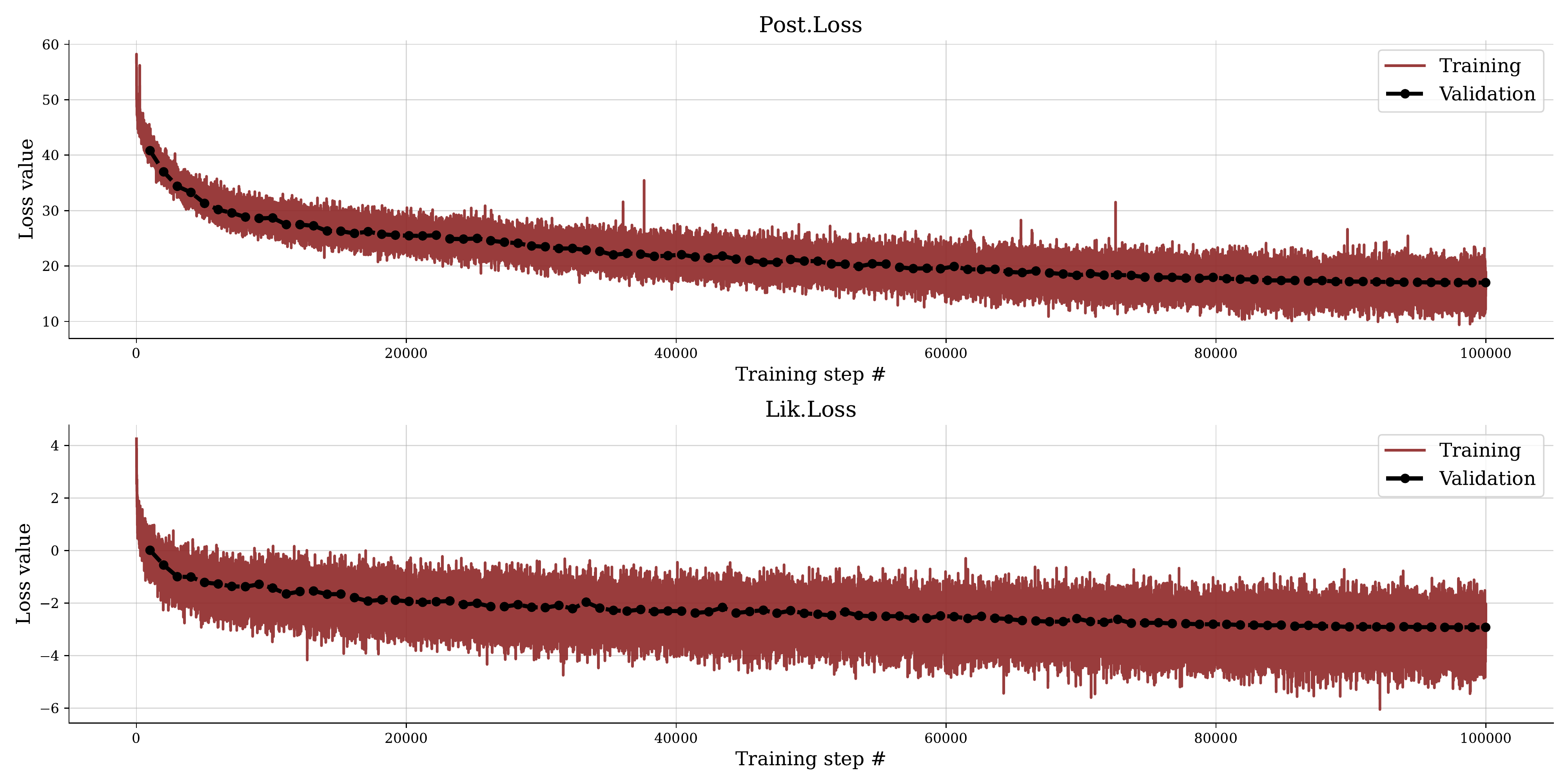}
	\caption{\textbf{Experiment 4.} Loss history}
	\label{fig:app:covid:loss}
\end{figure}

\begin{figure}
	\centering
	\covidscenario{Scenario I}{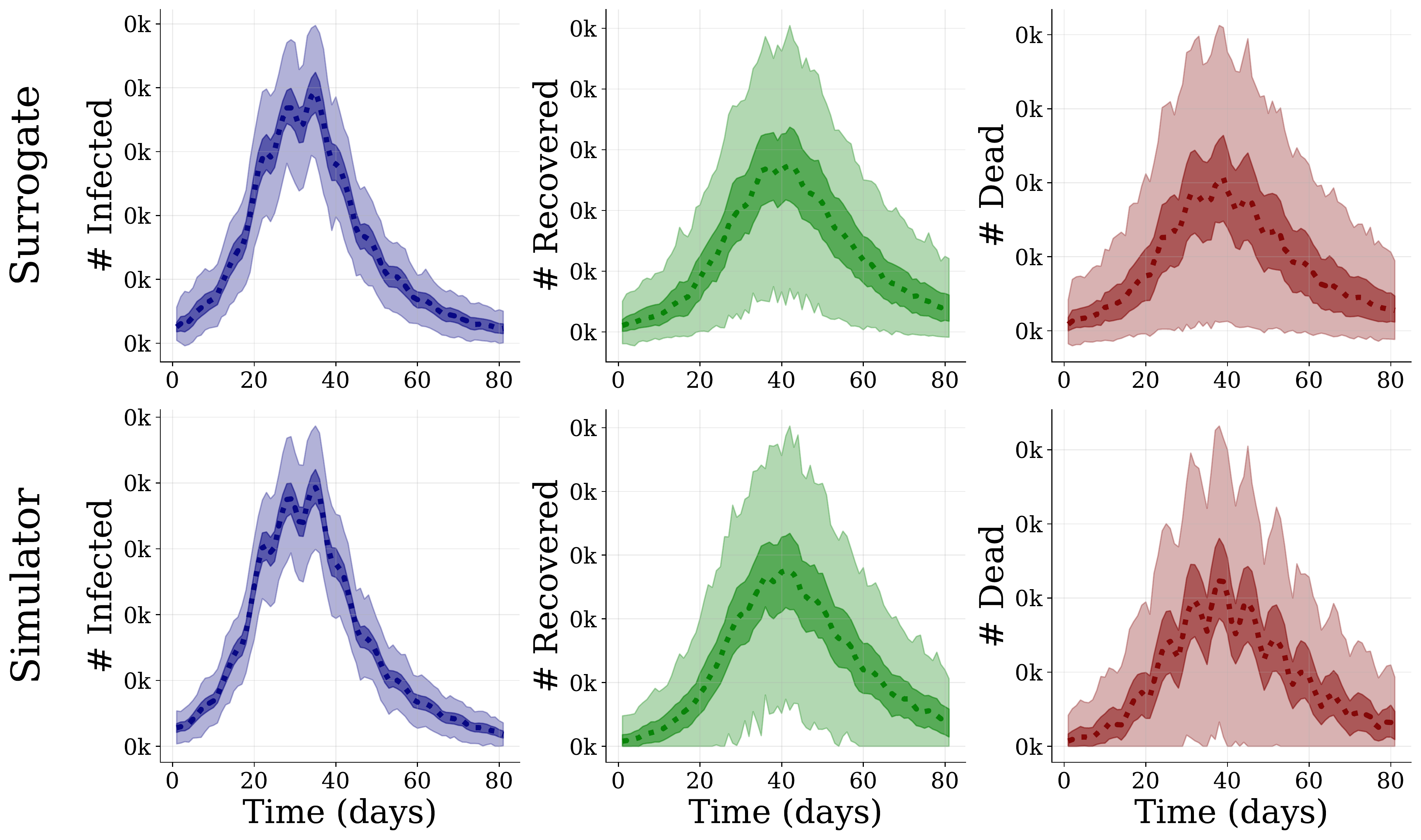}
	\hspace*{1cm}
	\covidscenario{Scenario II}{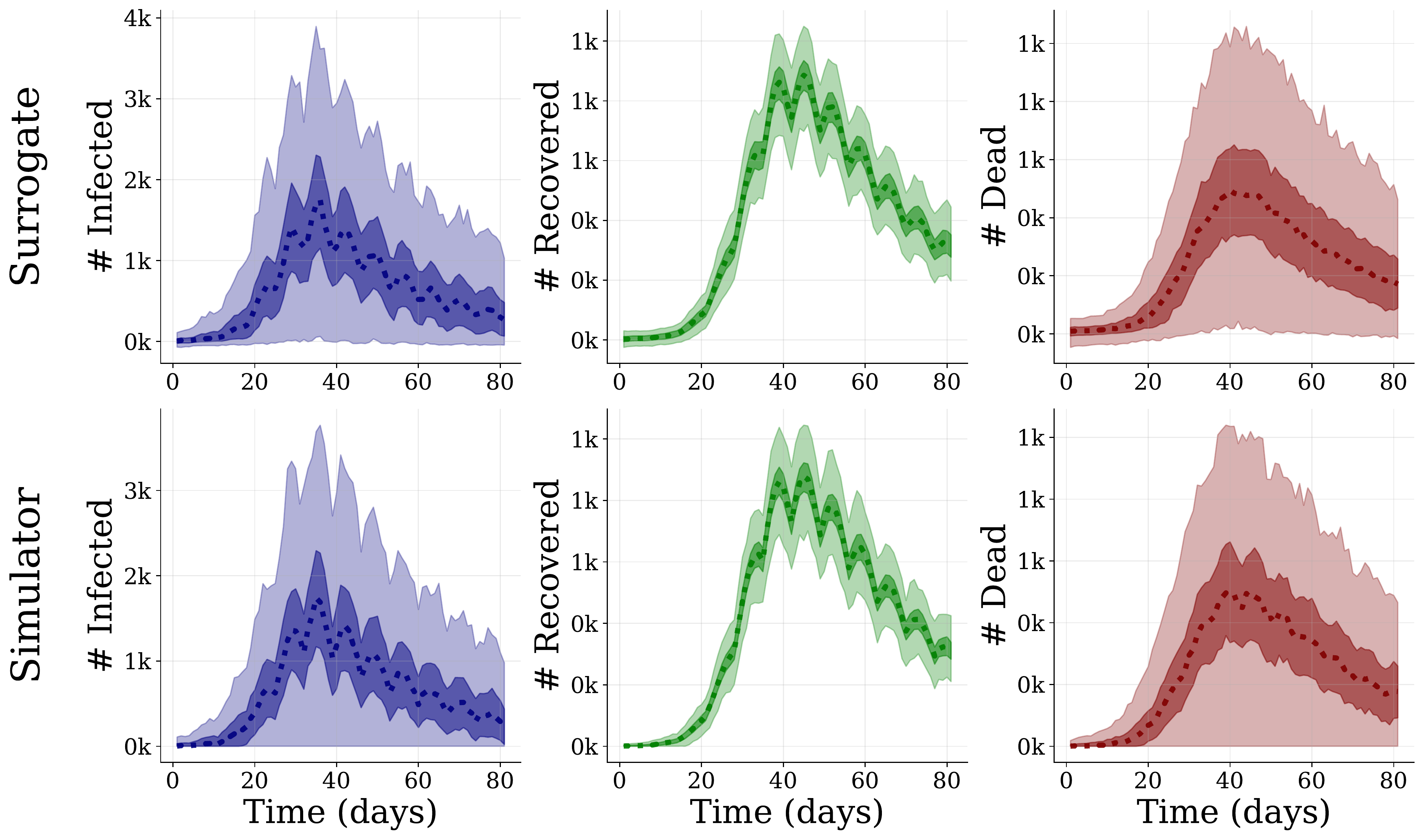}
	\\
	\covidscenario{Scenario III}{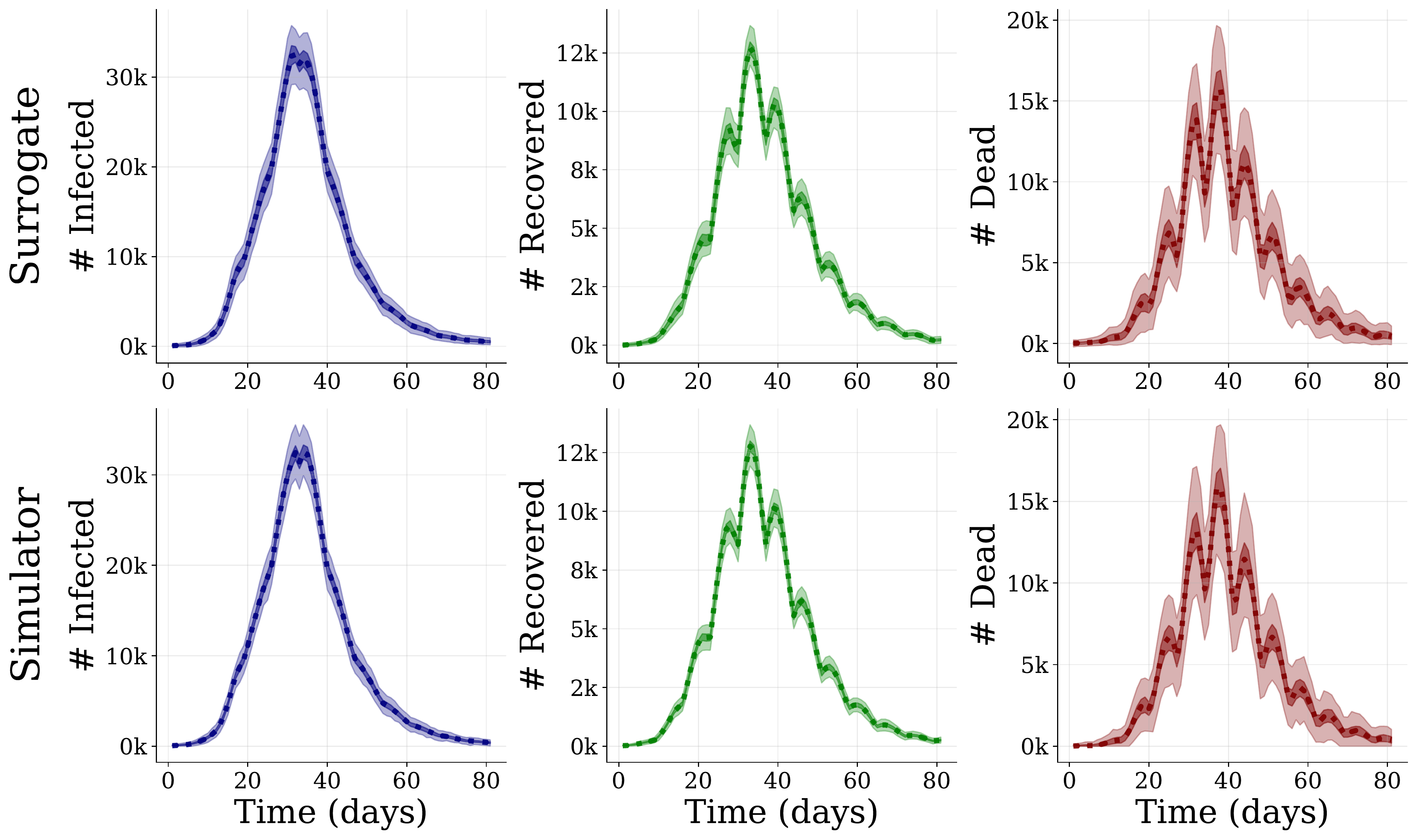}
	\hspace*{1cm}
	\covidscenario{Scenario IV}{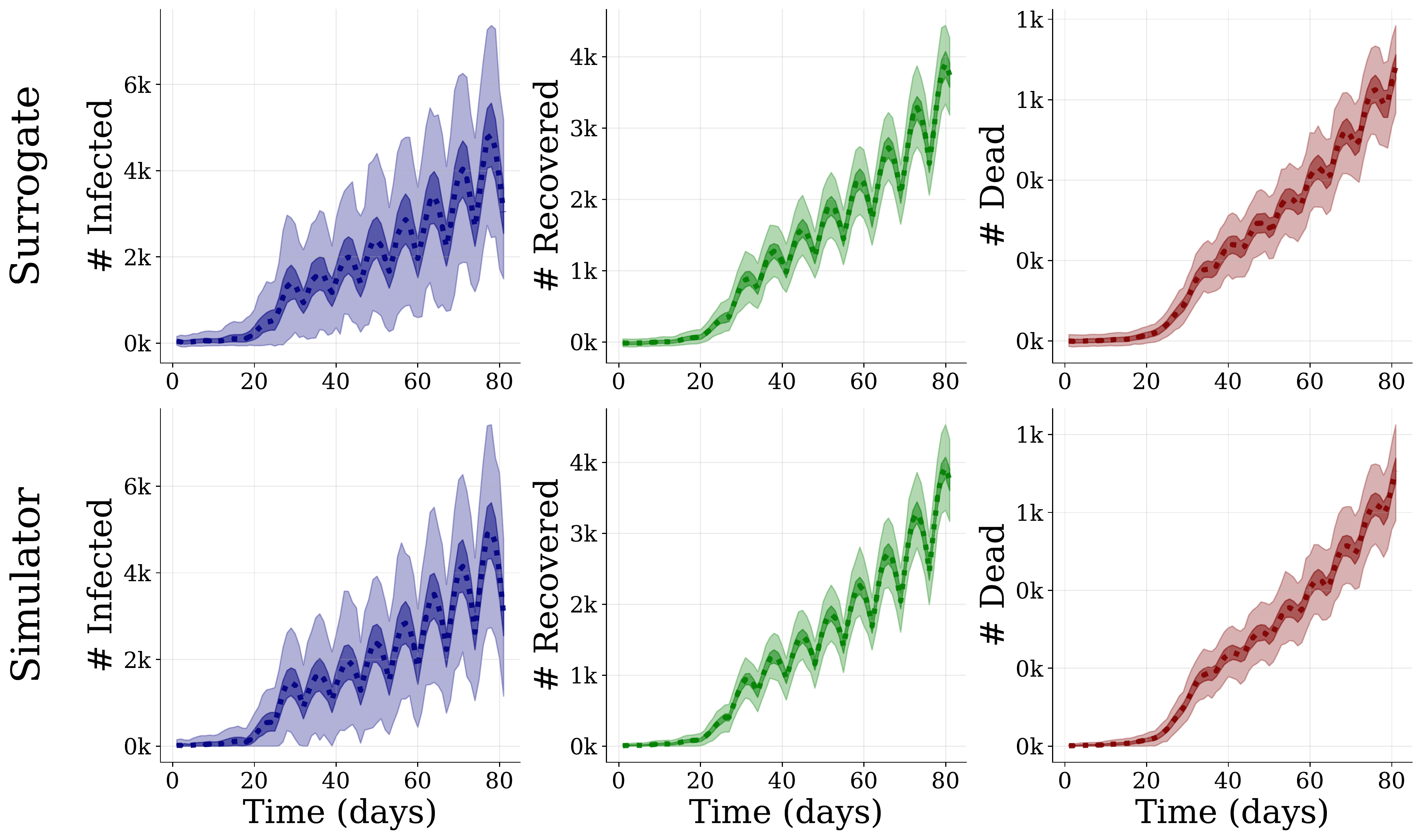}
	\\
	\covidscenario{Scenario V}{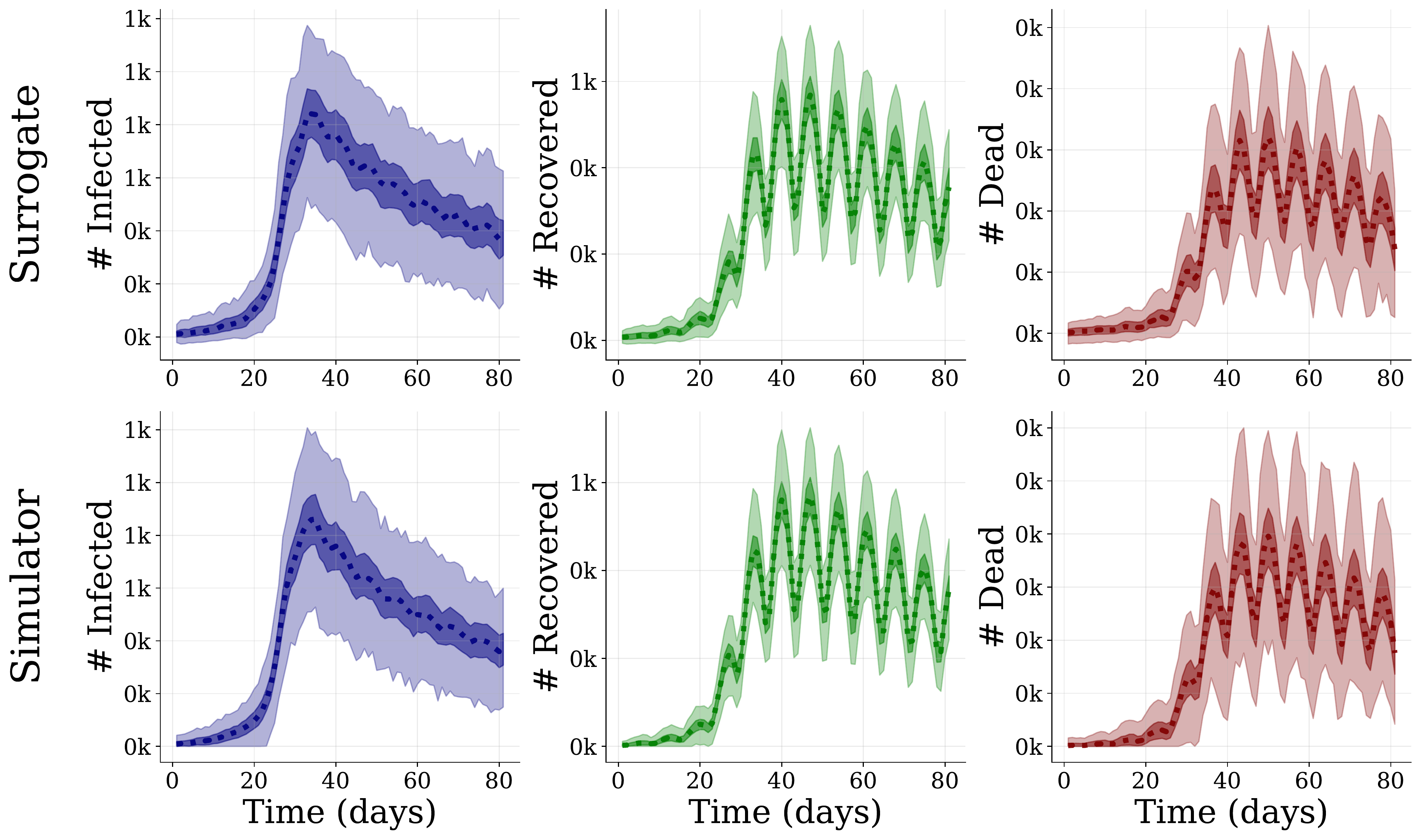}
	\hspace*{1cm}
	\covidscenario{Scenario VI}{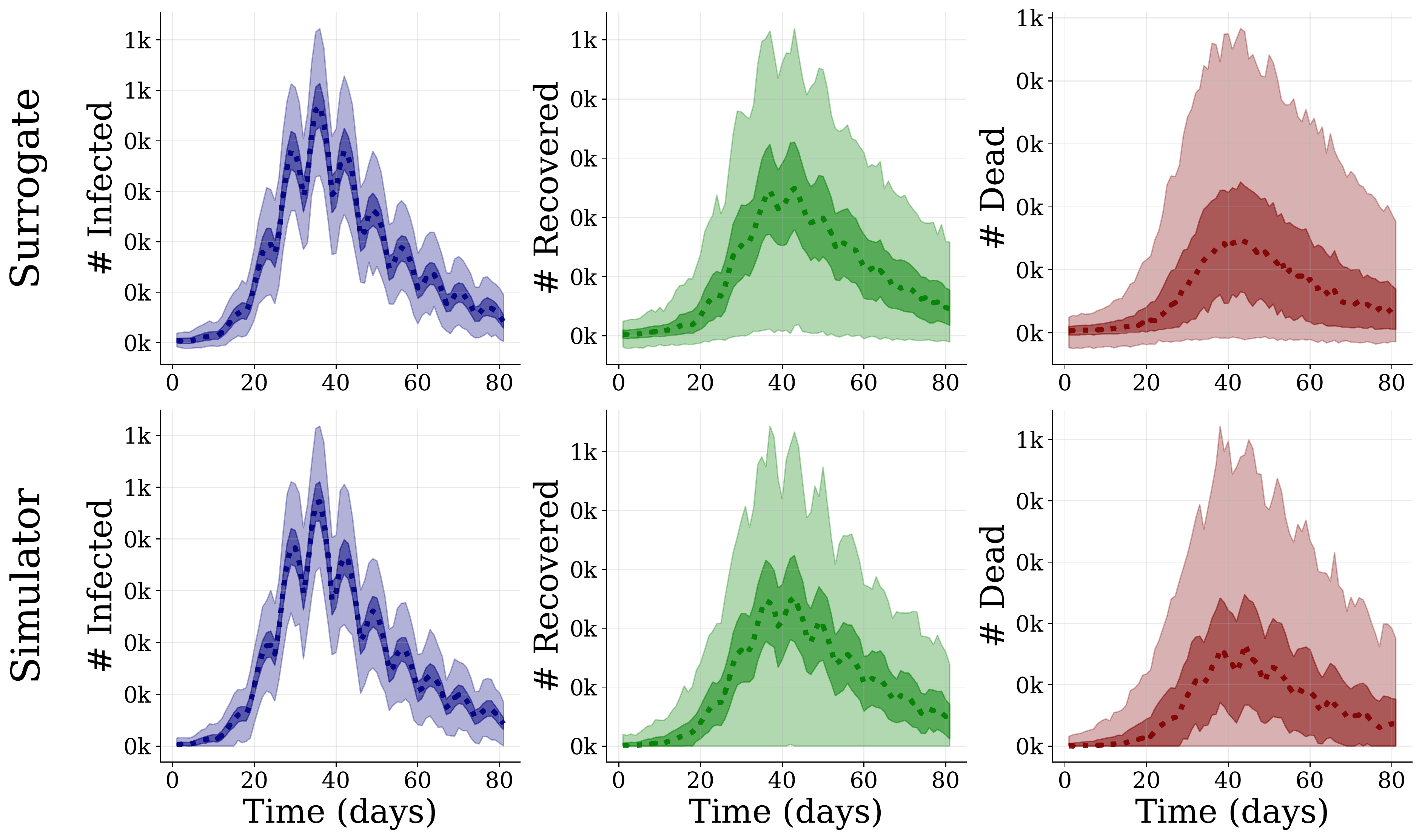}
	\\
	\covidscenario{Scenario VII}{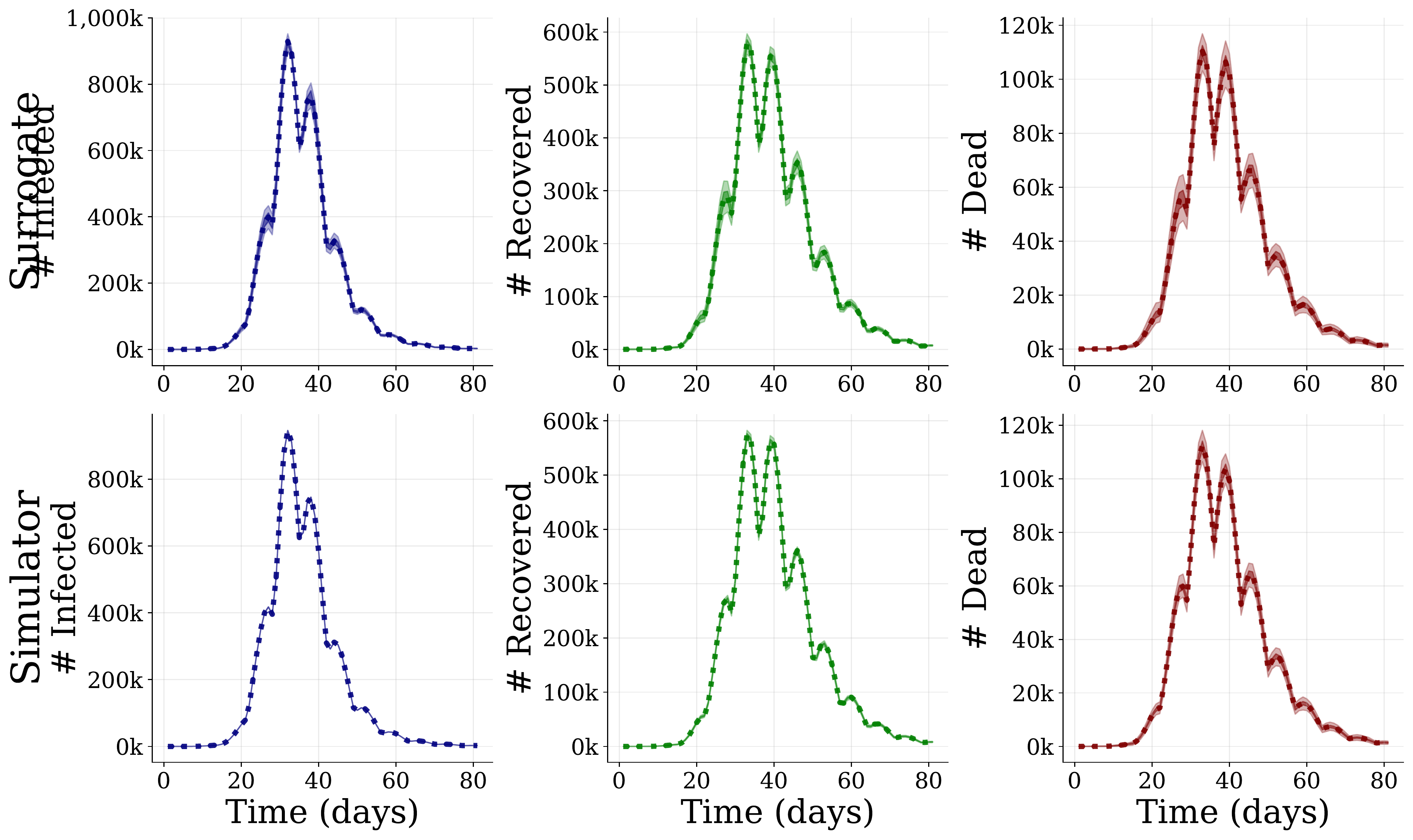}
	\hspace*{1cm}
	\covidscenario{Scenario VIII}{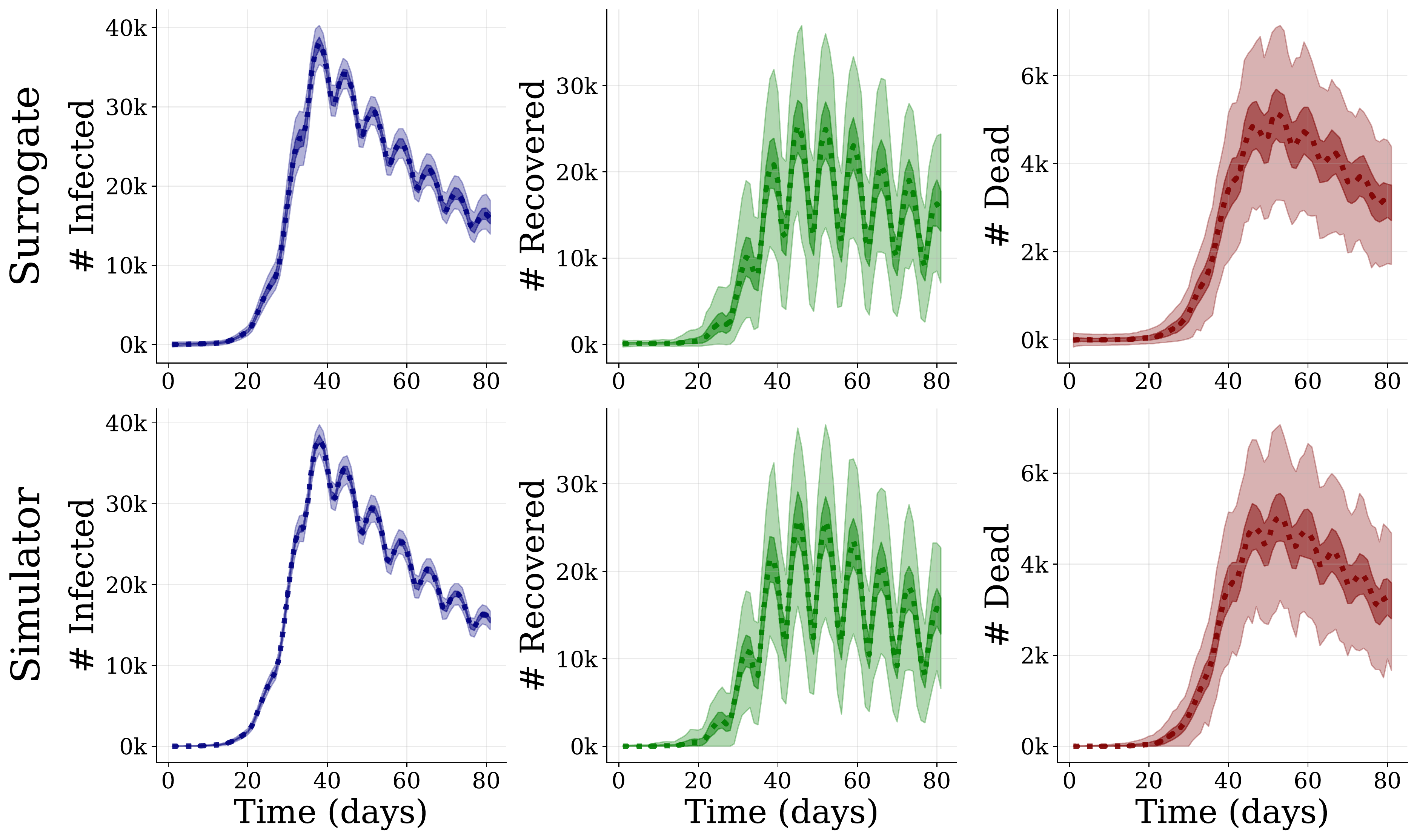}
	\\
	\includegraphics[width=0.17\linewidth]{plots/covid/legend_a.pdf}
	\includegraphics[width=0.17\linewidth]{plots/covid/legend_b.pdf}
	\includegraphics[width=0.17\linewidth]{plots/covid/legend_c.pdf}
	\caption{\textbf{Experiment 4.} The likelihood network can emulate the simulator and its aleatoric uncertainty strikingly well. Each sub-panel depicts $1000$ runs from the original and the surrogate neural simulator given the same parameter configuration, each leading to a qualitatively different outbreak scenario.}
\end{figure}

\begin{figure}
	\centering
	\begin{subfigure}[t]{0.62\linewidth}
		\includegraphics[width=\linewidth]{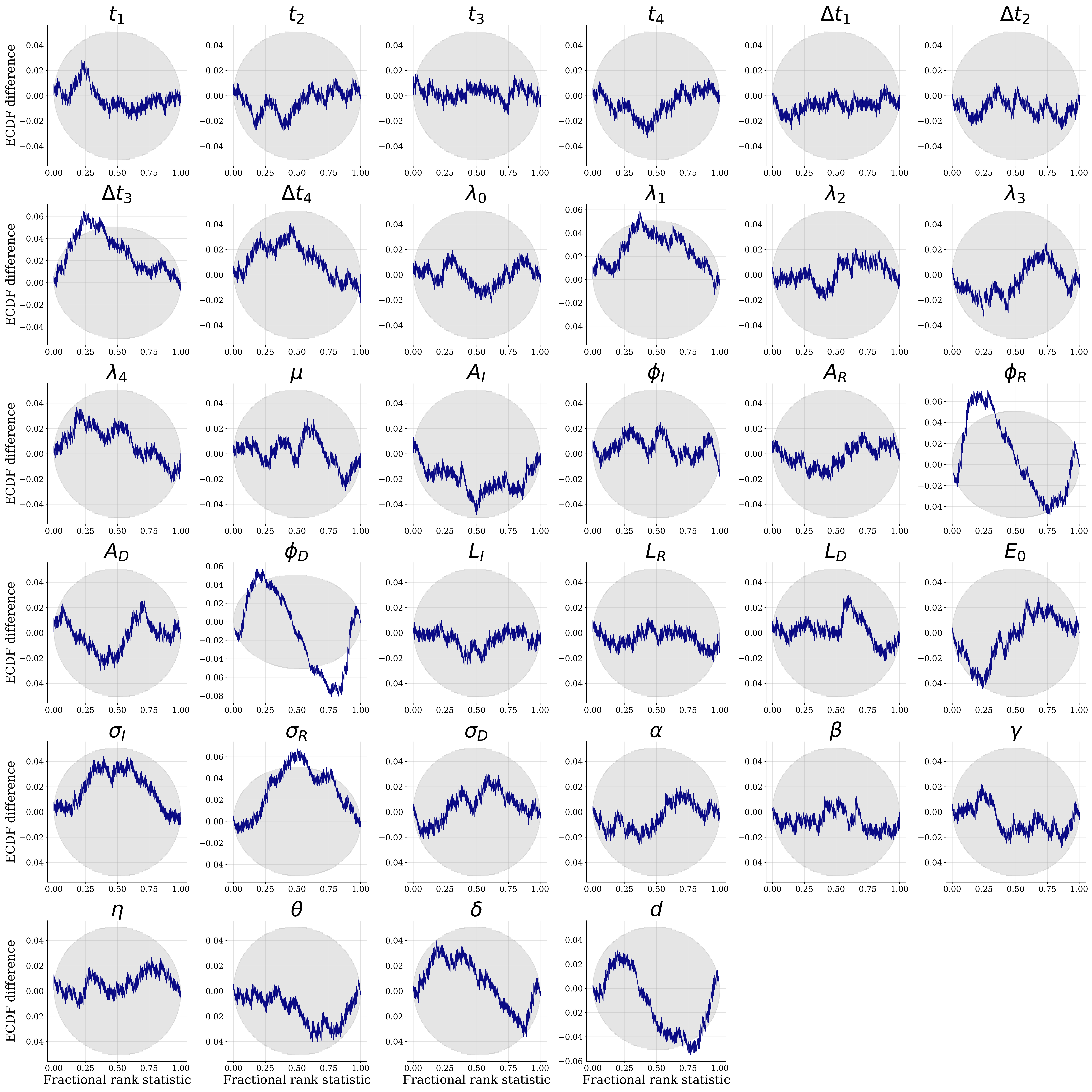}
		\caption{Posterior calibration}
	\end{subfigure}
	\begin{subfigure}[t]{0.62\linewidth}
		\includegraphics[width=\linewidth]{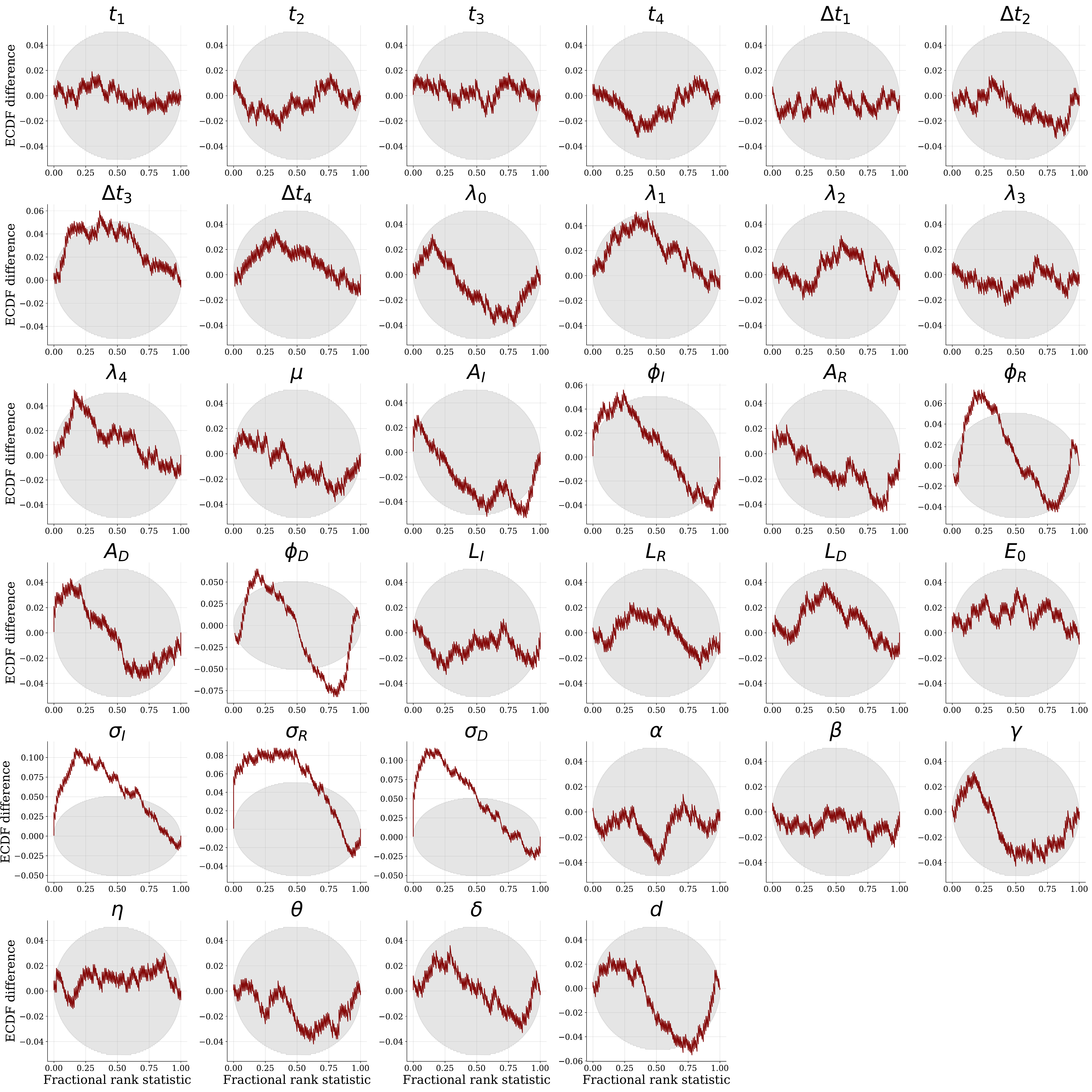}
		\caption{Joint calibration}
	\end{subfigure}
	\caption{\textbf{Experiment 4.} Posterior and joint calibration results}
	\label{fig:app:covid:calibration}
\end{figure}

\FloatBarrier
\clearpage
\subsection{Experiment 5: High-Dimensional Bayesian Denoising}

\paragraph{Model Details} This experiment follows the problem formulation from \textcite{ramesh2022gatsbi, pacchiardi2022score}.
However, we choose the Fashion MNIST data set because of its richer and more interesting structure. 
In this Bayesian denoising setup, a simulated noisy camera applies a multidimensional Gaussian filter (i.e., a blur) to each Fashion MNIST image.
Thus, the original image represents the ``parameters'' $\thetab \in \mathbb{R}^{784}$ and its blurry version $\mathbb{R}^{784}$ represents the ``observation''.
In order to make the problem more challenging, we do not use the class label as an additional conditioning input for the networks.
We also do not process the image data optimally \autocite[e.g., by applying a Haar wavelet downsampling or using convolutional couplings, as in][]{ardizzone2019guided, kingma2018glow}, as our goal is not to perform high-quality image reconstruction, but simply to illustrate the applicability of JANA for analyzing potentially high-dimensional Bayesian models.

\paragraph{Network and training details}

Since both ``data'' and ``parameters'' are images with a (theoretically) lower intrinsic dimensionality than the total number of pixels, both the likelihood and the posterior network utilize a separate summary network with identical architecture.
For each, we use a $4$-layer fully convolutional network with a final global average pooling layer yielding a $128$-dimensional summary representation of the original and blurry image, respective.
The posterior network is a conditional invertible neural network (cINN) comprising $12$ conditional affine coupling layers.
The internal networks of the coupling layers are fully connected (FC) networks with a single hidden layer of $512$ units and a \texttt{ReLU} non-linearity.
The likelihood network uses the same architecture as the posterior network.
Finally, we use a multivariate Student-T latent space \autocite{alexanderson2020robust}, as it allows us to perform a much more stable maximum likelihood training with higher learning rates.

We train the networks on the official training set of $60\,000$ Fashion MNIST images for $120$ epochs with a batch size of $32$ and a learning rate of $0.001$. 
This initial learning rate is reduced throughout the training phase following a cosine decay schedule with a minimum learning rate of $0$.
For each batch, we add a small amount of Gaussian noise with a scale of $0.001$ as a form of dequantization \autocite{ardizzone2019guided}.
We use $500$ images from the test set as a validation set to estimate the generalization error during training.
We utilize the remaining $9500$ images from the test set for evaluating the approximation quality and calibration of the networks.

\begin{figure}
	\centering
	\includegraphics[width=0.95\linewidth]{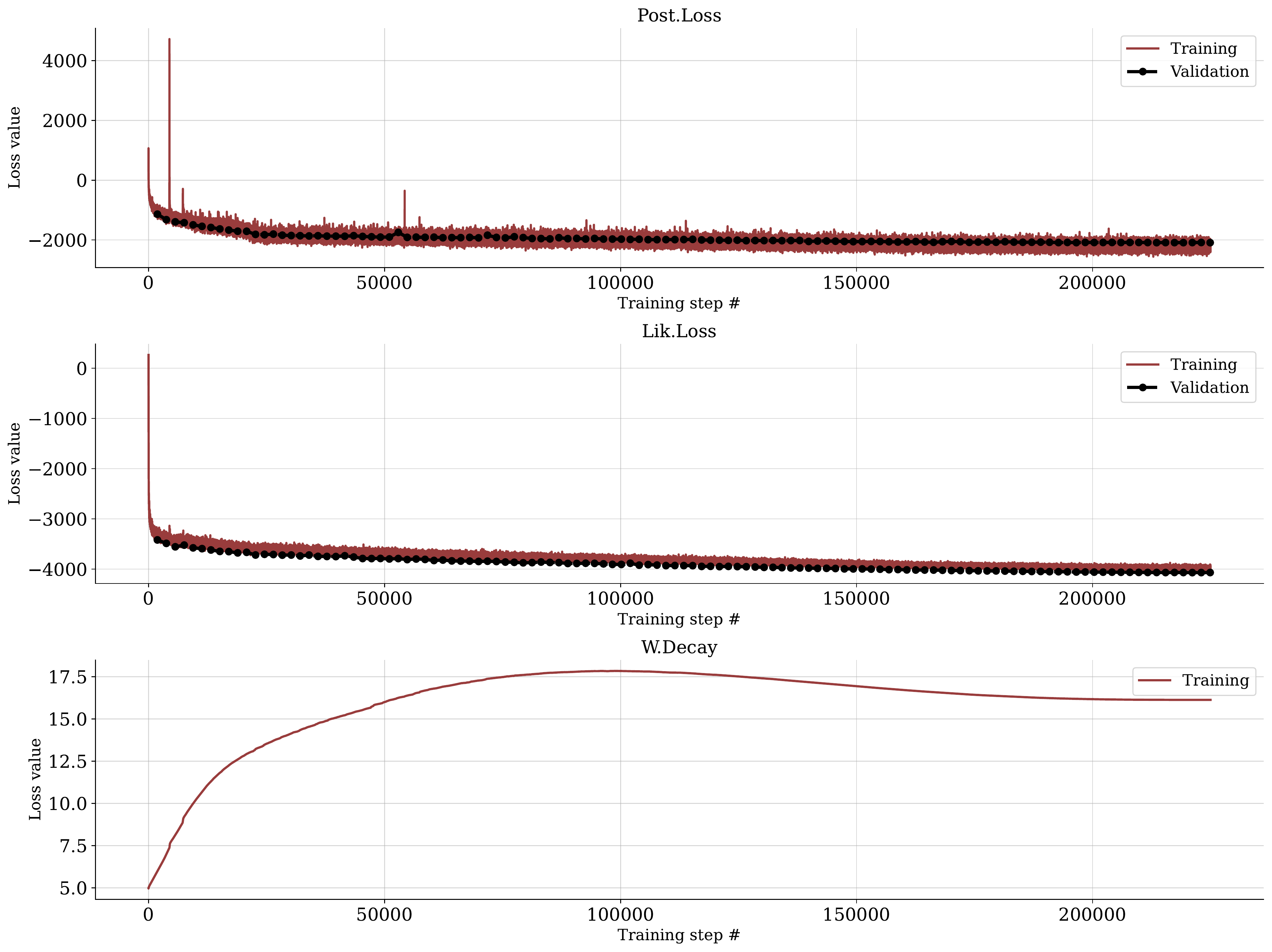}
	\caption{\textbf{Experiment 5.} Loss history}
	\label{fig:app:denoisinhg}
\end{figure}

\begin{figure}
	\centering
	\begin{subfigure}[t]{0.49\linewidth}
		\includegraphics[width=\linewidth]{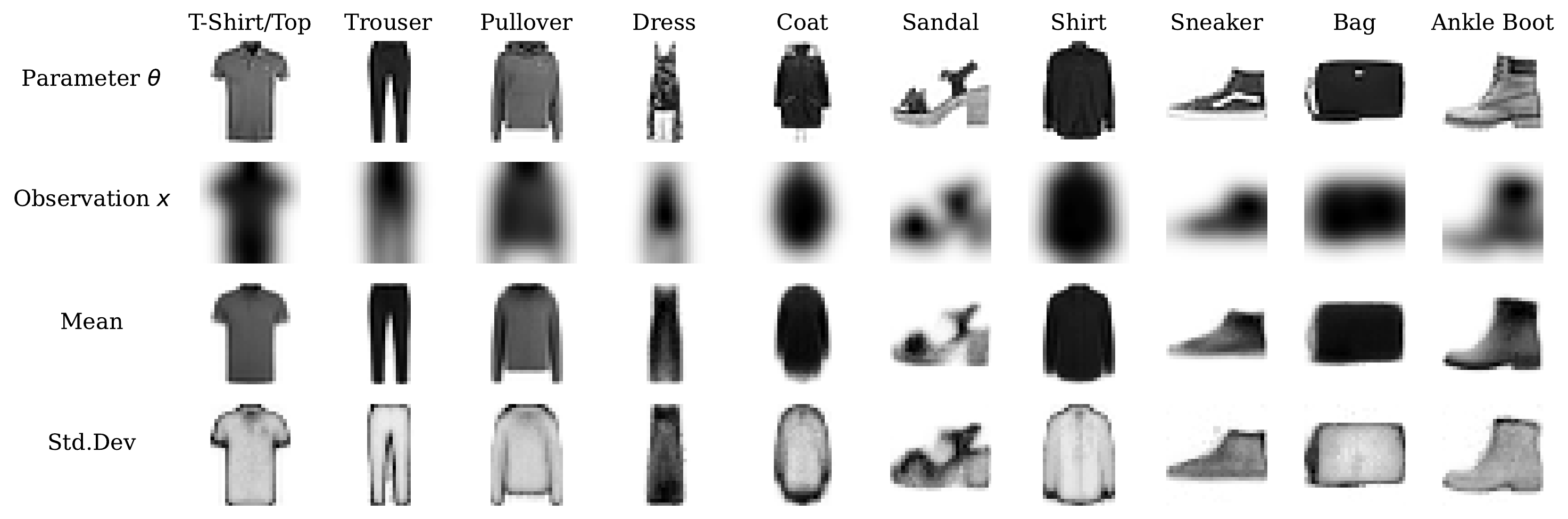}
		\caption{Posterior estimation \#1}
	\end{subfigure}
	\begin{subfigure}[t]{0.49\linewidth}
		\includegraphics[width=\linewidth]{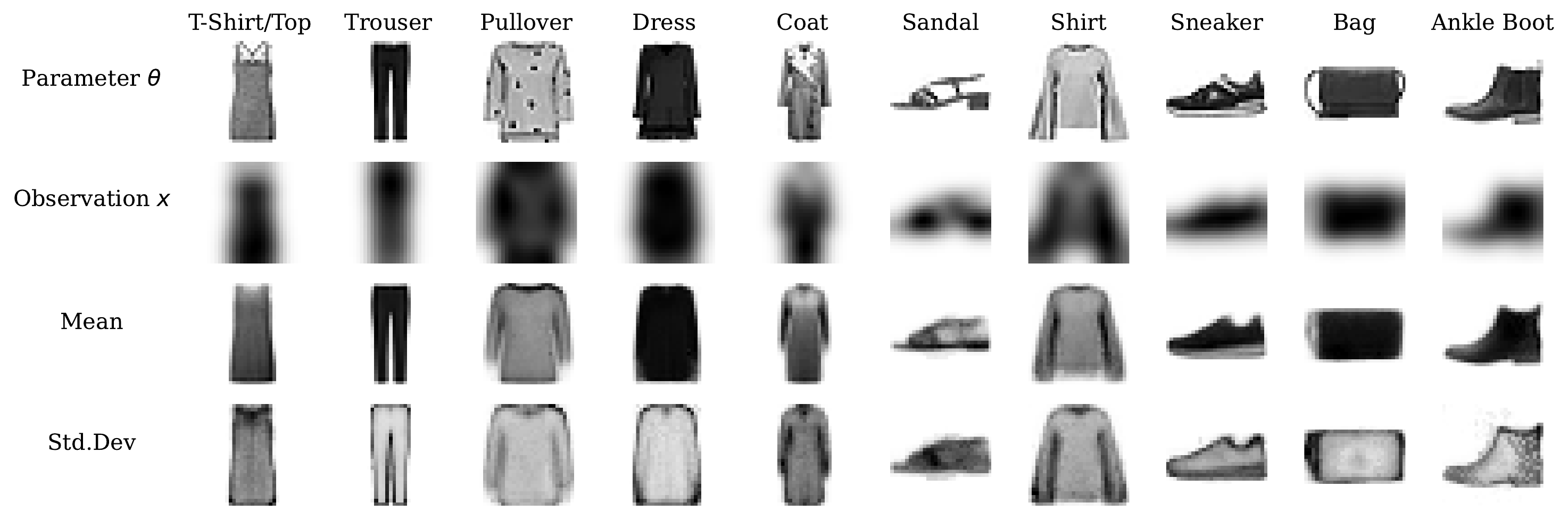}
		\caption{Posterior estimation \#2}
	\end{subfigure}
	\begin{subfigure}[t]{0.49\linewidth}
		\includegraphics[width=\linewidth]{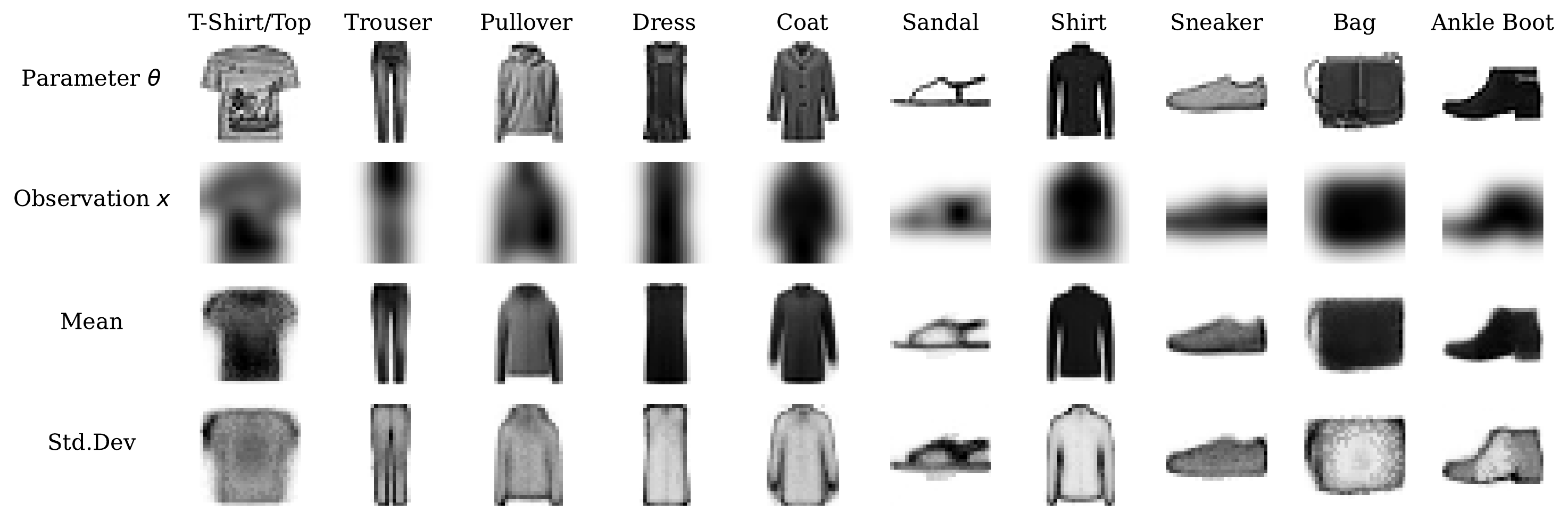}
		\caption{Posterior estimation \#3}
	\end{subfigure}
	\begin{subfigure}[t]{0.49\linewidth}
		\includegraphics[width=\linewidth]{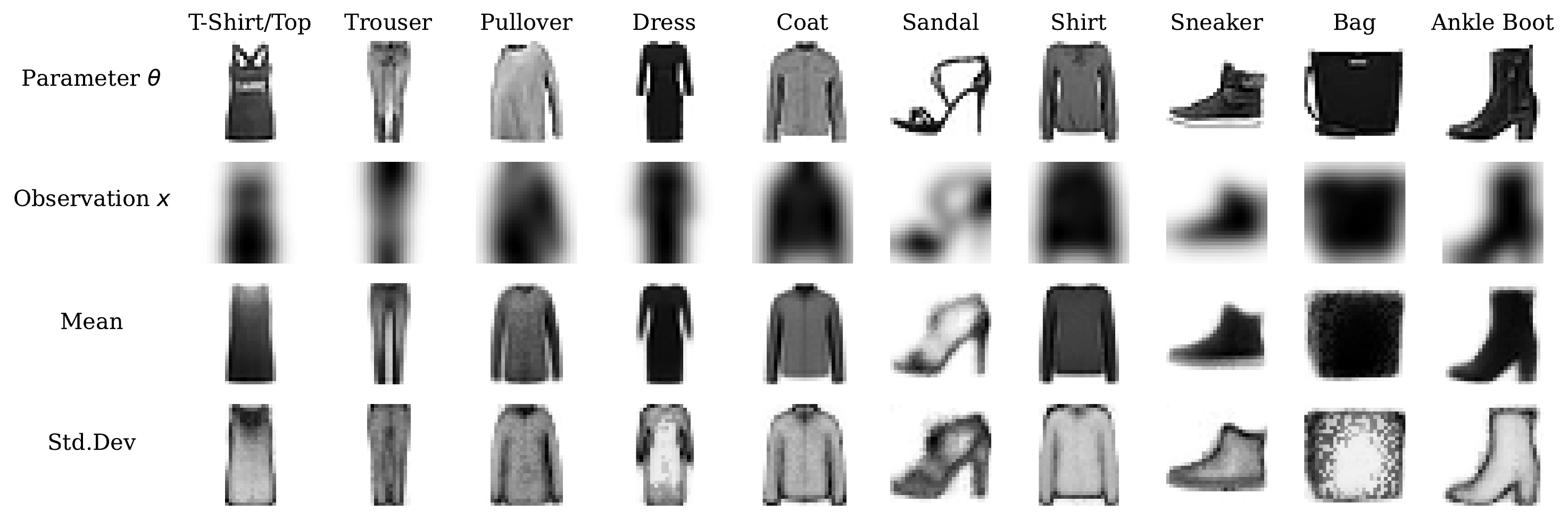}
		\caption{Posterior estimation \#4}
	\end{subfigure}
	\begin{subfigure}[t]{0.49\linewidth}
		\includegraphics[width=\linewidth]{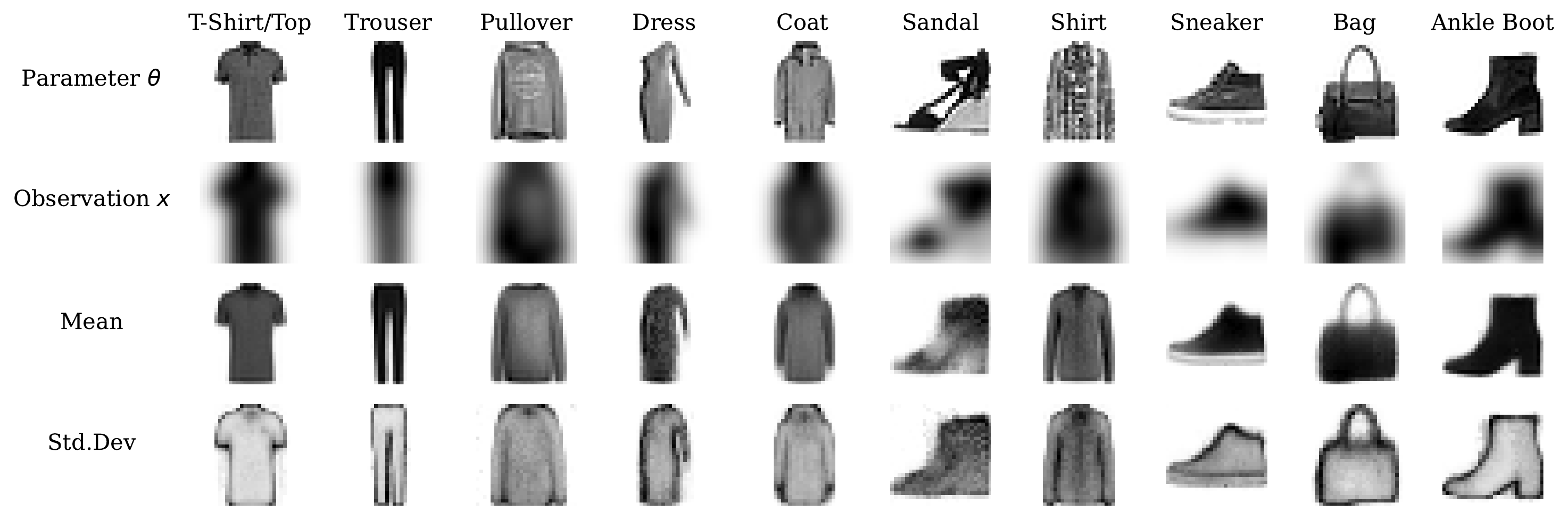}
		\caption{Posterior estimation \#5}
	\end{subfigure}
	\begin{subfigure}[t]{0.49\linewidth}
		\includegraphics[width=\linewidth]{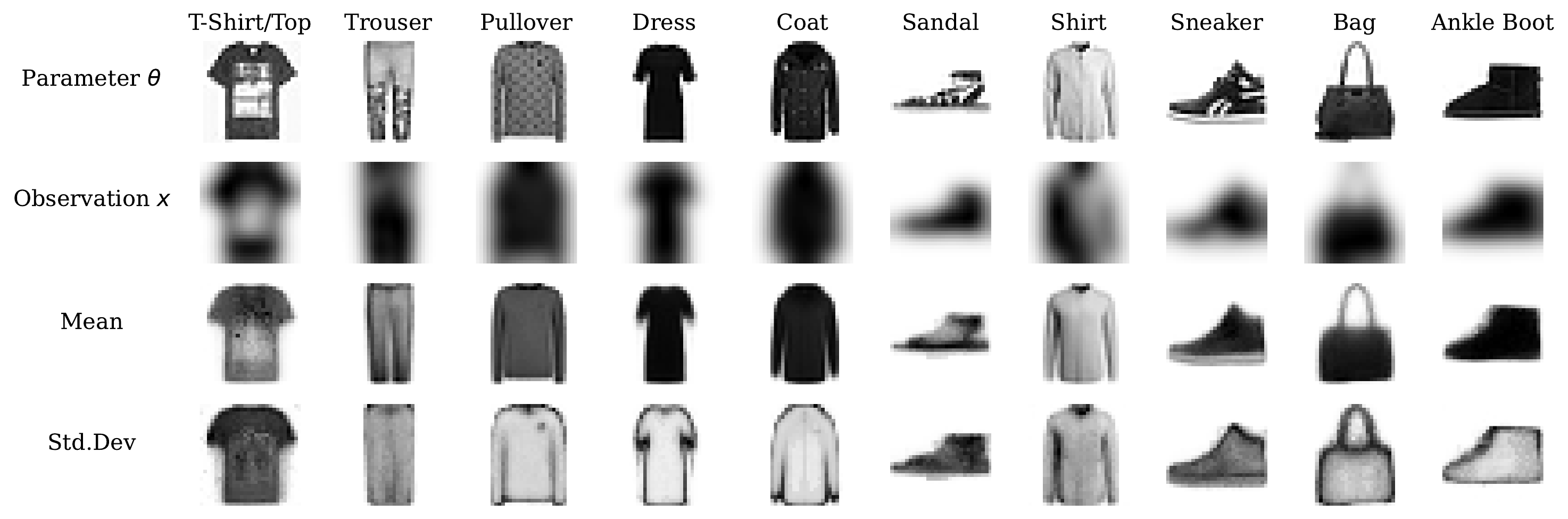}
		\caption{Posterior estimation \#6}
	\end{subfigure}
	\begin{subfigure}[t]{0.49\linewidth}
		\includegraphics[width=\linewidth]{plots/denoising/appendix_5.pdf}
		\caption{Posterior estimation \#7}
	\end{subfigure}
	\begin{subfigure}[t]{0.49\linewidth}
		\includegraphics[width=\linewidth]{plots/denoising/appendix_6.pdf}
		\caption{Posterior estimation \#8}
	\end{subfigure}
	\caption{\textbf{Experiment 5.} Posterior (denoising) results on $8$ randomly selected sets of images from the official Fashion MNIST test set.}
	\label{fig:app:denoising:posterior}
\end{figure}

\begin{figure}
	\centering
	\includegraphics[width=0.95\linewidth]{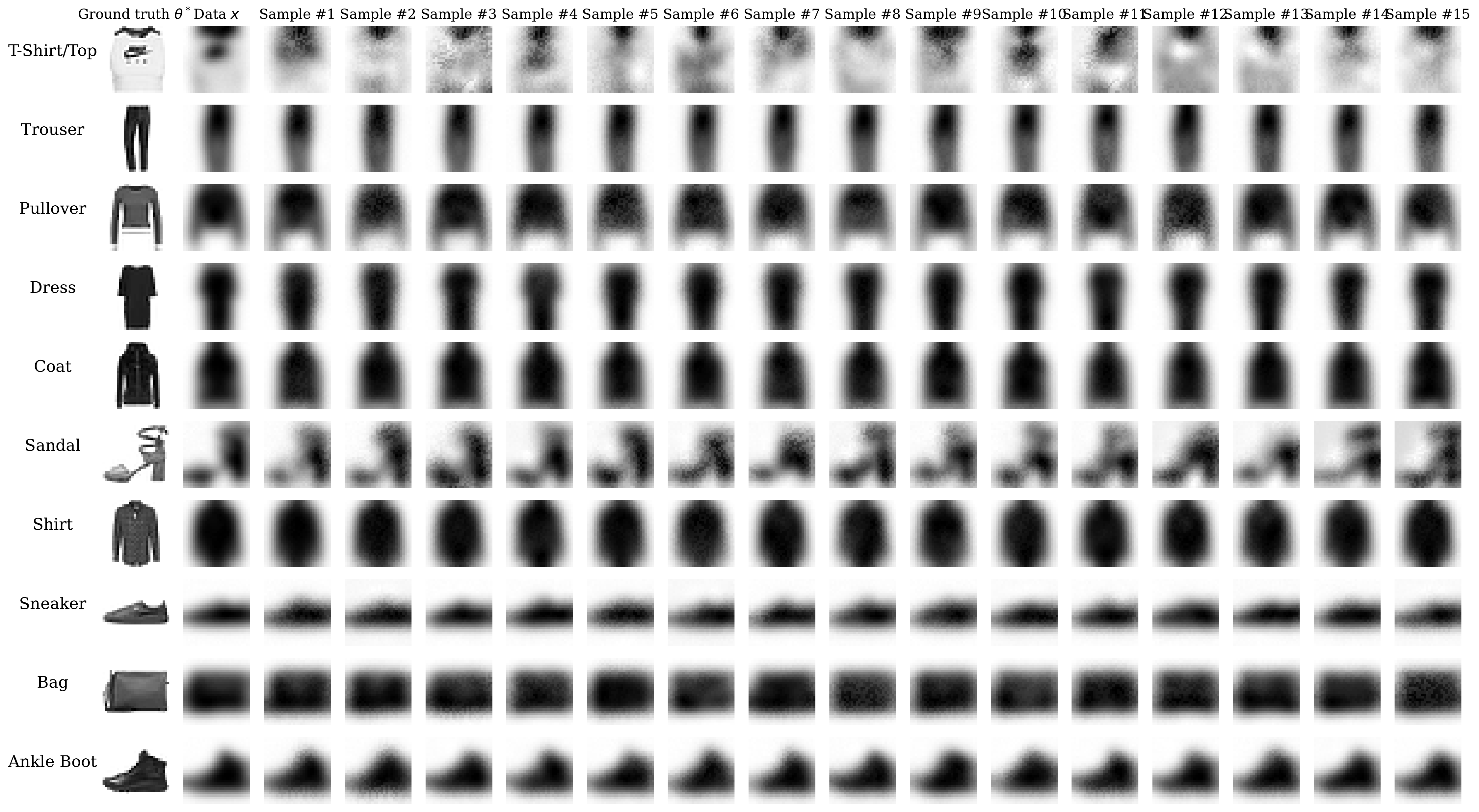}
	\caption{\textbf{Experiment 5.} Samples from the surrogate camera (i.e., the ``likelihood'' in the Bayesian denoising setup) given ten randomly selected clean images (i.e., ``parameters'') from each class.}
	\label{fig:app:denoising:likelihood1}
\end{figure}

\begin{figure}
	\centering
	\includegraphics[width=0.95\linewidth]{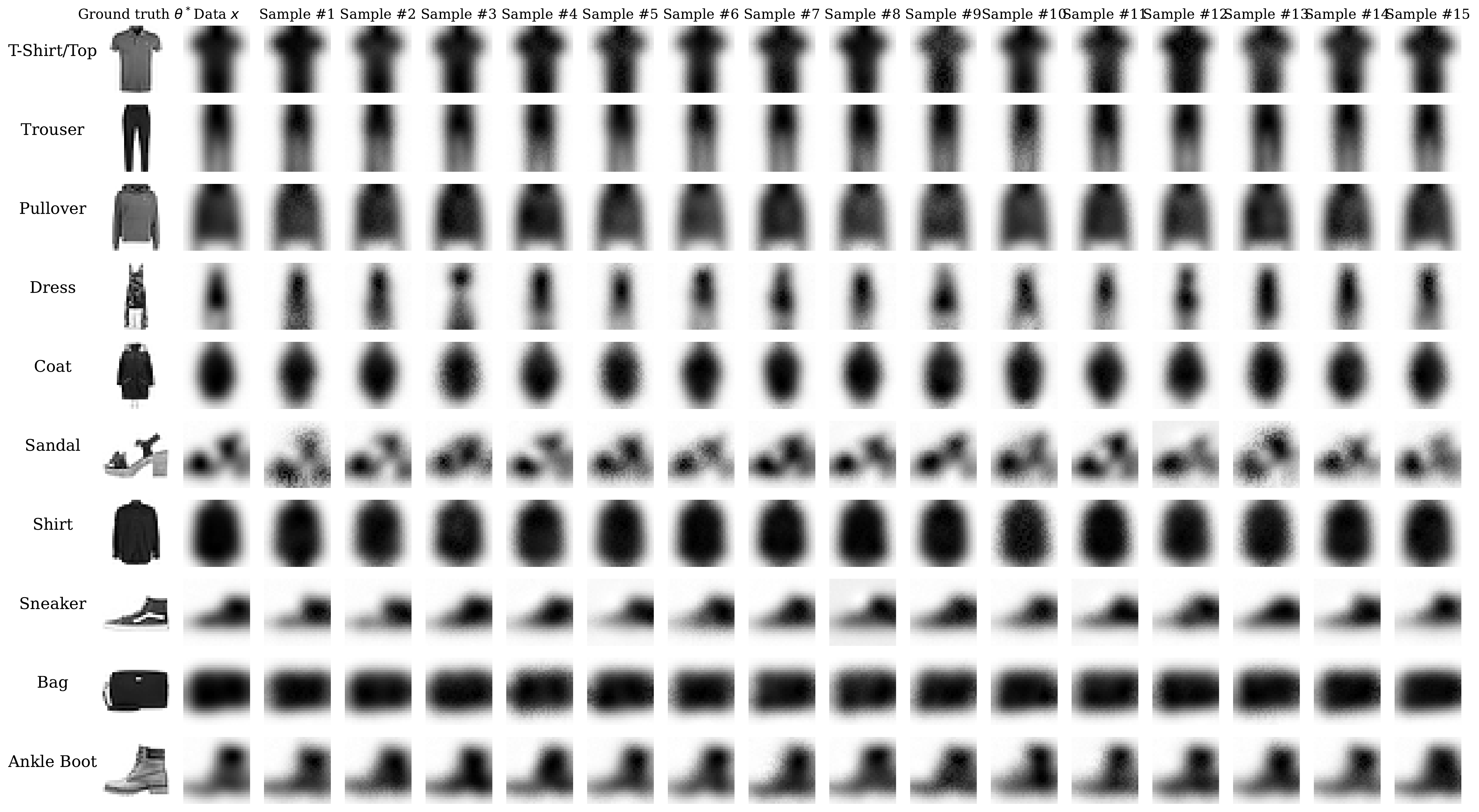}
	\caption{\textbf{Experiment 5.} Further samples from the surrogate camera.}
	\label{fig:app:denoising:likelihood2}
\end{figure}

\end{document}